\documentclass{article} 
\usepackage[final]{colm2026_conference}

\usepackage{microtype}
\usepackage{hyperref}
\usepackage{url}
\usepackage{booktabs}

\usepackage{graphicx}
\usepackage{subcaption} 
\usepackage{array}
\usepackage{enumitem}
\usepackage{booktabs}
\usepackage[table]{xcolor}
\usepackage{makecell}
\usepackage[most]{tcolorbox}
\usepackage{stfloats}
\fnbelowfloat
\usepackage{algpseudocodex}
\usepackage{algorithm}
\usepackage{amsmath, amssymb}
\usepackage{xcolor}
\usepackage{multirow}

\definecolor{stageblue}{HTML}{E8F0FE}
\definecolor{stagegreen}{HTML}{E6F4EA}
\definecolor{stageamber}{HTML}{FEF7E0}
\definecolor{stageblueborder}{HTML}{4285F4}
\definecolor{stagegreenborder}{HTML}{34A853}
\definecolor{stageamberborder}{HTML}{F9AB00}

\definecolor{codebg}{HTML}{F5F5F5}
\definecolor{codeframe}{HTML}{CCCCCC}
\definecolor{deepseek}{HTML}{4A6FA5}
\definecolor{qwen}{HTML}{7B2D8B}
\definecolor{olmo}{HTML}{2E7D32}
\definecolor{specialtoken}{HTML}{C62828}
\definecolor{commentgray}{HTML}{888888}

\definecolor{custompink}{RGB}{219, 48, 122}
 
\lstdefinestyle{chattemplate}{
    backgroundcolor=\color{codebg},
    frame=single,
    rulecolor=\color{codeframe},
    basicstyle=\ttfamily\scriptsize,
    breaklines=true,
    breakatwhitespace=false,
    tabsize=2,
    showstringspaces=false,
    xleftmargin=1em,
    xrightmargin=1em,
    aboveskip=0.8em,
    belowskip=0.8em,
}
 
\tcbuselibrary{skins, breakable}
 
\newtcolorbox{modelbox}[2][]{%
    enhanced,
    breakable,
    colback=white,
    colframe=#2,
    fonttitle=\bfseries\large,
    title=#1,
    boxrule=0.8pt,
    arc=3pt,
    left=6pt, right=6pt, top=4pt, bottom=4pt,
    toptitle=3pt, bottomtitle=3pt,
}

\def\githublogo{\raisebox{-1.5pt}{\includegraphics[height=1.05em]{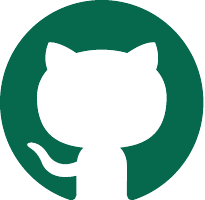}}}

\def\bloglogo{\raisebox{-2pt}{\includegraphics[height=1.05em]{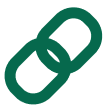}}}

\usepackage{lineno}

\definecolor{darkblue}{rgb}{0, 0, 0.5}
\definecolor{c-green}{RGB}{52,181,145}
\definecolor{c-green-light}{RGB}{221,247,240}
\definecolor{c-green-dark}{RGB}{6,105,77}
\definecolor{c-pink-light}{RGB}{255,246,249}
\definecolor{c-pink}{RGB}{218,123,153}
\definecolor{c-pink-dark}{RGB}{179,62,122}

\hypersetup{colorlinks=true, citecolor=darkblue, linkcolor=darkblue, urlcolor=darkblue}

\title{Why Does Self-Distillation (Sometimes) \\Degrade the Reasoning Capability of LLMs?}

\author{\parbox{\textwidth}{
\rule{0pt}{24pt}Jeonghye Kim$^{1,2}$\thanks{Work done during an internship at Microsoft Research. \textsuperscript{$\dagger$}Corresponding author.}, Xufang Luo\textsuperscript{1$\dagger$}, Minbeom Kim$^3$, Sangmook Lee$^4$, \\ Dohyung Kim$^4$,
Jiwon Jeon$^2$, Dongsheng Li$^1$, Yuqing Yang$^1$ \\[4pt]
\textmd{$^1$Microsoft Research\, $^2$KAIST\, $^3$SK hynix\, $^4$Seoul National University}
} \\
\bloglogo ~ {\small \href{https://beanie00.notion.site/why-does-self-distillation-degrade-reasoning?source=copy_link}{\texttt{blog post}}}
~~
\githublogo ~ {\small \href{https://github.com/beanie00/self-distillation-analysis}{\texttt{beanie00/self-distillation-analysis}}}
} 

\begin{document}

\ifcolmsubmission
\linenumbers
\fi

\maketitle


\begin{abstract}
Self-distillation has emerged as an effective post-training paradigm for LLMs, often improving performance while shortening reasoning traces. However, in mathematical reasoning, we find that it can reduce response length while degrading performance. We trace this degradation to the suppression of \textbf{epistemic verbalization}—the model's expression of uncertainty during reasoning. Through controlled experiments varying \textbf{conditioning context richness} and \textbf{task coverage}, we show that conditioning the teacher on rich information suppresses uncertainty expression, enabling rapid in-domain optimization with limited task coverage but harming OOD performance, where unseen problems benefit from expressing uncertainty and adjusting accordingly. Across Qwen3-1.7B/8B, DeepSeek-Distill-Qwen-7B, and Olmo3-7B-Instruct, we observe performance drops of up to 40\%. Our findings highlight that exposing appropriate levels of uncertainty is crucial for robust reasoning and underscore the importance of optimizing reasoning behavior beyond merely reinforcing correct answer traces.
\end{abstract}

\section{Introduction}

\begin{figure}[b]
    \centering
    \vspace{-0.2cm}
    \begin{subfigure}[b]{0.49\linewidth}
        \centering
        \includegraphics[width=\linewidth]{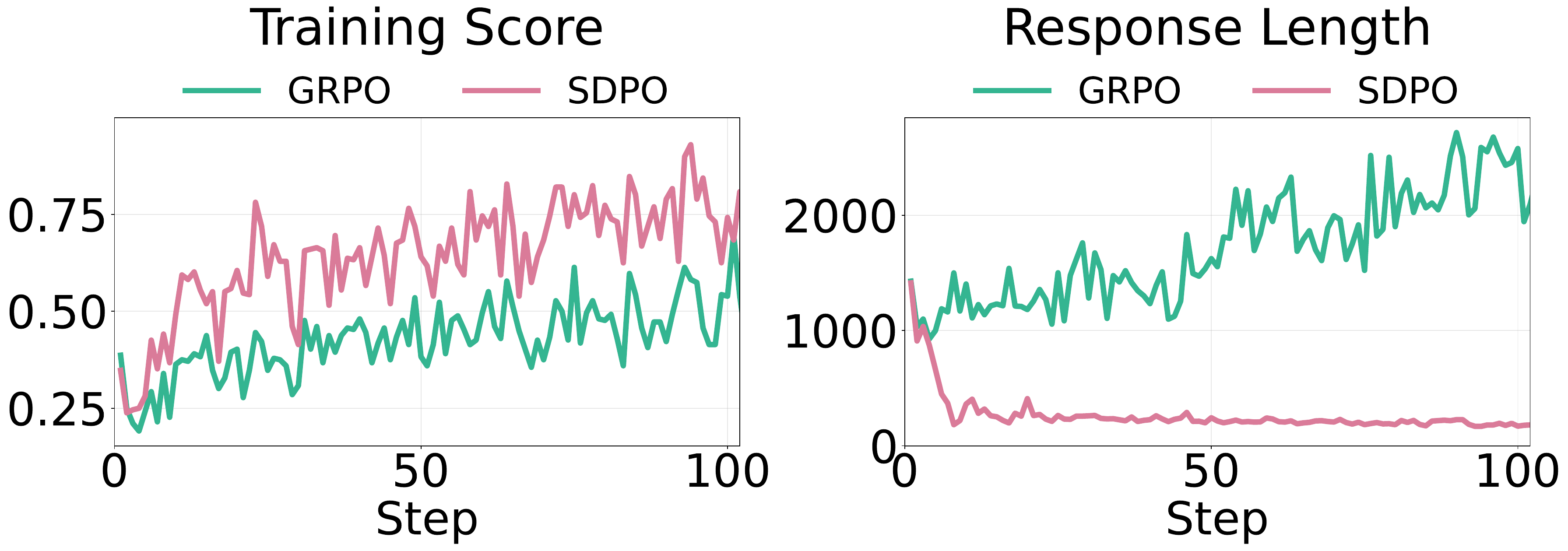}
        \caption{\textbf{Chemistry} (Olmo3-7B-Instruct)}
        \label{fig:sub-a}
    \end{subfigure}
    \hfill
    \begin{subfigure}[b]{0.49\linewidth}
        \centering
        \includegraphics[width=\linewidth]{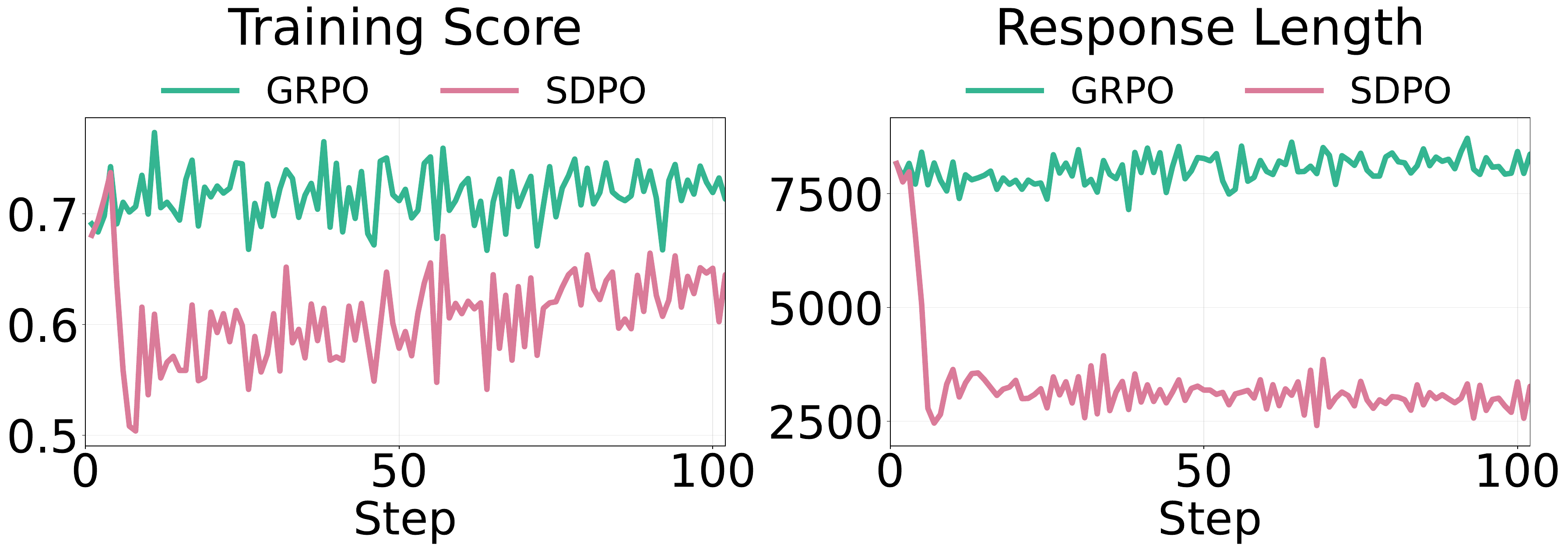}
        \caption{\textbf{DAPO-Math-17k} (DeepSeek-Distill-Qwen-7B)}
        \label{fig:sub-b}
    \end{subfigure}
    \caption{(a) Training score and response length changes for GRPO and Self-Distillation (SDPO) \citep{SDPO} in Chemistry, using results from SDPO Wandb logs \citep{wandb} 
\href{https://wandb.ai/jonhue/SDPO?nw=mgotcx6kk7}{\texttt{(link)}}. (b) Training score and response length changes on DAPO-Math-17k with GRPO and SDPO.}
    \label{fig:motivation}
\end{figure}

Recently, self-distillation \citep{self-distillation} has attracted increasing attention in the post-training of large language models (LLMs). In this paradigm, two instances of the same model are employed: one conditioned on privileged information (e.g., ground-truth solutions) serves as a teacher, providing dense reward signals for responses generated by another instance that does not have access to that information. Several studies have demonstrated that combining this framework with post-training methods such as Reinforcement Learning from Verifiable Rewards (RLVR) leads to highly efficient performance gains \citep{zhu2025token, SDPO, zhao2026self, privileged, empo2, opcd}. These methods have shown particularly strong improvements in domains such as agentic environments and scientific reasoning, especially under in-domain evaluation settings. Interestingly, a consistent trend observed across these works is that performance improves as response length decreases, suggesting that self-distillation promotes more concise and effective reasoning.

However, when we apply the same self-distillation approach to mathematical reasoning tasks, we observe a markedly different phenomenon. Figure~\ref{fig:motivation} compares the effects of a representative self-distillation algorithm, SDPO, in the Chemistry domain (a) and the Math domain (b). As shown in the figure, in the Chemistry domain, self-distillation substantially reduces response length compared to GRPO while rapidly improving performance. In contrast, in the Math domain, although response length consistently decreases as training progresses, performance drops significantly, contrary to prior findings.

This raises a question: \textbf{\emph{ "Why does performance sometimes degrade despite the model being trained to move toward the correct answer?"}} 

Our analysis reveals a consistent pattern: the more informative the context provided to the teacher, the more concise and confident the resulting reasoning becomes, with substantially fewer expressions of uncertainty and, particularly in math reasoning, degraded performance. We trace this effect to the suppression of \textit{epistemic verbalization} \citep{understanding}, whereby models explicitly verbalize and incorporate uncertainty during reasoning. Strong reasoning
models such as DeepSeek-R1 \citep{deepseek-r1} frequently express uncertainty using tokens like ``Wait'' or ``Hmm''. Although these expressions may not directly advance the reasoning, removing them discards important signals that a reasoning path may be flawed, leading to significant performance drops \citep{understanding}.

To systematically understand when and why self-distillation suppresses epistemic verbalization and thus affects performance, we conduct a comprehensive empirical study and identify two key factors: \textbf{information richness} and \textbf{task coverage}. The more privileged the information the teacher receives, such as the correct solution, the more its reasoning traces suppress expressions of uncertainty, encouraging the student to imitate a confident reasoning style that presupposes information unavailable at inference time. When task coverage is limited, this compression enables rapid in-domain optimization. However, as coverage increases, removing epistemic verbalization can interfere with optimization across diverse tasks, degrading performance on more challenging or unseen problems.

More broadly, our results show that even when the training objective faithfully guides the model toward correct reasoning traces, the resulting reasoning style can quietly shift in ways that hurt generalization. The suppression of epistemic verbalization is not penalized by standard objectives, yet negatively impacts out-of-distribution (OOD) performance. This suggests that post-training objectives need to account not only for answer correctness, but also for eliciting and preserving uncertainty-aware reasoning behaviors. We believe these findings offer a useful step toward a deeper understanding of reasoning in self-distillation and post-training more broadly.


\section{Preliminaries}

\paragraph{Self-Distillation} ~~ Let $x \in \mathcal{X}$ denote an input and $y=(y_1,\dots,y_T)$ a sequence generated by a language model $\pi_\theta$. 
The model defines an autoregressive distribution
$
\pi_\theta(y|x) = \prod_{t=1}^{T} \pi_\theta(y_t \mid x, y_{<t}).
$
In self-distillation, the same model acts as both a \emph{student} and a \emph{teacher} under different conditioning contexts. 
The student first generates a sequence $y \sim \pi_\theta(\cdot \mid x)$. 
The teacher policy is obtained by conditioning the model on a \emph{richer context} $c$ that provides additional information about the input (e.g., solutions, environment feedback, or other auxiliary signals). Training minimizes the divergence between the student and teacher next-token distributions:
\begin{equation}
\mathcal{L}_{\mathrm{\text{Self-Distillation}}}(\theta)
=
\sum_{t}
\mathrm{KL}\!\left(
\pi_\theta(\cdot \mid x, y_{<t})
\;\|\;
\mathrm{stopgrad}\big(\pi_\theta(\cdot \mid x,c,y_{<t})\big)
\right).
\end{equation}
This objective encourages the student to match the teacher's predictions under the richer context, enabling the model to improve by distilling privileged information available at training time without requiring an external teacher. 

\paragraph{Key Characteristics of Math Reasoning} ~~ 
In LLMs, math reasoning can be viewed as \textit{self-Bayesian reasoning}, where each step is generated conditioned only on the prompt $x$ and previously generated tokens $y=(y_1,\dots,y_T)$, allowing the model to iteratively update its belief over intermediate hypotheses \citep{understanding}. Such reasoning spans diverse tasks such as arithmetic, algebra, geometry, word problems, and logical pattern recognition, so evaluation benchmarks are frequently OOD relative to training data due to compositional and reasoning-depth shifts.

\captionsetup[subfigure]{labelformat=simple}
\renewcommand{\thesubfigure}{(\thefigure\alph{subfigure})}

\begin{figure}[t!]
    \centering
    \begin{subfigure}[b]{0.6\linewidth}
        \centering
        \includegraphics[width=\linewidth]{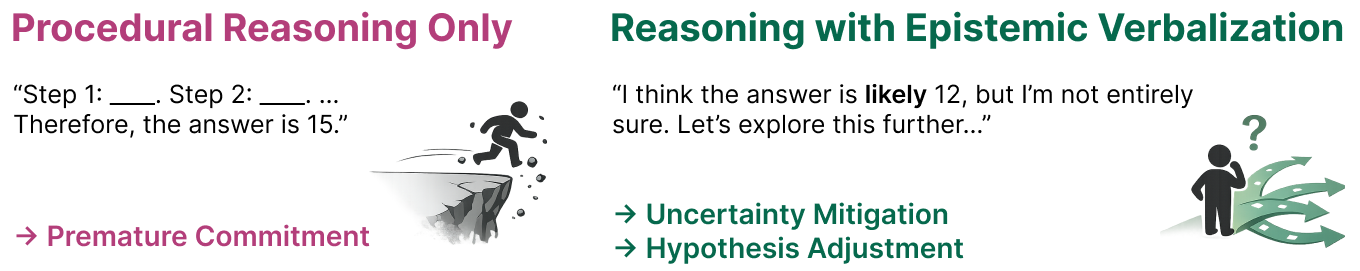}
        \caption{Reasoning with Epistemic Verbalization}
        \label{fig:epistemic}
    \end{subfigure}
    \hfill
    \begin{subfigure}[b]{0.39\linewidth}
        \centering
        \includegraphics[width=\linewidth]{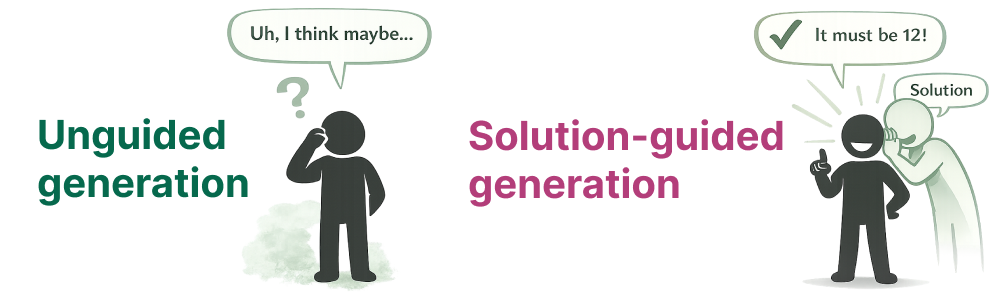}
        \caption{Unguided vs.\ Solution-Guided}
        \label{fig:guided}
    \end{subfigure}
\end{figure}

\captionsetup[subfigure]{labelformat=parens}
\renewcommand{\thesubfigure}{\alph{subfigure}}

Within this process, verbalized uncertainty toward correct response $y^*$, referred to as \textit{epistemic verbalization} \citep{understanding}, can serve as an informative signal rather than stylistic redundancy. As shown in Figure~\ref{fig:epistemic}, reasoning without such signals may prematurely commit to incorrect hypotheses with little room for recovery, whereas epistemic verbalization preserves alternative hypotheses and supports gradual uncertainty reduction.

\newcolumntype{P}[1]{>{\centering\arraybackslash}p{#1}}

\section{LLM Reasoning Behavior Under Richer Information} \label{sec:llm-reasoning-behavior}


In self-distillation, the teacher has access to a richer context $c$, which provides strong hints and enables it to generate reasoning traces with minimal expressed uncertainty (Figure \ref{fig:guided}). While this yields more concise responses, it could eliminate not only unnecessary verbosity but also valuable epistemic signals, potentially hindering the student's ability to perform uncertainty-aware reasoning. This raises two questions we investigate: (i) how does the richness of $c$ relate to the suppression of uncertainty expression and to reasoning performance? and (ii) if epistemic verbalization is beneficial, why does suppressing it improve performance in domains such as Chemistry while degrading it in Math?


To study these questions, we first need a way to quantify how informative the conditioning context is. Let $y^*$ denote a random variable representing an ideal correct response to $x$. We formalize the information that $c$ provides about $y^*$ as the conditional mutual information
\begin{equation}
I(y^*;\,c \mid x)
\;=\;
H(y^* \mid x) \;-\; H(y^* \mid x,\,c),
\end{equation}
which captures the reduction in uncertainty about $y^*$ once the additional context $c$ is given.

Building on this measure, we construct a controlled comparison to examine how the model's reasoning behavior changes as the conditioning information is varied. We use the DAPO-Math-17k dataset \citep{dapo} with DeepSeek-R1-Distill-Qwen-7B \citep{deepseek-r1} as the base model, selecting 100 problems on which the base model achieves accuracy between 0.125 and 0.5 over 8 rollouts.
Let $s$ denote the full solution (including chain-of-thought in 
\texttt{<think>} tags), $s_{\setminus\text{think}}$ the solution with 
\texttt{<think>} content removed, and $y_r \sim \pi_\theta(\cdot \mid x, s)$ 
a response previously generated under full solution guidance.
We compare the model's responses across four generation settings:

\begin{itemize}[itemsep=0pt, topsep=0pt, partopsep=0pt, leftmargin=*]
\item \textbf{(1) Unguided generation} ($c = \emptyset$): 
$I(y^*;\,c \mid x) = 0$ by definition.

\item \textbf{(2) Solution-guided generation} ($c = s$): 
provides maximal guidance and yields the largest $I(y^*;\,c \mid x)$.

\item \textbf{(3) Solution-guided generation without think contents} 
($c = s_{\setminus\text{think}}$): since $s_{\setminus\text{think}}$ is 
a strict informational subset of $s$, we have 
$I(y^*;\,s_{\setminus\text{think}} \mid x) \leq I(y^*;\,s \mid x)$.

\item \textbf{(4) Regeneration-conditioned generation} ($c = y_r$, 
where $y_r \sim \pi_\theta(\cdot \mid x, s)$): since $y_r$ is 
derived from $s$, the data processing inequality gives 
$I(y^*;\,y_r \mid x) \leq I(y^*;\,s \mid x)$.

\end{itemize}

\noindent \noindent These settings induce the following ordering over the conditional mutual information:
\begin{equation}\label{eq:mi_ordering}
\underbrace{I(y^*;\,\emptyset \mid x) = 0}_{\text{(1)}}
\;<\;
\underbrace{I(y^*;\,s_{\setminus\text{think}} \mid x)}_{\text{(3)}}
\;\leq\;
\underbrace{I(y^*;\,y_r \mid x)}_{\text{(4)}}
\;\leq\;
\underbrace{I(y^*;\,s \mid x)}_{\text{(2)}}.
\end{equation}

\paragraph{Prompts} ~~ The prompts used for unguided and solution-guided settings are as follows. For regeneration, we used the same prompts as in \cite{SDPO}. A comparative analysis with other regeneration prompts is provided in Appendix \ref{appendix:different_prompt}.

\begin{table}[h!]
\centering
\small
\begin{tabular}{p{3.2cm} p{9.8cm}}
\toprule
Prompt for unguided generation & \texttt{\{question\}} \newline Please reason step by step, and put your final answer within \textbackslash boxed\{\}. \\
\midrule
Regeneration prompt (followed the prompt in \cite{SDPO}) & \texttt{\{question\}} \newline Please reason step by step, and put your final answer within \textbackslash boxed\{\}. \newline \textcolor{c-pink-dark}{Correct solution: \texttt{\{previously correct solution\}}} \newline Correctly solve the original question. \\
\bottomrule
\end{tabular}
\end{table}

\paragraph{Epistemic Tokens} ~~ Following \cite{understanding}, we define a set of 10 epistemic markers $\mathcal{T} = \{\textit{wait},\, \textit{hmm},\, \textit{perhaps},\, \textit{maybe},\, \textit{actually},\, \textit{alternatively},\, \textit{seems},\, \textit{might},\, \textit{likely},\, \textit{check}\}$ as practical indicators of regions where the externalization of uncertainty is likely to occur. These tokens are intuitively recognizable as uncertainty expressions, with their
selection further supported by a cross-family analysis of 9,600 reasoning traces
(Appendix~\ref{sec:epistemic-token-selection}). We measure the epistemic token count of a response $y$ as $E(y) = \sum_{t \in \mathcal{T}} \mathrm{count}(t,\, y)$.

\paragraph{Results} ~~ We analyze how different forms of conditioning information toward the correct answer affect the model's reasoning behavior by comparing the average response length $\mathbb{E}[L(y)]$, score, and the epistemic token count $\mathbb{E}[E(y)]$ across the four settings. As shown in Table~\ref{tab:reasoning_behavior}, both quantities decrease monotonically as $I(y^*;\,c \mid x)$ increases:
\begin{equation}
\mathbb{E}\bigl[L(y)\bigr]\Big|_{(1)}
\;>\;
\mathbb{E}\bigl[L(y)\bigr]\Big|_{(3)}
\;>\;
\mathbb{E}\bigl[L(y)\bigr]\Big|_{(4)}
\;>\;
\mathbb{E}\bigl[L(y)\bigr]\Big|_{(2)},
\end{equation}
and analogously for $\mathbb{E}[E(y)]$, confirming that richer conditioning information leads to more concise and confident reasoning.

\begin{table}[h!]
    \centering
    \small
    \renewcommand{\arraystretch}{1.3}
    \caption{Comparison of response characteristics under varying levels of rich information.}
    \label{tab:reasoning_behavior}
    \begin{tabular}{p{4.9cm}|P{1.85cm}P{2.15cm}P{3.2cm}}
        \hline
        & Avg. Score & Avg. Length & Epistemic Token Count \\
        \hline
        (1) Unguided & 0.30 & 13,054 & 182.5 \\
        \rowcolor{c-pink-light} (2) Solution-Guided ($c=s$) & 0.98 & 1,873 & 8.8 \\
        (3) Solution-Guided ($c = s_{\setminus\text{think}}$) & 0.78 & 12,036 & 159.8 \\
        \rowcolor{c-pink-light} (4) Regeneration-Conditioned & 0.95 & 2,808 & 24.1 \\
        \hline
    \end{tabular}
\end{table}

Specifically, unguided generation ($c = \emptyset$) produces substantially longer responses with the highest epistemic token counts. When the full solution $s$ is provided in~(2), the model follows the given reasoning trace with high confidence, and its concise output can be viewed as a compressed representation of the reasoning in $s$.
In~(3), removing the \texttt{<think>} portion retains only $s_{\setminus\text{think}}$ (640 out of 13{,}054 response tokens), and both $\mathbb{E}[L(y)]$ and $\mathbb{E}[E(y)]$ increase again toward the unguided level, reflecting the substantial information loss.
Setting~(4), conditioning on the regenerated response $y_r$, yields intermediate values (lower than~(3) but higher than~(2)), indicating that $y_r$ preserves much of the informative structure of the full solution.
Detailed per-token breakdowns are reported in Appendix~\ref{appendix:more_epistemic_behavior}.

\begin{tcolorbox}[
  colback=white,
  colframe=c-green-dark,
  enhanced,
  coltitle=black,
  colbacktitle=c-green-light,
  title={\small{\textcolor{c-green-dark}{\textbf{Takeaway 1}}: Information Richness and Epistemic Verbalization}},
  boxsep=2.7pt,
  top=1pt,
  bottom=1pt,
  left=4pt,
  right=4pt
]
\small
As the conditioning context $c$ becomes more informative and directly useful, the LLM generates answers more confidently and with fewer expressions of epistemic uncertainty.
\end{tcolorbox}

\section{Supervised Finetuning with Self-Distillation} \label{sec:sft}

A natural follow-up question is whether the suppression of epistemic verbalization under high $I(y^*;\,c \mid x)$ is merely stylistic or has a tangible impact on reasoning capability. To test this, we conduct off-policy self-distillation (SFT) using DeepSeek-R1-Distill-Qwen-7B \citep{deepseek-r1} on two datasets, each containing 800 correct responses:
 
\begin{itemize}[itemsep=0pt, topsep=0pt, partopsep=0pt, leftmargin=*]
\item $\mathcal{D}_{\text{ug}}$: unguided responses ($c = \emptyset$), with high $\mathbb{E}[E(y)]$ and $\mathbb{E}[L(y)] \approx 12\text{k}$ tokens.
\item $\mathcal{D}_{\text{sg}}$: solution-guided responses ($c = s$), with low $\mathbb{E}[E(y)]$ and $\mathbb{E}[L(y)] \approx 2\text{k}$ tokens.
\end{itemize}
 
\noindent Both datasets consist entirely of correct trajectories. The key difference lies in the epistemic density of the training signal. We evaluate the resulting checkpoints across multiple math benchmarks (examples from each dataset are presented in Appendix~\ref{appendix:qualitative_analysis}).
 
\begin{table}[h!]
    \centering
    \small
    \renewcommand{\arraystretch}{1.3}
    \caption{Math benchmark performance of the base model DeepSeek-R1-Distill-Qwen-7B and its SFT checkpoints trained on unguided and solution-guided datasets.}
    \begin{tabular}{p{4.5cm}|P{1.8cm}P{1.8cm}P{1.8cm}P{1.8cm}}
        \hline
        DeepSeek-R1-Distill-Qwen-7B  & AIME24 & AIME25 & AMC23 & MATH500 \\
        \hline
        Base & \textbf{54.79} & 37.92 & \textbf{89.06} & \textbf{92.19} \\
        SFT on $\mathcal{D}_{\text{ug}}$ & 51.04 & \textbf{40.00} & 87.66 & 90.93 \\
        \rowcolor{c-pink-light}
        SFT on $\mathcal{D}_{\text{sg}}$ & 20.21 & 12.71 & 57.03 & 65.52 \\
        \hline
    \end{tabular}
    \label{tab:sft_performance}
\end{table}
 
As shown in Table~\ref{tab:sft_performance}, training on $\mathcal{D}_{\text{sg}}$ leads to substantial degradation across all benchmarks, despite the dataset consisting of correct answers, whereas training on $\mathcal{D}_{\text{ug}}$ produces no significant performance change.
This asymmetry arises because solution-guided responses are concise precisely due to the external context $s$. Using them as SFT targets forces the model to imitate a reasoning style that presupposes information unavailable at inference time, effectively suppressing the epistemic tokens that support autonomous exploration and error correction.
 
\begin{tcolorbox}[
  colback=white,
  colframe=c-green-dark,
  enhanced,
  coltitle=black,
  colbacktitle=c-green-light,
  title={\small{\textcolor{c-green-dark}{\textbf{Takeaway 2}}: Epistemic Suppression and Reasoning Performance}},
  boxsep=2.7pt,
  top=1pt,
  bottom=1pt,
  left=4pt,
  right=4pt
]
\small
Even when trained on correct trajectories, excessively suppressing epistemic verbalization can substantially degrade reasoning performance.
\end{tcolorbox}

\section{On-Policy Self-Distillation} \label{sec:opsd}

We now turn to on-policy self-distillation \citep{SDPO, zhao2026self, privileged}, where the model learns from dense reward signals provided by a self-teacher with access to the correct solution, based on its own responses. We compare GRPO with Reinforcement Learning via Self-Distillation (SDPO) \citep{SDPO} on the DAPO-Math-17k dataset \citep{dapo}, using Qwen3-8B \citep{qwen3} and DeepSeek-R1-Distill-Qwen-7B \citep{deepseek-r1}. Additional results for Qwen3-1.7B and Olmo-3-7B-Instruct \citep{olmo} are in Appendix \ref{appendix:olmo}. For each model, we track training score, response length, and OOD performance on AIME24 and AMC23. We fix the teacher to the initial policy, which performs better than a moving target (see Section~\ref{appendix:fixed_vs_moving_target} for a comparison).

The behavior of on-policy self-distillation depends on two factors: (i) the extent of epistemic verbalization already present in the base model, and (ii) the richness of the conditioning context $c$. To disentangle these, we compare GRPO and SDPO under two settings: $c = s$ (full solution) and $c = s_{\setminus\text{think}}$ (solution without \texttt{<think>} content).

\subsection{DeepSeek-R1-Distill-Qwen-7B}

\begin{figure}[h!]
    \centering
    \begin{subfigure}[b]{0.49\linewidth}
        \centering
        \includegraphics[width=\linewidth]{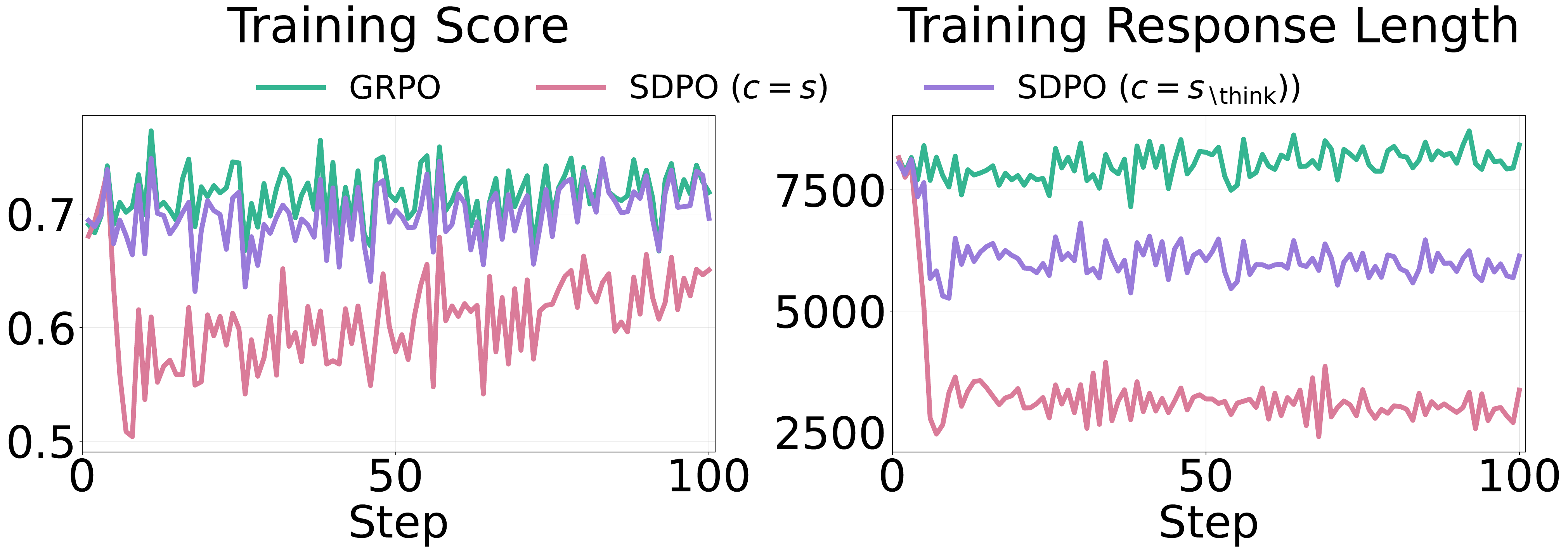}
        \caption{Training Score-Length Comparison}
        \label{fig:ds_score_length}
    \end{subfigure}
    \begin{subfigure}[b]{0.49\linewidth}
        \centering
        \includegraphics[width=\linewidth]{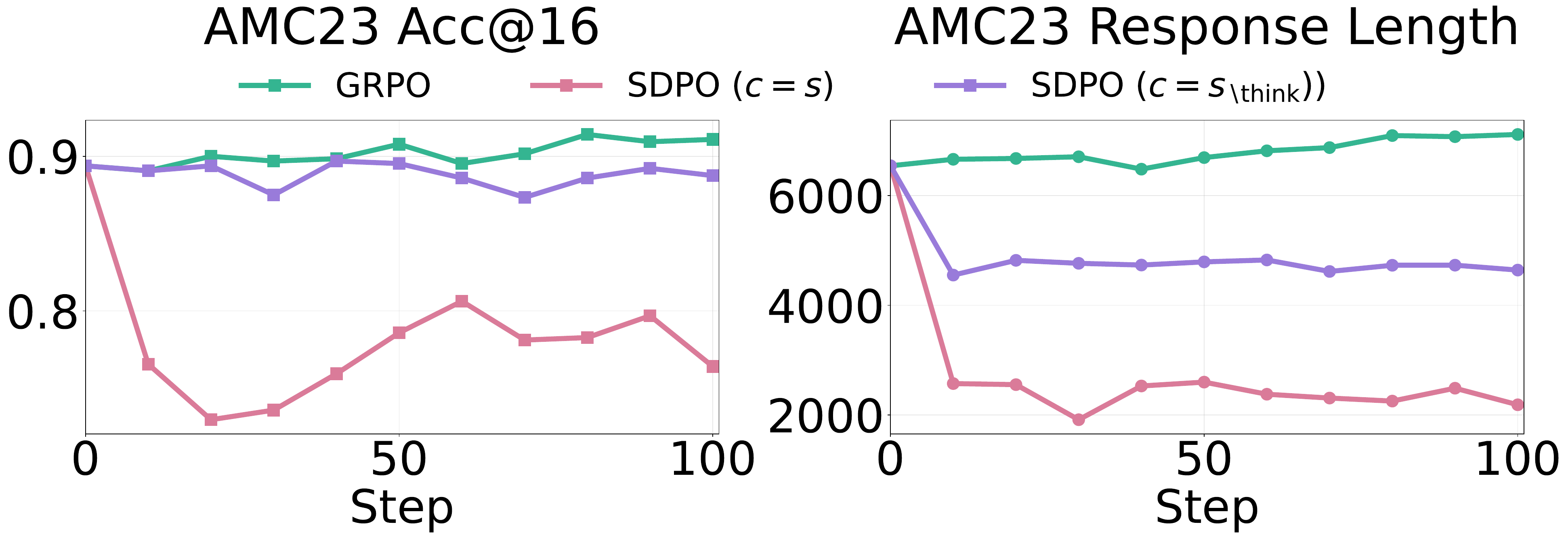}
        \caption{AMC23 Score and Response Length}
        \label{fig:ds_amc23}
    \end{subfigure} \\
    \vspace{0.3cm}
    \begin{subfigure}[b]{0.49\linewidth}
        \centering
        \includegraphics[width=\linewidth]{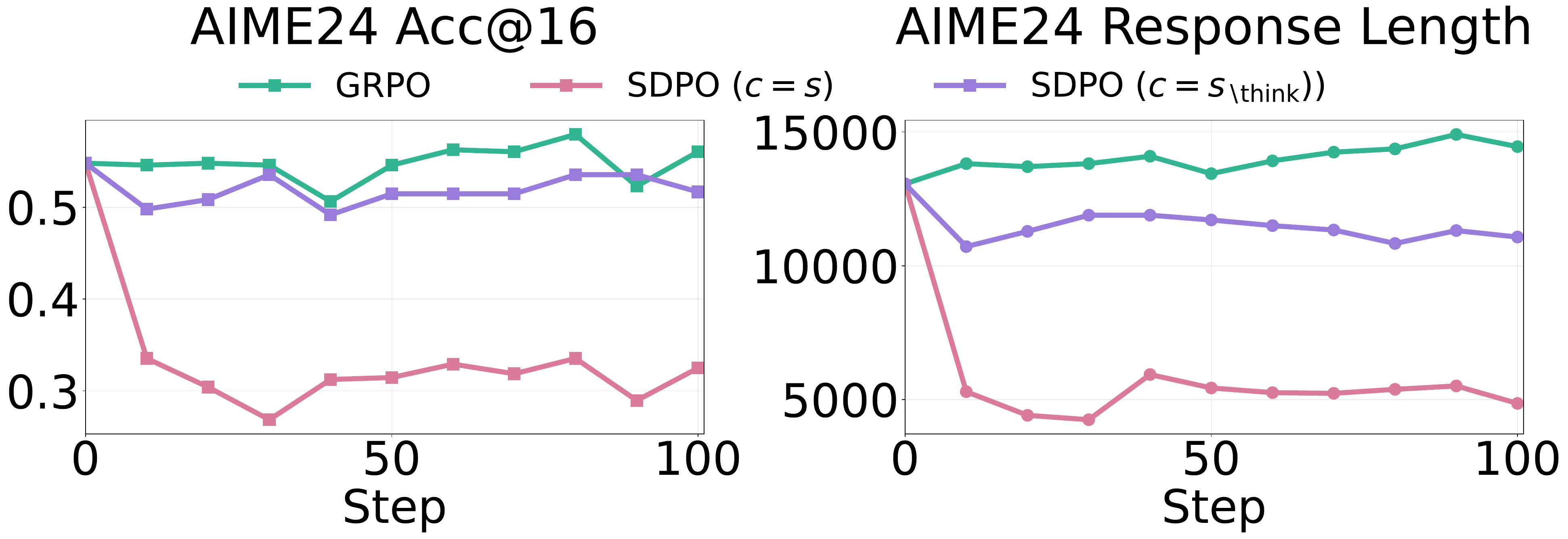}
        \caption{AIME24 Score and Response Length}
        \label{fig:ds_aime24}
    \end{subfigure}
    \begin{subfigure}[b]{0.49\linewidth}
        \centering
        \includegraphics[width=\linewidth]{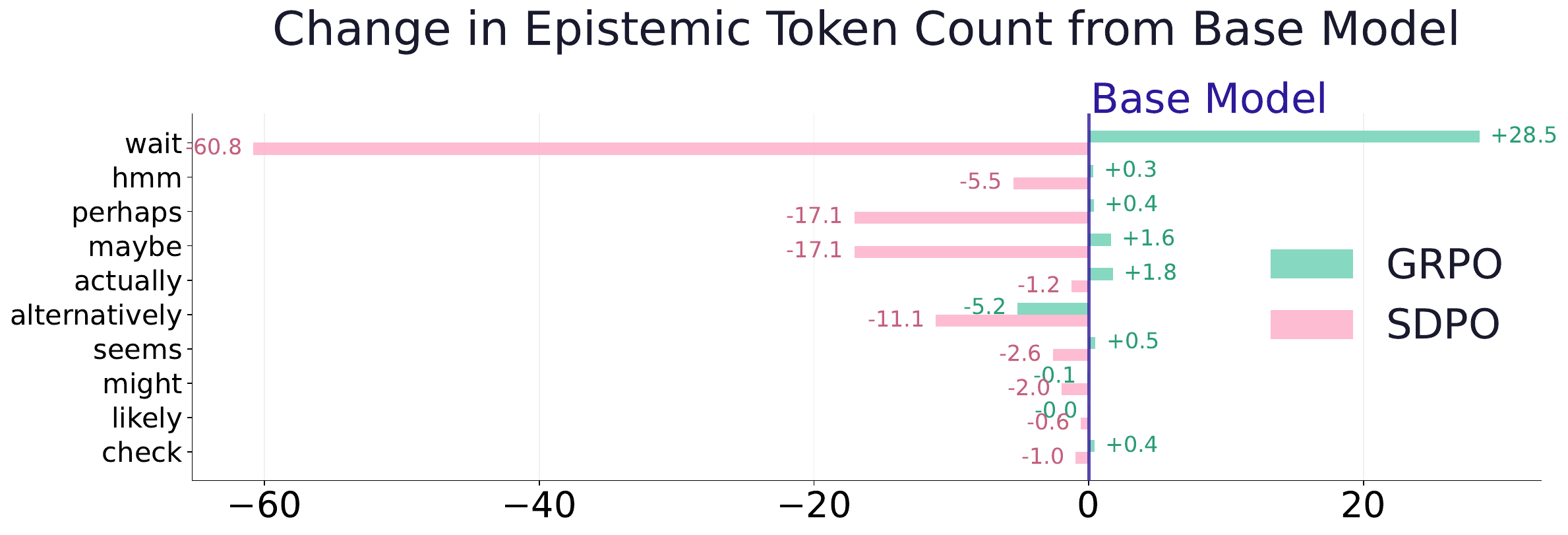}
        \caption{Change in Epistemic Token Usage on AIME24}
        \label{fig:ds_epistemic_change}
    \end{subfigure}
    \caption{On-policy self-distillation results for DeepSeek-R1-Distill-Qwen-7B. GRPO yields modest OOD gains with a slight increase in epistemic verbalization, whereas SDPO degrades both performance and epistemic token usage, particularly with $c = s$.}
\end{figure}

DeepSeek-R1-Distill-Qwen-7B is a representative high-reasoning model that generates rich epistemic verbalizations within \texttt{<think>} tags, averaging about 180 epistemic tokens per response, as shown in Section~\ref{sec:llm-reasoning-behavior}, leading to lengthy yet high-performing reasoning outputs.

\paragraph{Training Performance} ~~ As shown in Figure~\ref{fig:ds_score_length}, GRPO training slightly increases $\mathbb{E}[L(y)]$ with a modest improvement in score. In contrast, SDPO with $c = s$ causes a sharp initial drop in both $\mathbb{E}[L(y)]$ and score. Performance gradually recovers but remains below GRPO throughout training. When the conditioning is reduced to $c = s_{\setminus\text{think}}$, the drop in $\mathbb{E}[L(y)]$ is attenuated and the score trajectory approaches that of GRPO, consistent with the relationship between $I(y^*;\,c \mid x)$ and epistemic suppression discussed in Section~\ref{sec:llm-reasoning-behavior}.

\paragraph{OOD Evaluation - AIME24, AMC23} ~~ Consistent with the training trends, GRPO yields modest gains on both OOD benchmarks (AIME24: 54.7 $\to$ 56.0; AMC23: 89.3 $\to$ 91.1, Figures \ref{fig:ds_amc23} and \ref{fig:ds_aime24}) with a slight increase in $\mathbb{E}[L(y)]$. SDPO with $c = s$ degrades performance substantially (${\sim}40\%$ on AIME24, ${\sim}15\%$ on AMC23). Reducing the conditioning to $c = s_{\setminus\text{think}}$ mitigates the drop, though performance still remains below the base model.
 
\paragraph{Reasoning Pattern} ~~ Figure~\ref{fig:ds_epistemic_change} illustrates the epistemic token counts of the trained models. GRPO increases $\mathbb{E}[E(y)]$, whereas SDPO suppresses it more aggressively, consistent with the observed correlation between epistemic suppression and performance degradation throughout our analysis. Additional analyses on total versus epistemic token changes, as well as the number of uncertainty expressions per reasoning trace, evaluated using an LLM-as-Judge, are provided in Appendices~\ref{appendix:total_vs_epistemic} and \ref{appendix:epistemic_llm_as_judge}. Furthermore, a comparison of the KL divergence of epistemic tokens with respect to the mean is presented in Appendix~\ref{appendix:kl}. Example responses generated by GRPO and SDPO are included in Appendix~\ref{appendix:qualitative_grpo_sdpo}.

\subsection{Qwen3-8B (Thinking Mode: ON)} 

Qwen3-8B is a hybrid reasoning model that can operate in either a thinking mode for deliberate reasoning or a non-thinking mode for direct responses. With thinking mode enabled, Qwen3-8B produces longer responses than DeepSeek-R1-Distill-Qwen-7B, with a higher number of epistemic tokens (see Appendix~\ref{appendix:comparison_across_models} for a comparison of epistemic tokens across models).

\begin{figure}[h!]
    \centering
    \begin{subfigure}[b]{0.49\linewidth}
        \centering
        \includegraphics[width=\linewidth]{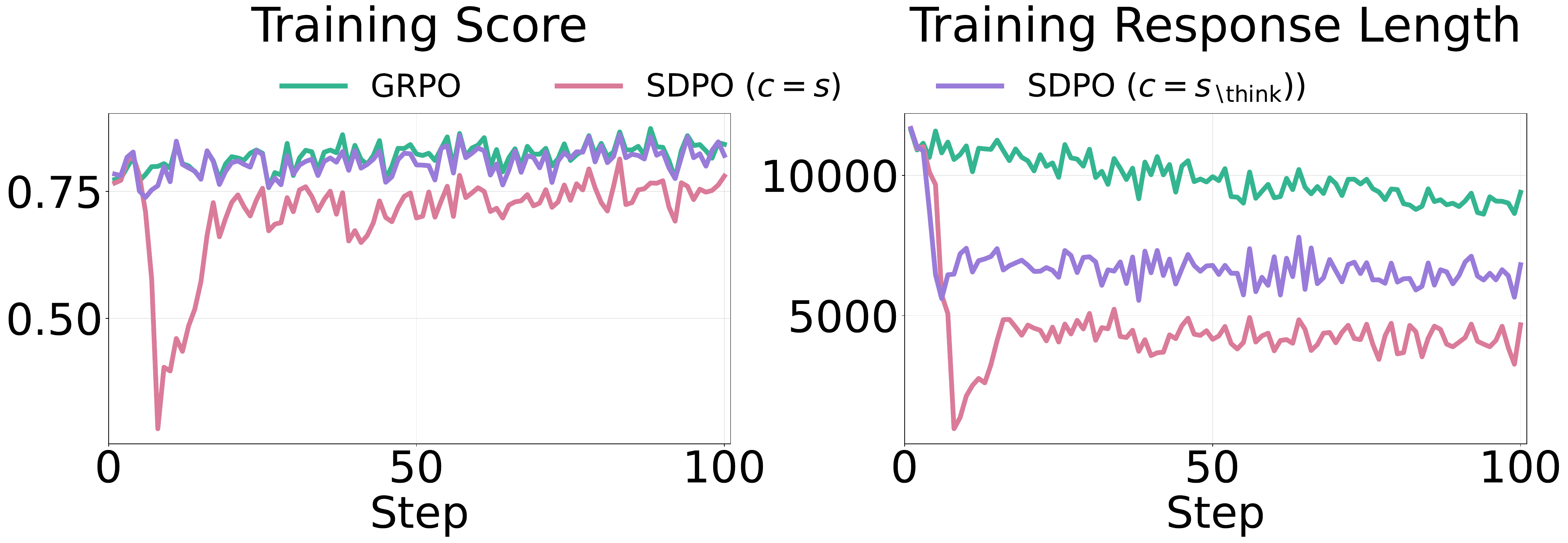}
        \caption{Training Score-Length Comparison}
        \label{fig:qwen3_8b_think_score_length}
    \end{subfigure}
    \begin{subfigure}[b]{0.49\linewidth}
        \centering
        \includegraphics[width=\linewidth]{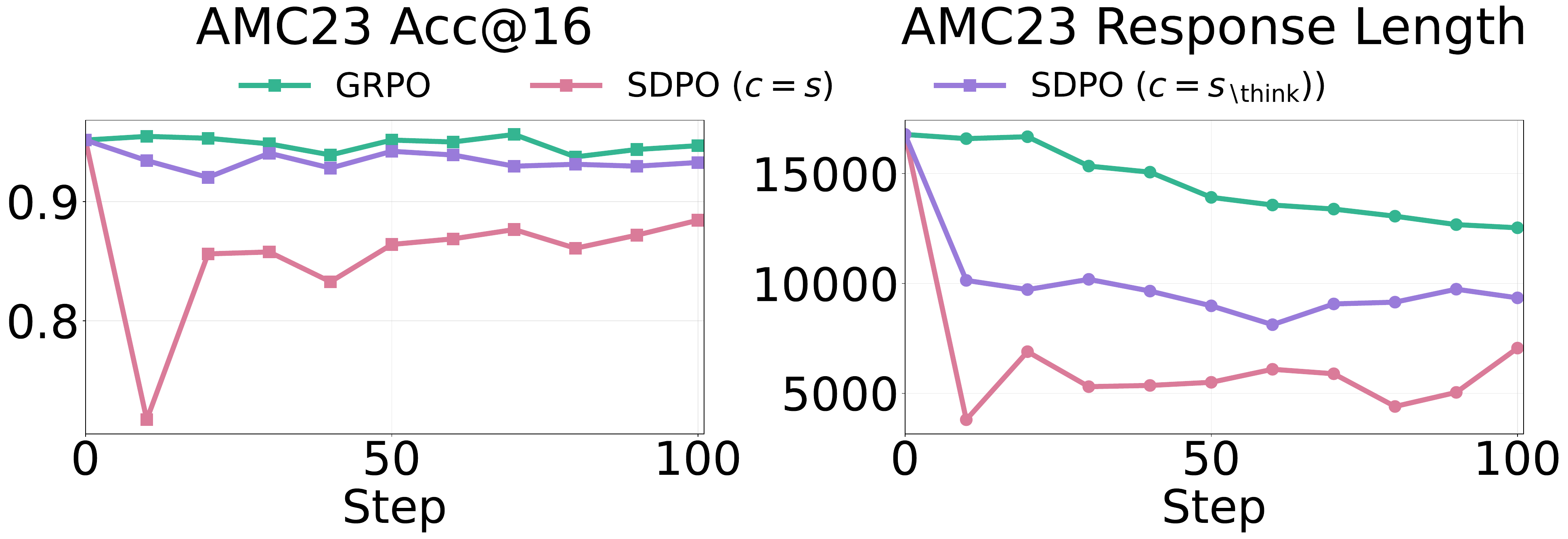}
        \caption{AMC23 Score and Response Length}
        \label{fig:qwen3_8b_think_amc23}
    \end{subfigure} \\
    \vspace{0.3cm}
    \begin{subfigure}[b]{0.49\linewidth}
        \centering
        \includegraphics[width=\linewidth]{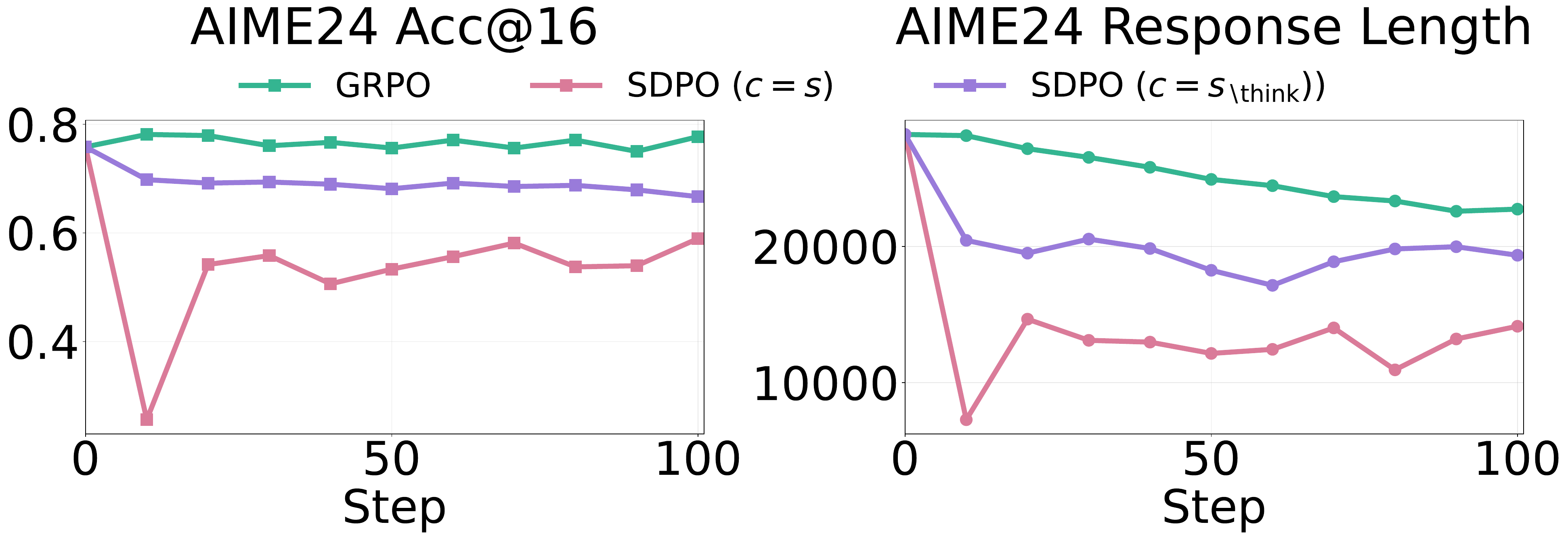}
        \caption{AIME24 Score and Response Length}
        \label{fig:qwen3_8b_think_aime24}
    \end{subfigure}
    \begin{subfigure}[b]{0.49\linewidth}
        \centering
        \includegraphics[width=\linewidth]{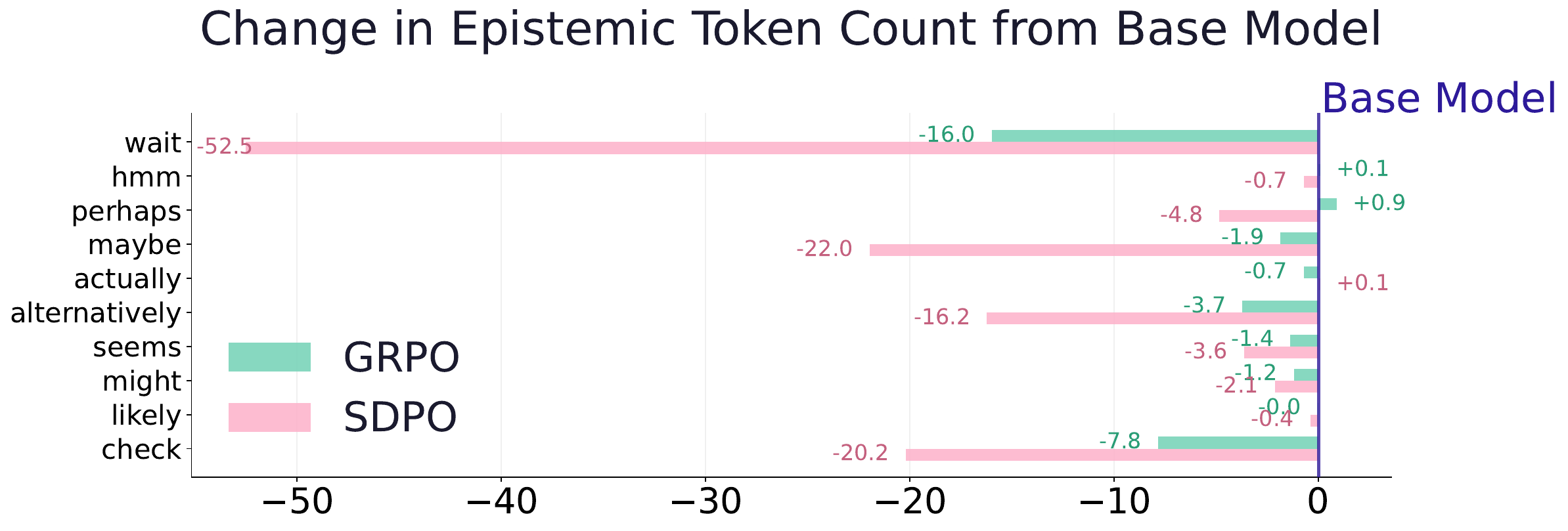}
        \caption{Change in Epistemic Token Usage on AIME24}
        \label{fig:qwen3_8b_think_epistemic_change}
    \end{subfigure}
    \caption{On-policy self-distillation results for Qwen3-8B (Thinking Mode: ON). Both GRPO and SDPO reduce response length and epistemic verbalization, but SDPO's more aggressive suppression leads to greater degradation in OOD performance, particularly on AIME24.}
\end{figure}

\paragraph{Training Performance} ~~ As shown in Figure~\ref{fig:qwen3_8b_think_score_length}, $\mathbb{E}[L(y)]$ decreases under both GRPO and SDPO, with SDPO exhibiting a larger reduction and a correspondingly larger performance drop. Notably, $\mathbb{E}[L(y)]$ first drops sharply, then increases slightly. Since the teacher policy is fixed as the reference policy, shortening the response by ${\sim}900$ tokens reduces the informativeness of $c$, i.e., decreases $I(y^*,c \mid x)$. As the context becomes less informative, the model compensates by increasing epistemic verbalization, causing the length to partially recover.
 
\paragraph{OOD Evaluation - AIME24, AMC23} ~~ The gap becomes more pronounced on OOD benchmarks (Figures~\ref{fig:qwen3_8b_think_amc23} and \ref{fig:qwen3_8b_think_aime24}): GRPO maintains largely stable performance with gradually decreasing $\mathbb{E}[L(y)]$, whereas SDPO falls below the base model, particularly with $c = s$. Notably, although GRPO and SDPO with $c = s_{\setminus\text{think}}$ achieve comparable training performance, their OOD results diverge, especially on the more challenging AIME24, where SDPO with $c = s_{\setminus\text{think}}$ shows progressive performance degradation as training proceeds.
 
\paragraph{Reasoning Pattern} ~~ Figure~\ref{fig:qwen3_8b_think_epistemic_change} reveals that both methods reduce $\mathbb{E}[E(y)]$ relative to the base model, with SDPO more aggressively so. This suggests that Qwen3-8B originally generates more epistemic verbalization than necessary. While both methods mitigate this redundancy, overly aggressive suppression risks removing epistemic signals that carry useful reasoning information.
 
\subsection{Qwen3-8B (Thinking Mode: OFF)} \label{sec:qwen3-8b-nothink}

\begin{figure}[h!]
    \centering
    \begin{subfigure}[b]{0.49\linewidth}
        \centering
        \includegraphics[width=\linewidth]{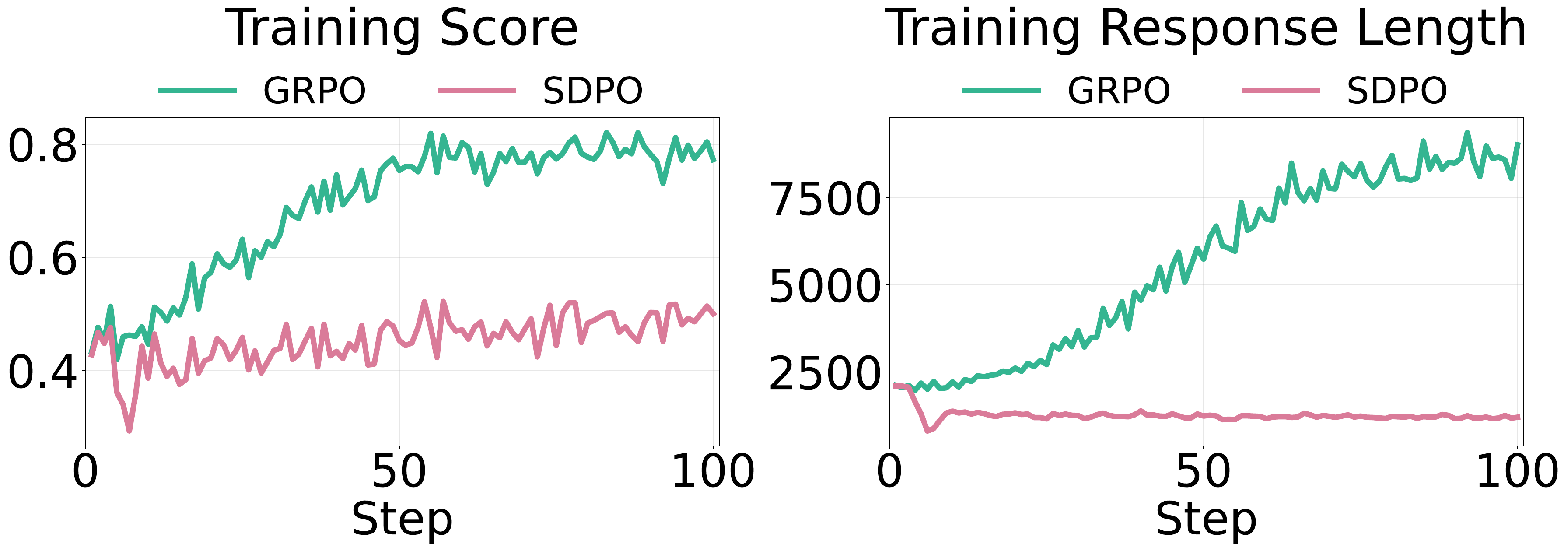}
        \caption{Training Score-Length Comparison}
        \label{fig:nothink_score_length}
    \end{subfigure}
    \begin{subfigure}[b]{0.49\linewidth}
        \centering
        \includegraphics[width=\linewidth]{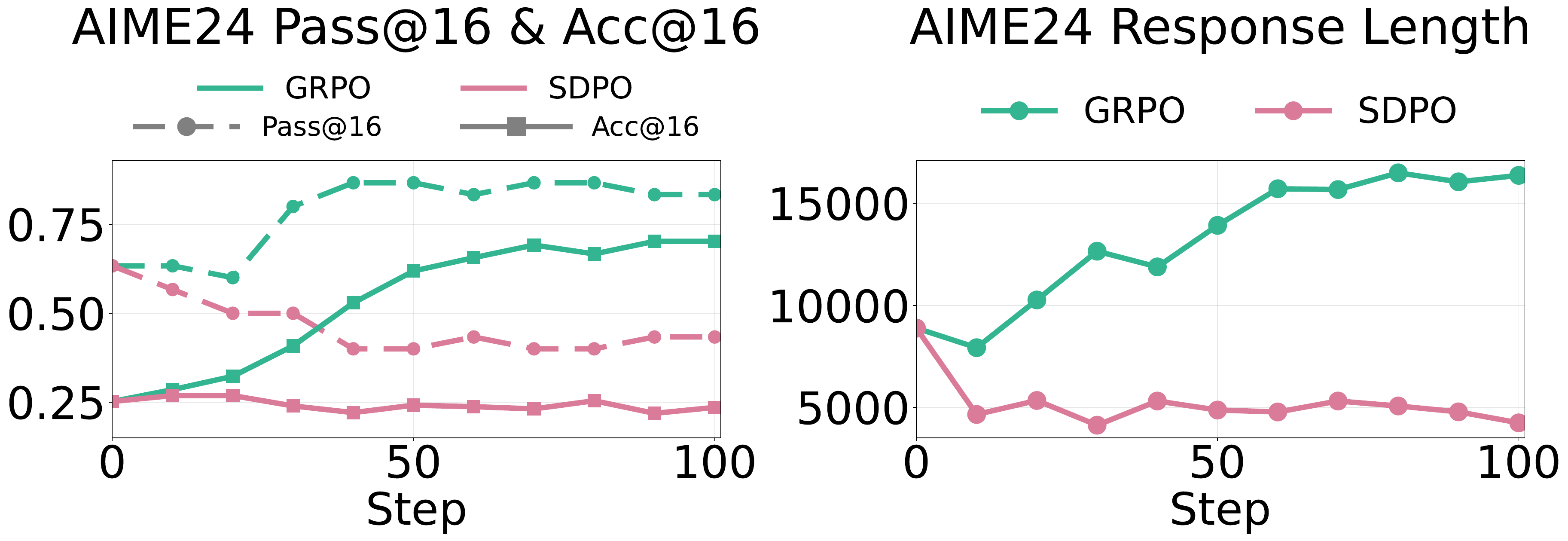}
        \caption{AIME24 Score and Response Length}
        \label{fig:nothink_aime24}
    \end{subfigure}
    \caption{On-policy self-distillation results for Qwen3-8B (Thinking Mode: OFF). GRPO rapidly increases response length via epistemic verbalization and achieves strong training gains, while SDPO reduces response length and struggles to improve, with slight OOD degradation on AIME24.}
    \label{fig:qwen3-8b-nothink}
\end{figure}

When Qwen3-8B is used without thinking mode, the \texttt{<think>} tag is absent, so we consider only $c = s$. The model initially produces much shorter responses and exhibits significantly lower performance. GRPO rapidly increases $\mathbb{E}[L(y)]$ by promoting epistemic verbalization (as shown in Appendix~\ref{appendix:more_on_policy_qwen3_8b_nothink}), quickly achieving a high training score. In contrast, SDPO reduces $\mathbb{E}[L(y)]$ and improves much more slowly. Even when the training score slightly increases, as shown in Figure~\ref{fig:nothink_aime24}, performance on AIME24 slightly declines ($0.25 \to 0.23$), further illustrating the cost of epistemic suppression under self-distillation.

\subsection{Causal Study: Epistemic Token Suppression during GRPO}

To further examine whether the suppression of epistemic verbalization is a
causal contributor to the observed performance degradation, we explicitly
penalized epistemic tokens during GRPO training and tested whether this
intervention induces a similar degradation pattern.

Following the epistemic marker set $\mathcal{T}$ defined in Section~\ref{sec:llm-reasoning-behavior},
we override the standard GRPO advantage at every position $t$ whose generated token realizes
a marker in $\mathcal{T}$, setting $\tilde{A}_{t} = \min(A_{t}, -\kappa)$ if $y_{t} \in \mathcal{T}$
and $\tilde{A}_{t} = A_{t}$ otherwise, with suppression strength $\kappa = 0.5$. Since
$\tilde{A}_{t} \le -\kappa$ holds even for correct rollouts, the policy gradient consistently
reduces $\mathbb{E}[E(y)]$ throughout training.


\begin{figure}[h!]
    \centering
    \begin{subfigure}[b]{0.49\linewidth}
        \centering
        \includegraphics[width=\linewidth]{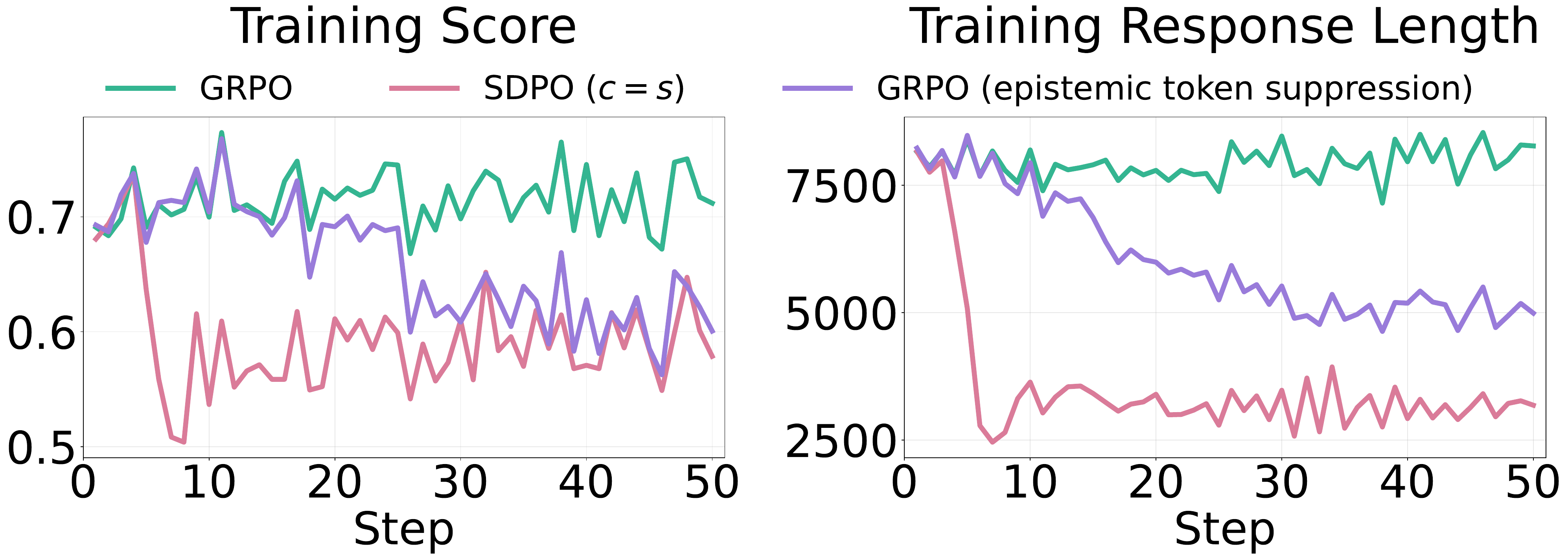}
        \caption{DeepSeek-Distill-7B Training Comparison}
        \label{fig:grpo_epistemic_score_length}
    \end{subfigure}
    \begin{subfigure}[b]{0.49\linewidth}
        \centering
        \includegraphics[width=\linewidth]{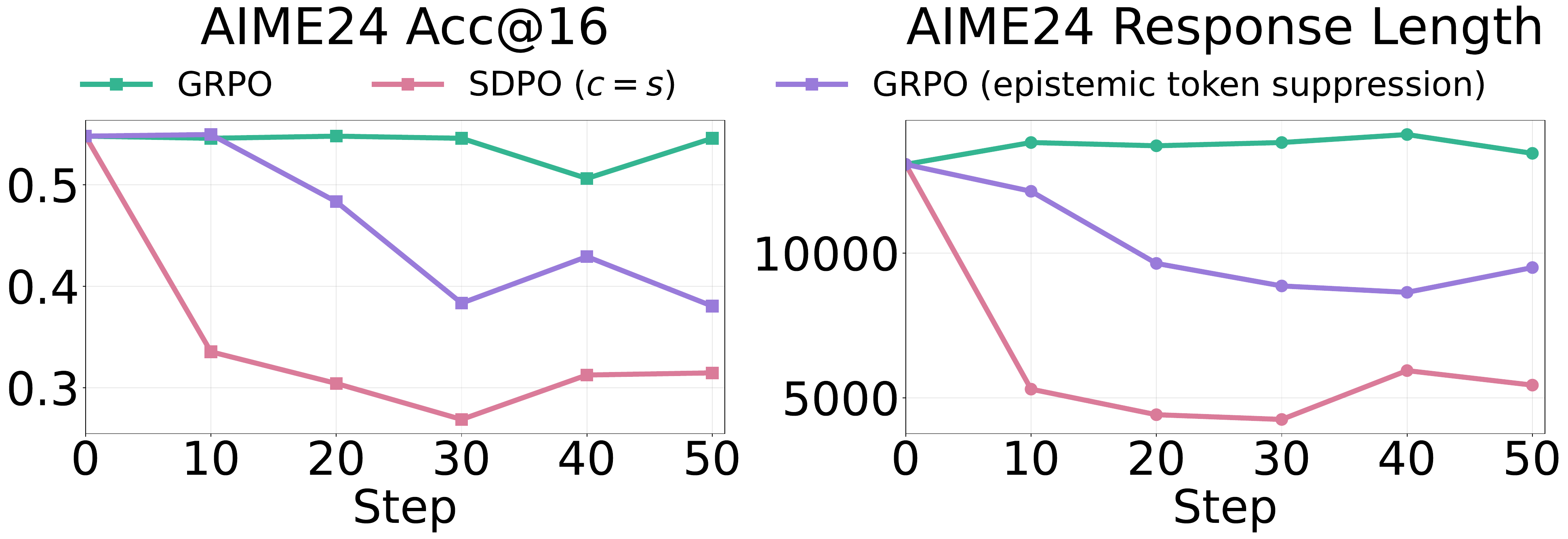}
        \caption{DeepSeek-Distill-7B AIME24 Comparison}
        \label{fig:grpo_epistemic_aime24}
    \end{subfigure}
    \caption{Effect of epistemic-token suppression during GRPO for DeepSeek-Distill-7B.} \label{fig:grpo_epistemic_suppression}
    
    \end{figure} 
    
As shown in Figure~\ref{fig:grpo_epistemic_suppression}, explicitly penalizing epistemic tokens during GRPO produces shorter responses and lower performance, qualitatively reproducing the degradation pattern observed with SDPO. The degradation is less pronounced than under SDPO, possibly because the predefined epistemic token set does not capture all linguistic realizations of uncertainty or epistemic reasoning, as discussed further in Appendix~\ref{appendix:epistemic_llm_as_judge}. Nevertheless, this result provides causal evidence that suppressing uncertainty expression contributes to the performance decline under SDPO.


\subsection{Ablation Study: Fixed vs.\ Moving Target Teacher} \label{appendix:fixed_vs_moving_target}

In naive on-policy self-distillation, the teacher and student share a continuously updated policy, making the teacher a moving target that can introduce training instability \citep{zhao2026self, opcd}. To mitigate this, SDPO uses an EMA-smoothed teacher (EMA rate: 0.05). However, we find that setting the EMA rate to 0.0 (i.e., fixing the teacher to the initial policy) yields better performance, and thus Section~\ref{sec:opsd} follows this setting. 

\begin{figure}[h!]
    \centering
    \begin{subfigure}[b]{0.49\linewidth}
        \centering
        \includegraphics[width=\linewidth]{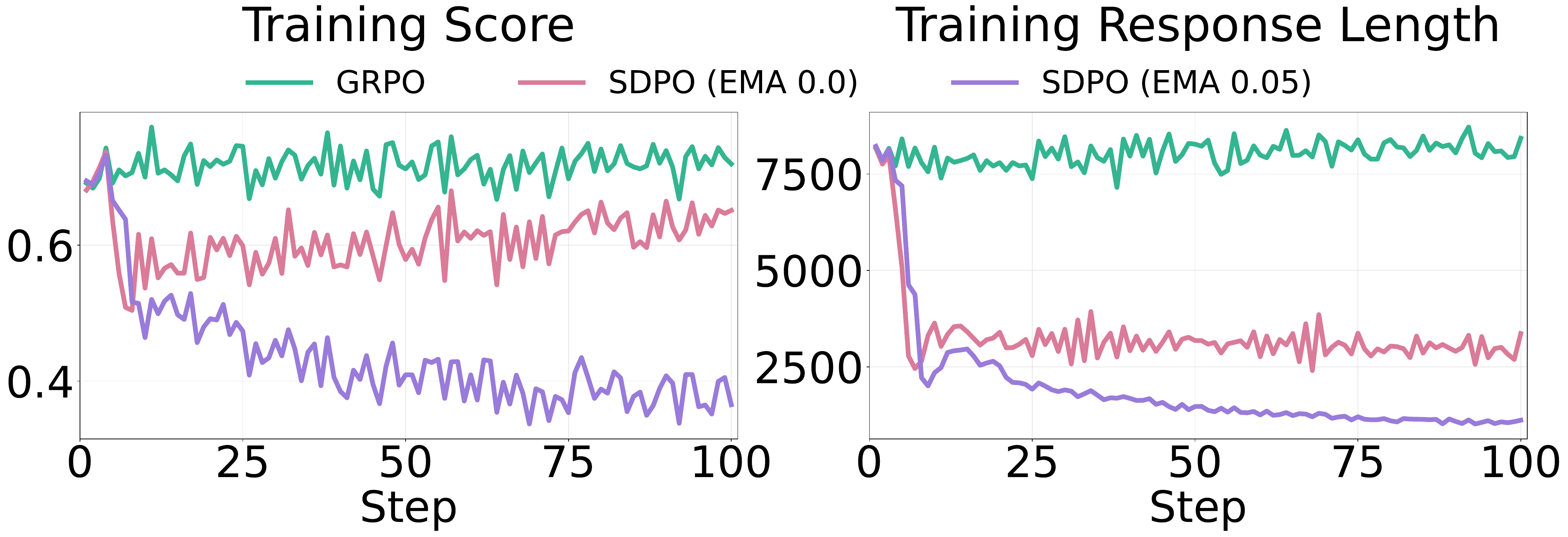}
        \caption{DeepSeek-Distill-7B Training Comparison}
        \label{fig:deepseek_ema_score_length}
    \end{subfigure}
    \begin{subfigure}[b]{0.49\linewidth}
        \centering
        \includegraphics[width=\linewidth]{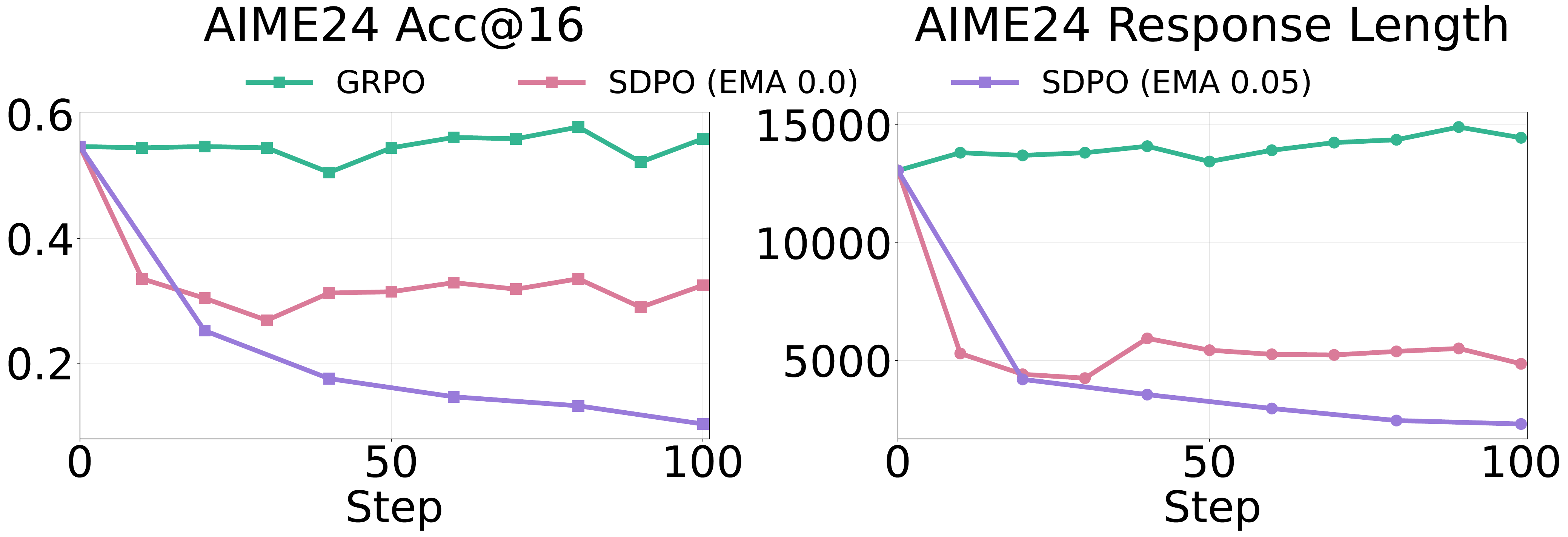}
        \caption{DeepSeek-Distill-7B AIME24 Comparison}
        \label{fig:deepseek_ema_aime24}
    \end{subfigure}
    \caption{Fixed vs.\ moving target teacher for DeepSeek-R1-Distill-Qwen-7B. Even slow EMA updates (rate 0.05) amplify epistemic suppression via a feedback loop, causing greater performance degradation than a fixed teacher.}
\end{figure}

Figure~\ref{fig:deepseek_ema_score_length} shows additional comparison results when the teacher is updated during training. As shown, even slow updates (e.g., rate 0.05) lead to a sharper reduction in response length, resulting in larger performance degradation. This can be interpreted as a feedback loop in self-distillation. The model is trained to produce increasingly confident outputs, and when this is used as the teacher, it generates even more confident responses, amplifying the effect over iterations. Further ablations on learning rate, top-$k$ logits, and the effect of limiting the maximum response length to 4,096 tokens are provided in Appendix~\ref{appendix:more_ablation}.

\begin{tcolorbox}[
  colback=white,
  colframe=c-green-dark,
  enhanced,
  coltitle=black,
  colbacktitle=c-green-light,
  title={\small{\textcolor{c-green-dark}{\textbf{Takeaway 3}}: Epistemic Verbalization Changes and Performance in On-Policy Self-Distillation}},
  boxsep=2.7pt,
  top=1pt,
  bottom=1pt,
  left=4pt,
  right=4pt
]
As the teacher context $c$ becomes more information-rich, on-policy self-distillation reduces epistemic verbalization and shortens responses. This effect varies depending on the level of uncertainty originally expressed by the base model.
\end{tcolorbox}

\section{Relationship Between Task Coverage, Epistemic Verbalization and Generalization Ability} \label{sec:generalization}

Across the off-policy and on-policy settings analyzed above, self-distillation consistently produces more confident responses with reduced $\mathbb{E}[E(y)]$. This aligns with the findings of \cite{SDPO}, which reports that SDPO learns to reason concisely: on Science Q\&A (Chemistry, Physics, Biology, and Materials Science) \citep{sciknoweval}, tool use \citep{toolalpaca}, and LiveCodeBench v6 \citep{livecodebench}, SDPO achieves higher accuracy than GRPO while producing substantially shorter outputs with fewer epistemic markers.
 
In other words, in these domains, self-distillation suppresses epistemic verbalization and improves performance simultaneously. The key question is \emph{why the same mechanism leads to performance degradation in our math-focused setup}. We hypothesize that the answer lies in differences in task coverage between the training and evaluation distributions.
 
\subsection{Comparison of Task Coverage}
 
To test this hypothesis, we compare the dataset characteristics of the settings where SDPO outperformed GRPO against our experimental setup. As shown in Table~\ref{tab:task_coverage}, the Chemistry dataset, despite its large size, draws from only six main problem types that differ primarily in surface details rather than underlying structure. LiveCodeBench v6 contains diverse problems but only 131 in total, leading to repeated exposure during training with identical train/eval splits. In contrast, DAPO-Math-17k exposes the model to ~14,000 distinct problems (78\% of the 25,600 samples drawn over 100 steps, due to repeated sampling), spanning a broad, non-overlapping range of problem types, and evaluation is performed on unseen problem types.

\begin{table}[h!]
    \centering
    \scriptsize
    \renewcommand{\arraystretch}{1.3}
    \caption{Comparison of the total number of problems, problem composition, and train–evaluation splits in ScienceQ\&A, LiveCodeBench v6, and our experiments.}
    \label{tab:task_coverage}
    \begin{tabular}{p{2cm}|p{11cm}}
        \hline
        \textbf{Domain} & \textbf{Dataset Analysis} \\
        \hline
        ScienceQ\&A (Chemistry) & 
        Total 2,400 questions: 
        Reaction Balancing (300), 
        Molecular Descriptor Counting (300), 
        Molecular Weight Calculation (600), 
        Property Prediction (e.g., logS) (500), 
        Precursor/Reactant Selection (Retrosynthesis) (300), 
        Product Prediction (Organic Reactions) (400). 
        \newline
        \textbf{\textit{Train/Eval split:}} 90\% for training, 10\% for evaluation. \\
        \hline
        LiveCodeBench v6 & 
        \textbf{\textit{Train/Eval split:}} Total 131 questions for training and all 131 questions are used for both training and evaluation. Only 50\% of the public test cases are used during training, while the full set including hidden test cases is used for evaluation. \\
        \hline
        DAPO-Math-17k & \textbf{\textit{Train/Eval split:}} Total ~14,000 questions for training and evaluation is conducted on standard benchmarks (AIME24/25, AMC23, MATH 500) using questions not part of the training data. \\
        \hline
    \end{tabular}
\end{table}

\subsection{Relationship Between Task Coverage and Learning Performance} \label{sec:task_coverage_performance}
To further investigate the interplay between task coverage and generalization, we vary the number of training questions $|\mathcal{D}| \in \{1, 8, 64, 128, 512\}$ from DAPO-Math-17k and train with both GRPO and SDPO. All experiments use Qwen3-8B (Thinking Mode OFF).

\begin{figure}[h!]
    \centering
    \includegraphics[width=0.49\linewidth]{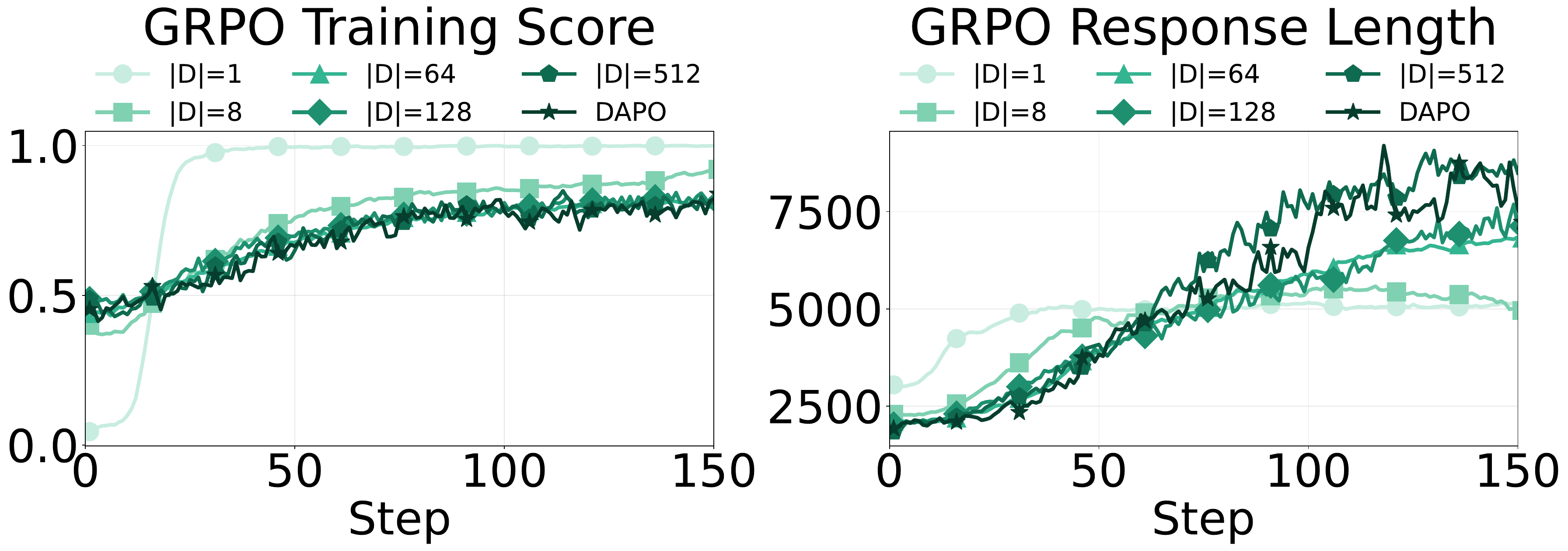}
    \includegraphics[width=0.49\linewidth]{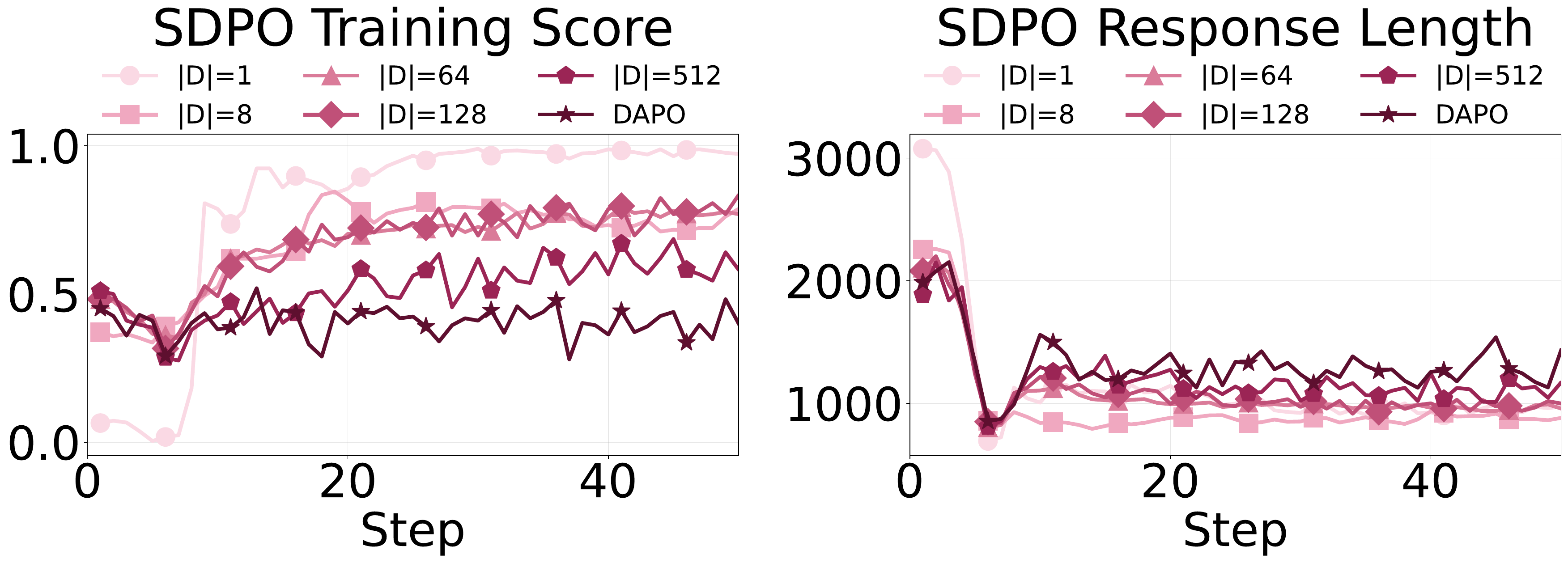}
    \caption{Training score and response length comparison between GRPO and SDPO for $|D| \in \{1, 8, 64, 128, 512\}$.}
    \label{fig:generalization_training}
\end{figure}

\paragraph{Training Logs} ~~ As shown in Figure \ref{fig:generalization_training}, GRPO and SDPO exhibit distinct training dynamics as $|\mathcal{D}|$ varies. When $|\mathcal{D}| \leq 128$, SDPO quickly achieves high scores while reducing $\mathbb{E}[L(y)]$ by up to $8\times$, indicating higher training efficiency on a small task set. However, at $|\mathcal{D}| = 512$, further reductions in $\mathbb{E}[L(y)]$ begin to hurt the training score relative to GRPO, whose $\mathbb{E}[L(y)]$ gradually increases with $|\mathcal{D}|$.

This difference can be interpreted through \textbf{task coverage}. As $|\mathcal{D}|$ grows, the model must accommodate a broader range of reasoning patterns. GRPO addresses this by increasing $\mathbb{E}[E(y)]$, allowing the model to express greater uncertainty and adapt its reasoning accordingly. SDPO instead encourages confident, concise responses---\textbf{effective when task coverage is small but limiting when the problem set becomes larger and more diverse.}

\begin{figure}[h!]
    \centering
    \includegraphics[width=0.49\linewidth]{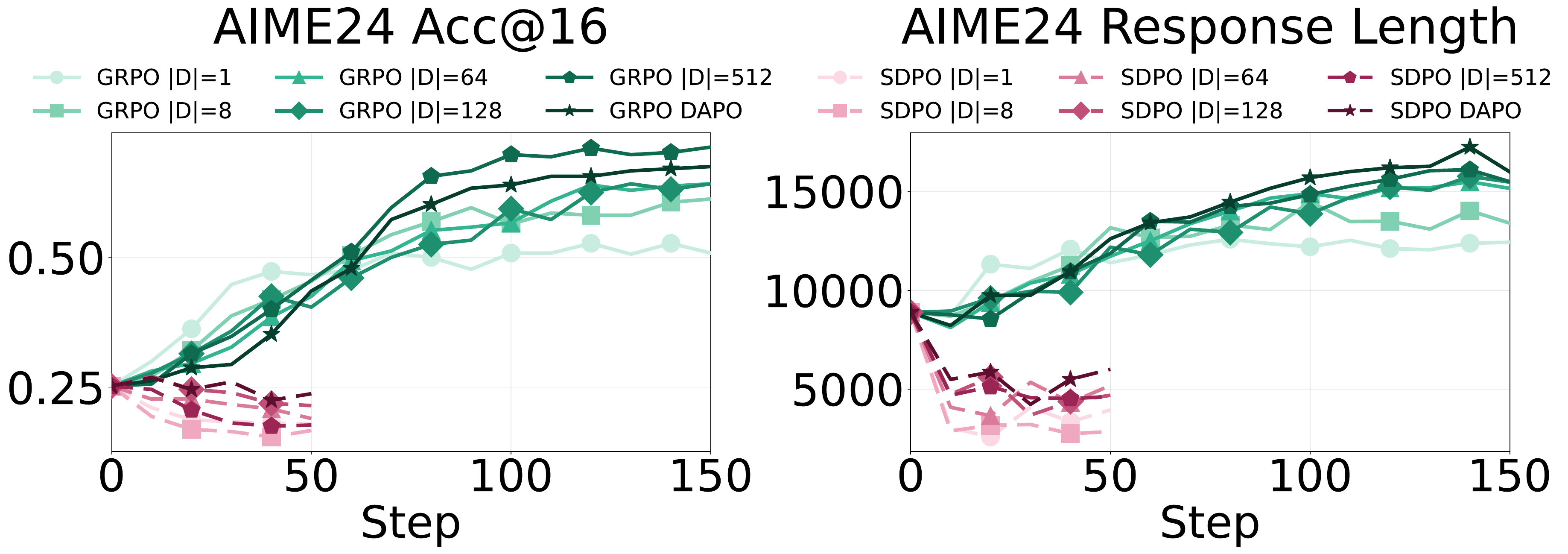}
    \includegraphics[width=0.49\linewidth]{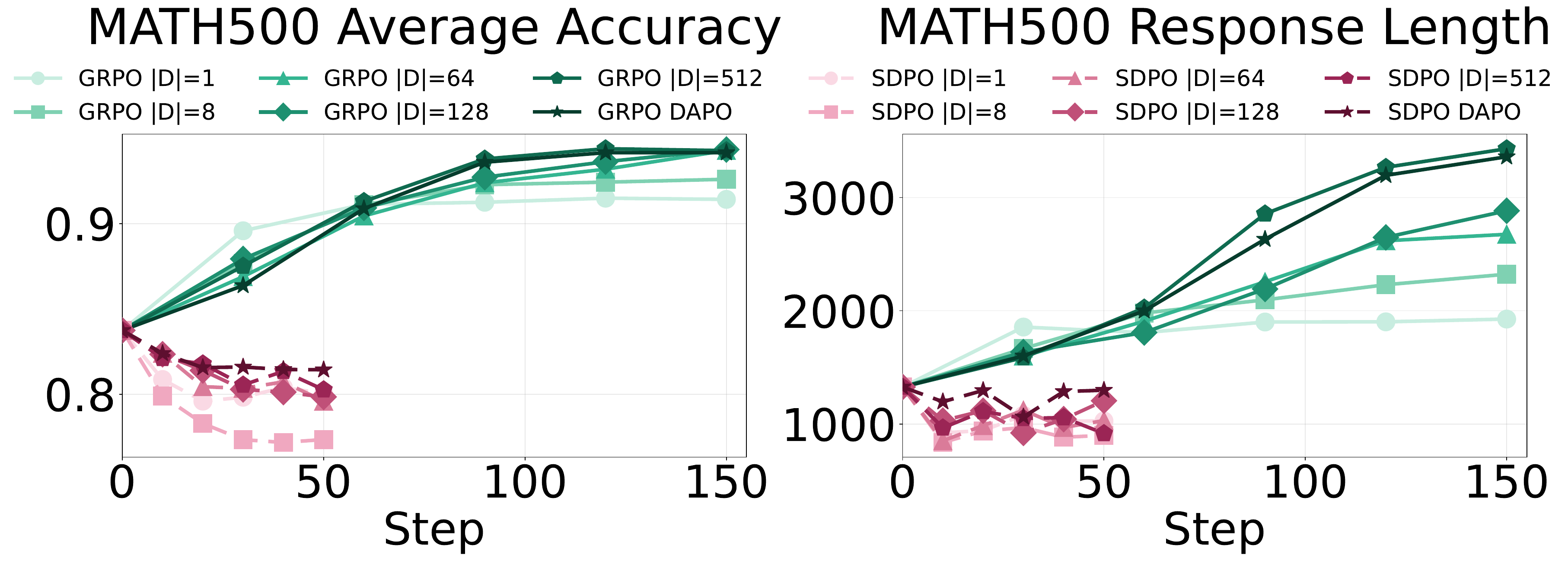}
    \caption{Evaluation performance on AIME24 and MATH500 and response length as $|D|$ varies over \{1, 8, 64, 128, 512\}.}
    \label{fig:generalization_ood_eval}
\end{figure}
 
\paragraph{OOD Evaluation - AIME24, MATH500} ~~ The distinction between GRPO and SDPO becomes more pronounced on OOD benchmarks (Figure~\ref{fig:generalization_ood_eval}). Under GRPO, performance scales consistently with $|\mathcal{D}|$: $|\mathcal{D}|=1$ converges quickly but soon stops improving, while larger $|\mathcal{D}|$ yields progressively higher final scores accompanied by increasing $\mathbb{E}[L(y)]$. Under SDPO, the pattern reverses: smaller $|\mathcal{D}|$ leads to more severe OOD degradation. Even at the largest $|\mathcal{D}|$ (DAPO setting), SDPO still underperforms the base model. Changes in GRPO and SDPO reasoning patterns across task coverage are provided in Appendix~\ref{appendix:more_epistemic_generalization}.

\begin{tcolorbox}[
  colback=white,
  colframe=c-green-dark,
  enhanced,
  coltitle=black,
  colbacktitle=c-green-light,
  title={\small{\textcolor{c-green-dark}{\textbf{Takeaway 4}}: Epistemic Verbalization Depends on Task Generalization}},
  boxsep=2.7pt,
  top=1pt,
  bottom=1pt,
  left=4pt,
  right=4pt
]
The value of epistemic verbalization scales with generalization demands: largely redundant for familiar, repetitive tasks (small $|\mathcal{D}|$) and removable for efficiency, but increasingly important as task diversity grows.
\end{tcolorbox}

\section{Conclusion}

In this work, we provide an empirical analysis of on-policy self-distillation, motivated by an information-theoretic view of conditioning context richness. Our experiments suggest that the effectiveness of self-distillation is closely tied to how information is provided to the model and how the model expresses uncertainty during reasoning. We observe that self-distillation tends to produce answers with higher confidence and shorter reasoning traces. While this effect enables more compact reasoning and can quickly improve in-domain performance when task coverage is limited, it becomes less effective when task coverage is broad and may even harm OOD performance.

Importantly, the fact that performance can degrade even when a mathematically sound objective function is designed to elicit correct CoT reasoning suggests that the choice of optimization objective alone may not be sufficient for preserving robust reasoning, and that we need to pay closer attention to how training reshapes the model's reasoning behavior, beyond answer correctness. We hope that our analysis contributes to a deeper understanding of self-distillation and other LLM post-training methods, and provides insights for developing more robust training strategies.


\section*{Acknowledgments}
We thank Li Dong, Tianzhu Ye, and Sojeong Rhee for their valuable discussions. This work was supported by Microsoft Research and partly by the National Research Foundation of Korea(NRF) grant funded by the Korea government(MSIT)(RS-2025-00557589).

\section*{Ethics Statement}

This work aims to analyze the causes of reasoning performance degradation during self-distillation of LLMs. All experiments were conducted using publicly available models (Qwen3, DeepSeek-R1-Distill-Qwen, OLMo) and datasets (DAPO-Math-17k, AIME, AMC, MATH500). As this study does not involve any new data collection or human subjects research, concerns related to privacy or human subjects ethics do not apply.

A key finding of this work is that self-distillation can suppress epistemic verbalization, thereby diminishing a model's ability to express uncertainty during reasoning. This observation carries important implications for the design of LLM post-training methods. The ability to appropriately express uncertainty is directly related to model reliability and safety. Training procedures that indiscriminately suppress such expressions may increase the risk of models producing incorrect answers with unwarranted confidence. We hope that our analysis contributes to the development of more robust and trustworthy reasoning models.

\section*{Reproducibility Statement}

We release our implementation and experiment scripts at
\href{https://github.com/beanie00/self-distillation-analysis}
{GitHub (beanie00/self-distillation-analysis)},
along with our training logs on
\href{https://wandb.ai/beanie/SDPO-beanie/reports/Why-Does-Self-Distillation-Sometimes-Degrade-the-Reasoning-Capability-of-LLMs---VmlldzoxNjI1MTk5Mw?accessToken=i1ohtk0lbrcm63kkv1vd99f6jvb71sqoj32rgsokrs4l8r4vjz8q5mywtt8upwho}
{Weights \& Biases},
and trained checkpoints for the main model settings on
\href{https://huggingface.co/collections/beanie00/self-distillation-analysis}
{Hugging Face (beanie00/self-distillation-analysis)}.

The repository provides instructions and scripts for data preparation,
GRPO/SDPO training, task-coverage experiments, and evaluation.
Detailed prompts, training hyperparameters, and evaluation configurations
are provided in Appendix~\ref{appendix:experimental_details}.

\bibliography{colm2026_conference}
\bibliographystyle{colm2026_conference}

\appendix

\newpage
\section{Related Works} 

\paragraph{Self-Distillation for LLM Reasoning.}
Self-distillation, where a single model acts as both teacher and student under asymmetric conditioning~\citep{self-distillation}, has recently been widely adopted for LLM post-training.
SDPO~\citep{SDPO} conditions the self-teacher on its previously generated correct trajectories or rich environment feedback to provide dense credit assignment, while OPSD~\citep{zhao2026self} leverages ground-truth solutions as privileged information, achieving strong efficiency gains in both cases.
Our analysis builds upon the SDPO codebase, and a detailed comparison with OPSD is provided in Appendix~\ref{appendix:comparison_opsd}. \citet{empo2, privileged, opcd} further utilize this paradigm in settings that incorporate additional signals, such as privileged information, self-generated hints derived from environmental interactions, or contextual guidance, with the goal of internalizing this external knowledge into the model’s parameters.

While these works demonstrate the effectiveness of self-distillation, they do not investigate when or why self-distillation may degrade reasoning performance, especially in scenarios where a model must solve problems entirely on its own, without interaction with the external environment, which is the central question of our work.

\paragraph{Uncertainty Expression in LLM Reasoning.} \citet{understanding} introduce an information-theoretic framework that identifies epistemic verbalization, the explicit externalization of uncertainty, as informationally necessary for robust reasoning. Our work answers a question this framework cannot readily explain: \emph{Under what conditions is this behavior encouraged or suppressed? And if epistemic verbalization is necessary, why does suppressing it lead to better performance in some domains, such as Chemistry, while degrading performance in others?} We trace this to two factors: \textbf{conditioning context richness} (Section~\ref{sec:llm-reasoning-behavior}) and \textbf{task coverage} (Section~\ref{sec:generalization}), supported by a systematic analysis of recent on-policy self-distillation methods across various models with varying pretrained capabilities.

\paragraph{Reasoning Compression.}
Several recent methods aim to shorten reasoning traces while preserving accuracy.
GFPO~\citep{sample-more} filters training rollouts by length and token efficiency within enlarged sample groups, reducing GRPO's length inflation by 46--85\% on STEM and coding benchmarks.
OPSDC~\citep{OPSDC} uses a conciseness instruction as privileged context, achieving 57--59\% token reduction on MATH-500.
ConPress~\citep{ConPress} exploits multi-question contextual pressure to elicit self-compressed traces, yielding 30--60\% reductions via SFT.
Accordion-Thinking~\citep{Accordion-Thinking} trains models to produce periodic step summaries that can be discarded, achieving 3$\times$ throughput with a gap-vanishing phenomenon during RL training.
CEEH~\citep{CEEH} proposes difficulty-aware entropy regularization to prevent entropy collapse under length constraints.

Our analysis complements these efforts by showing that compression which suppresses uncertainty expression, even when producing correct answers, can harm OOD generalization, and that this effect is modulated by task coverage. Therefore, the key challenge is to filter out non-informative content while retaining epistemic expressions that enable iterative belief refinement, rather than blindly compressing the reasoning process.

\newpage
\section{Basis for Selecting Epistemic Tokens}
\label{sec:epistemic-token-selection}

To provide an empirical basis for our selection of epistemic tokens, we analyzed uncertainty expressions across a diverse set of reasoning models. We collected 9,600 reasoning traces from eight models spanning four model families, including the Qwen3 family \citep{qwen3} (Qwen3-1.7B, Qwen3-8B, Qwen3-14B, and Qwen3-32B), the QwQ family \citep{qwen3} (QwQ-32B), the DeepSeek-R1-Distill-Qwen family \citep{deepseek-r1} (7B and 32B), and the OLMo family \citep{olmo} (OLMo-3-7B-Instruct). 

The traces were generated on 600 math problems drawn from AIME 2024/2025,
AMC 2023, and MATH500, with two samples per problem for each model.
Each reasoning trace was then provided to GPT-5 with the following prompt:

\begin{tcolorbox}[
  colback=white,
  title=\small{Prompt for Epistemic Token Extraction},
  colframe=gray,
  coltitle=black,
  colbacktitle=gray!10,
  breakable,
]
\small
From the reasoning trace below, extract every span where the model expresses
uncertainty or doubt (hesitation, second-guessing, hedging, suspicion something
is wrong).

For each span return:
\begin{enumerate}[nosep, leftmargin=*]
  \item \texttt{`span'}: the exact uncertainty sentence(s), copied verbatim;
  \item \texttt{`trigger\_token'}: the single short token or phrase
  (1--3 words, lowercased) inside that span that signals the uncertainty.
\end{enumerate}

\end{tcolorbox}

This procedure yielded over 99,000 $(\textit{span}, \textit{trigger\_token})$
pairs. We then ranked trigger tokens by frequency within each model family
(Qwen3, QwQ, DeepSeek, and OLMo).

As shown in Table~\ref{tab:uncertainty_triggers}, the ten markers
\textit{maybe}, \textit{wait}, \textit{check}, \textit{hmm}, \textit{seems},
\textit{perhaps}, \textit{might}, \textit{actually}, \textit{likely}, and
\textit{alternatively}, together with their natural variants, consistently
appeared among the top-ranked uncertainty expressions, supporting
their inclusion across model families.
While several tokens, such as \textit{let me try/verify}, \textit{I think},
and \textit{make sure}, were excluded, our selected markers cover
75--93\% of the occurrences of the phrases in
Table~\ref{tab:uncertainty_triggers}.

\begin{table*}[h!]
\centering
\small
\caption{Top-15 uncertainty trigger tokens by frequency across model families.
Percentages indicate the proportion of extracted trigger-token occurrences
within each family.}
\label{tab:uncertainty_triggers}

\setlength{\tabcolsep}{5pt}
\resizebox{\textwidth}{!}{%
\begin{tabular}{c l l l l}
\toprule
\textbf{Rank}
& \textbf{Qwen3 ($n=55{,}743$)}
& \textbf{QwQ ($n=10{,}817$)}
& \textbf{DeepSeek ($n=27{,}386$)}
& \textbf{OLMo ($n=4{,}948$)} \\
\midrule

1  & \textbf{maybe} (37.9\%)
   & \textbf{maybe} (34.3\%)
   & \textbf{maybe} (23.5\%)
   & \textbf{perhaps} (31.4\%) \\

2  & let me \textbf{check} (12.3\%)
   & \textbf{hmm} (16.0\%)
   & \textbf{hmm} (19.5\%)
   & \textbf{maybe} (19.9\%) \\

3  & \textbf{wait} (12.1\%)
   & \textbf{wait} (12.3\%)
   & I think (13.0\%)
   & \textbf{wait} (18.4\%) \\

4  & \textbf{hmm} (8.6\%)
   & let me \textbf{check} (8.6\%)
   & \textbf{wait} (10.6\%)
   & \textbf{likely} (5.1\%) \\

5  & I think (7.0\%)
   & \textbf{perhaps} (7.2\%)
   & \textbf{perhaps} (6.2\%)
   & \textbf{actually} (3.1\%) \\

6  & let me think (4.4\%)
   & I think (5.0\%)
   & double-\textbf{check} (4.3\%)
   & double-\textbf{check} (3.1\%) \\

7  & \textbf{perhaps} (3.0\%)
   & \textbf{might} (2.5\%)
   & \textbf{seems} (4.1\%)
   & I think (2.9\%) \\

8  & \textbf{seems} (2.7\%)
   & let me see (2.4\%)
   & but \textbf{wait} (3.3\%)
   & let's see (2.7\%) \\

9  & let me (2.4\%)
   & let me think (2.2\%)
   & let me think (3.0\%)
   & \textbf{might} (2.3\%) \\

10 & make sure (2.3\%)
   & make sure (1.9\%)
   & make sure (2.9\%)
   & \textbf{hmm} (2.0\%) \\

11 & but \textbf{wait} (2.1\%)
   & hold on (1.8\%)
   & just to make sure (2.2\%)
   & let's \textbf{check} (2.0\%) \\

12 & let me try (1.5\%)
   & \textbf{seems} (1.8\%)
   & let me (2.1\%)
   & not sure (1.8\%) \\

13 & \textbf{might} (1.4\%)
   & let me confirm (1.5\%)
   & \textbf{might} (1.7\%)
   & but \textbf{actually} (1.8\%) \\

14 & let me verify (1.2\%)
   & but \textbf{wait} (1.3\%)
   & hold on (1.7\%)
   & \textbf{wait}, \textbf{actually} (1.8\%) \\

15 & not sure (1.1\%)
   & \textbf{wait} a second (1.3\%)
   & let me \textbf{check} (1.7\%)
   & \textbf{alternatively} (1.8\%) \\

\bottomrule
\end{tabular}%
}
\end{table*}

Moreover, to account for uncertainty expressions not captured by our token
set, we conducted an independent LLM-as-a-Judge annotation experiment
(Appendix~\ref{appendix:epistemic_llm_as_judge}) that captures the full range of uncertainty expressions.
The resulting distribution closely mirrors that obtained with our
token-based approach.

Finally, although the linguistic patterns used to express uncertainty vary
slightly across models, Qwen3, QwQ, and DeepSeek exhibit highly similar
distributions. OLMo also shares largely the same set of uncertainty tokens,
but differs in the relative frequency with which individual tokens are used.
The selected markers largely overlap with expressions that human readers
would naturally recognize as signals of uncertainty, plausibly because these
models are ultimately trained on human-generated text.

\newpage
\section{Experimental Details} \label{appendix:experimental_details}

\paragraph{Training} ~~ For GRPO and SDPO training, we built upon the SDPO implementation~(\url{https://github.com/lasgroup/SDPO}) and additionally incorporated the DAPO-Math-17k dataset. The original DAPO-Math-17k dataset uses the following prompt format:

\begin{quote}
\texttt{Solve the following math problem step by step. The last line of your response should be of the form Answer: \$Answer (without quotes) where \$Answer is the answer to the problem.\textbackslash n\textbackslash n\{question\}\textbackslash nRemember to put your answer on its own line after "Answer:".}
\end{quote}

\noindent We replaced this with a simpler format:

\begin{quote}
\texttt{\{question\}\textbackslash nPlease reason step by step, and put your final answer within \textbackslash boxed\{\}.}
\end{quote}

\noindent as we observed that this format consistently yielded higher evaluation performance. For reward verification, we used the scoring function from the verl framework, which extracts the answer from the \texttt{\textbackslash boxed\{\}} expression and verifies correctness via exact match followed by mathematical equivalence checking using math-verify, adapted from EleutherAI's lm-evaluation-harness~\citep{eval-harness}.

The GRPO and SDPO training hyperparameters are listed in Tables \ref{tab:hyperparameters_common}, \ref{tab:hyperparameters_grpo}, and \ref{tab:hyperparameters_sdpo}. For the experiments on the relationship between task coverage and learning performance in Figure \ref{fig:generalization_ood_eval}, we reduced the question batch size to 64 due to the use of smaller training questions.

\begin{table}[h!]
\centering
\small
\caption{Common hyperparameters shared by GRPO and SDPO.}
\label{tab:hyperparameters_common}
\newcolumntype{C}[1]{>{\centering\arraybackslash}p{#1}}
\begin{tabular}{p{3cm}p{4.5cm}C{5cm}}
\toprule
\textbf{Category} & \textbf{Parameter} & \textbf{Value} \\
\midrule
\multirow{2}{*}{Data}
& Max.\ prompt length & 2048 \\
& Max.\ response length & 20480 \\
\midrule
\multirow{3}{*}{Batching}
& Question batch size & 256 \\
& Mini batch size & 64 or 128 \\
& Number of rollouts & 8 \\
\midrule
\multirow{2}{*}{Rollout}
& Inference engine & vllm \\
& Temperature & 1.0 \\
\midrule
\multirow{4}{*}{Training}
& Optimizer & AdamW \\
& Warmup steps & 10 \\
& Weight decay & 0.01 \\
& Gradient clip norm & 1.0 \\
\bottomrule
\end{tabular}
\end{table}

\begin{table}[h!]
\centering
\small
\caption{GRPO-specific hyperparameters.}
\label{tab:hyperparameters_grpo}
\newcolumntype{C}[1]{>{\centering\arraybackslash}p{#1}}
\begin{tabular}{p{3cm}p{4.5cm}C{5cm}}
\toprule
\textbf{Category} & \textbf{Parameter} & \textbf{Value} \\
\midrule
\multirow{3}{*}{Loss}
& $\epsilon$-high & 0.28 \\
& Rollout IS clip & 2 \\
& KL coefficient ($\lambda$) & 0.0 \\
\midrule
Training & Learning rate & $1 \times 10^{-6}$ \\
\bottomrule
\end{tabular}
\end{table}

\begin{table}[h!]
\centering
\small
\caption{SDPO-specific hyperparameters.}
\label{tab:hyperparameters_sdpo}
\newcolumntype{C}[1]{>{\centering\arraybackslash}p{#1}}
\begin{tabular}{p{3cm}p{4.5cm}C{5cm}}
\toprule
\textbf{Category} & \textbf{Parameter} & \textbf{Value} \\
\midrule
\multirow{3}{*}{Loss}
& Distillation divergence & Jensen--Shannon \\
& Top-$k$ distillation & 100 \\
& EMA update rate & 0.0 \\
\midrule
Training & Learning rate & $1 \times 10^{-5}$ \\
\bottomrule
\end{tabular}
\end{table}

\paragraph{Evaluation}
We evaluated all models using the hyperparameters listed in Table~\ref{tab:hyperparameters_eval}, following the recommended settings from each model's official documentation.\footnote{\url{https://huggingface.co/Qwen/Qwen3-8B}}\textsuperscript{,}\footnote{\url{https://huggingface.co/allenai/OLMo-3-7B-Instruct}}

\begin{table}[h!]
\centering
\small
\caption{Evaluation hyperparameters.}
\label{tab:hyperparameters_eval}
\newcolumntype{C}[1]{>{\centering\arraybackslash}p{#1}}
\begin{tabular}{p{4cm}C{2cm}C{1.8cm}C{1.8cm}C{1.8cm}}
\toprule
\textbf{Model} & \textbf{Max Tokens} & \textbf{Temp.} & \textbf{Top-$p$} & \textbf{Top-$K$} \\
\midrule
DeepSeek-R1-Distill-7B & 38912 & 0.6 & 0.95 & 20 \\
Qwen3-8B (thinking) & 38912 & 0.6 & 0.95 & 20 \\
Qwen3-8B (non-thinking) & 38912 & 0.7 & 0.8 & 20 \\
OLMo-3-7B-Instruct & 38912 & 0.6 & 0.95 & 20 \\
\bottomrule
\end{tabular}
\end{table}

\paragraph{Chat Templates for Different Model Series}
Table~\ref{tab:chat-templates} summarizes the chat template formats used by several open-weight model families, each using distinct special tokens to delineate user and assistant turns.
The placeholder \texttt{\{prompt\}} denotes the user message.

\begin{table}[h]
\centering
\caption{Chat template formats by model series.}
\label{tab:chat-templates}
\small
\begin{tabular}{@{}lp{10cm}@{}}
\toprule
\textbf{Model} & \textbf{Template} \\
\midrule
DeepSeek-R1-Distill-7B &
\texttt{<|begin\_of\_sentence|><|User|>}\,\{prompt\}\newline
\texttt{<|Assistant|><think>} \\
\addlinespace
Qwen3-8B (think ON) &
\texttt{<|im\_start|>user\textbackslash n}\,\{prompt\}\,\texttt{<|im\_end|>}\newline
\texttt{<|im\_start|>assistant} \\
\addlinespace
Qwen3-8B (think OFF) &
\texttt{<|im\_start|>user\textbackslash n}\,\{prompt\}\,\texttt{<|im\_end|>}\newline
\texttt{<|im\_start|>assistant\textbackslash n<think>\textbackslash n</think>} \\
\addlinespace
OLMo-3-7B-Instruct &
\texttt{<|im\_start|>system\textbackslash n}\,\{sys\}\,\texttt{<|im\_end|>}\newline
\texttt{<|im\_start|>user\textbackslash n}\,\{prompt\}\,\texttt{<|im\_end|>\textbackslash n<|im\_start|>assistant} \\
\bottomrule
\end{tabular}
\end{table}

\newpage
\section{More Ablation Study} \label{appendix:more_ablation}

\subsection{Effect of Maximum Response Length}

In our main experiments, as detailed in Table \ref{tab:hyperparameters_common}, we set the maximum training response length to 20,480 tokens to avoid prematurely truncating the long reasoning trajectories. Under a budget-constrained setting, we additionally examine how the training behavior of GRPO and SDPO changes when the maximum response length is reduced to 4,096 tokens.

\begin{figure}[h!]
    \centering
    \includegraphics[width=0.44\linewidth]{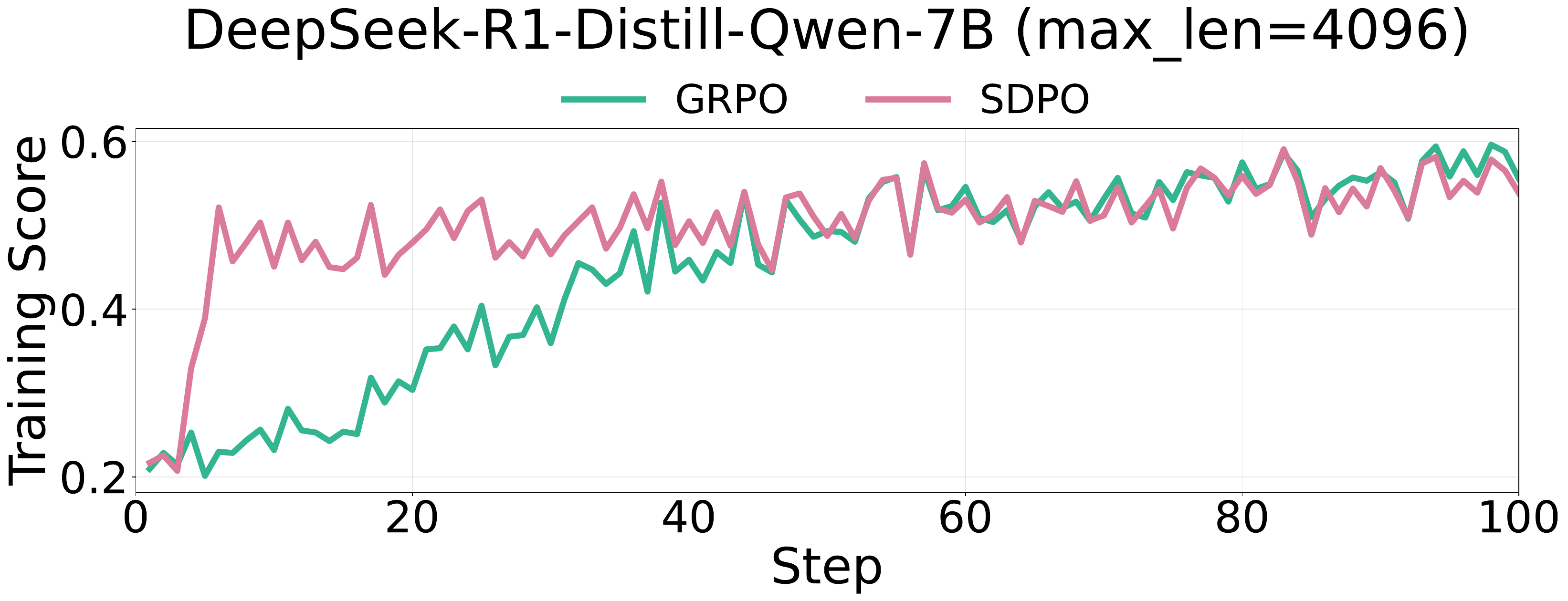}
    \hspace{0.1cm}
    \includegraphics[width=0.475\linewidth]{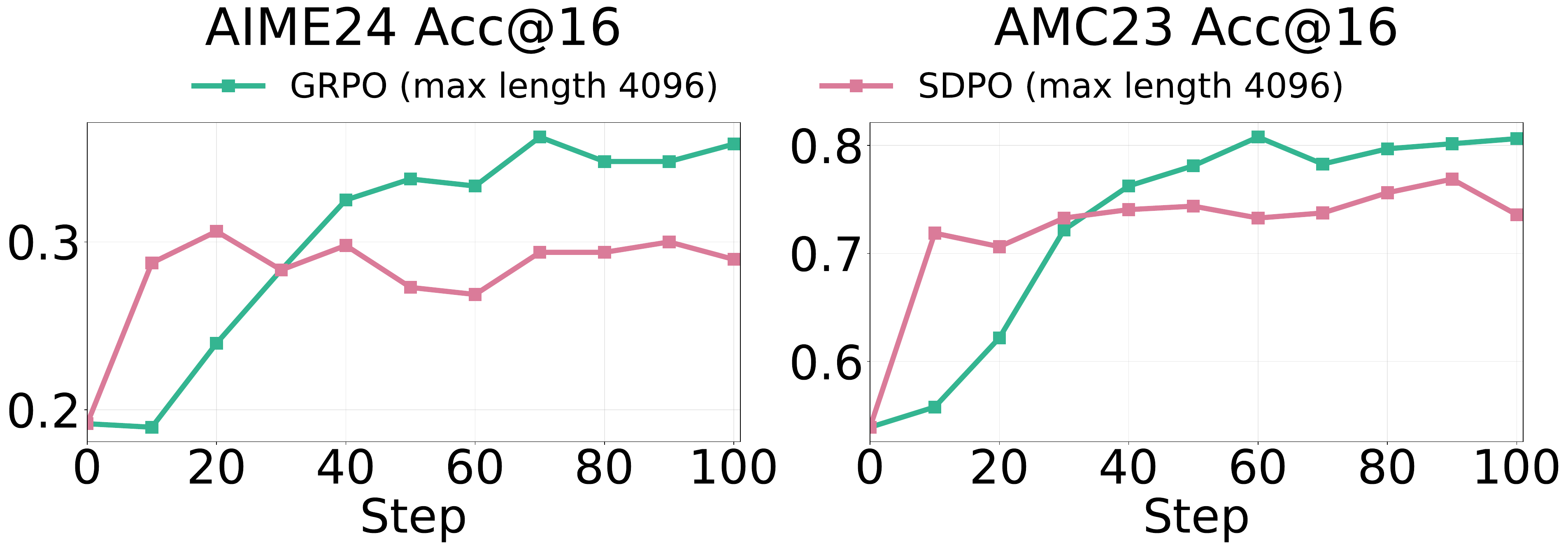}
    \caption{GRPO and SDPO with the maximum response length restricted to 4,096 tokens.}
    \label{fig:max_length_ablation}
\end{figure}

As shown in Figure~\ref{fig:max_length_ablation}, under the 4,096-token constraint, SDPO achieves relatively high training scores early in training, whereas GRPO improves more gradually and eventually reaches comparable training performance. More importantly, GRPO consistently achieves higher performance on AIME24 and AMC23, while SDPO plateaus at lower performance. Under a strict response budget, SDPO's tendency to produce shorter reasoning traces allows more responses to be completed within the available token budget, which likely contributes to its rapid initial gains. However, this early advantage does not translate into continued improvement in reasoning performance.

\subsection{Top-$k$ Distillation and Learning Rate}

\begin{figure}[h!]
    \centering
    \begin{subfigure}[b]{0.49\linewidth}
        \centering
        \includegraphics[width=\linewidth]{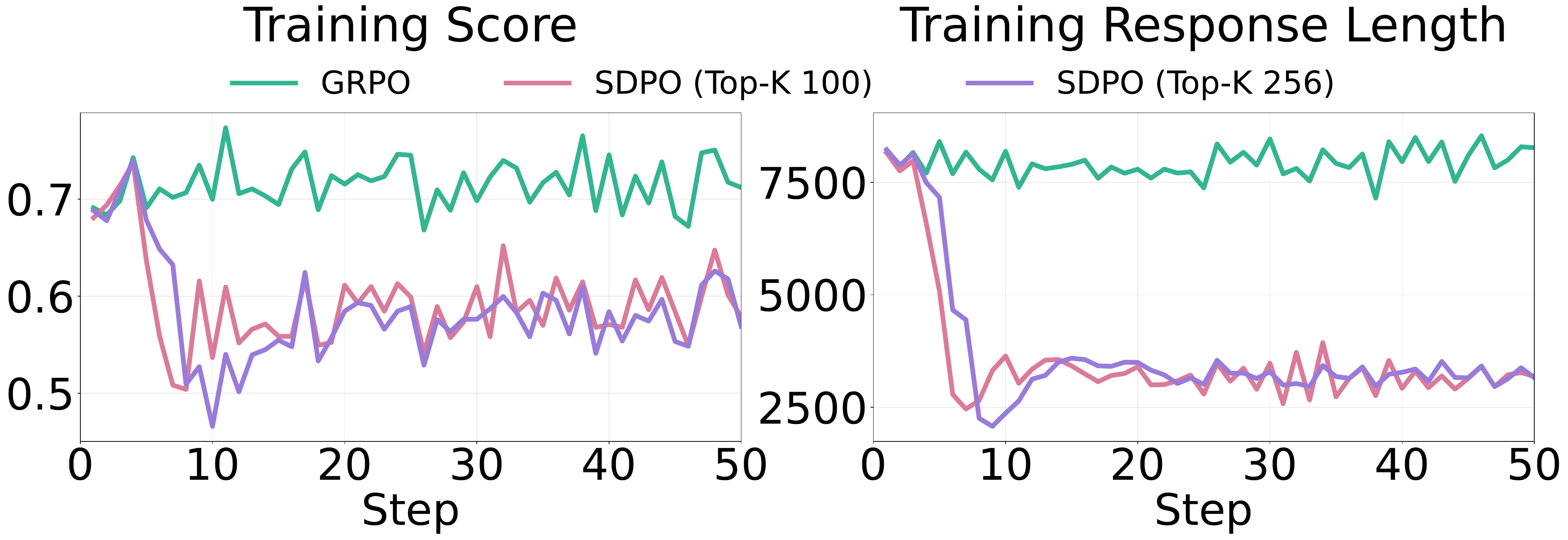}
        \caption{Effect of top-$k$ distillation ($k=100$ vs.\ $k=256$).}
        \label{fig:ablation_top_k}
    \end{subfigure}
    \begin{subfigure}[b]{0.49\linewidth}
        \centering
        \includegraphics[width=\linewidth]{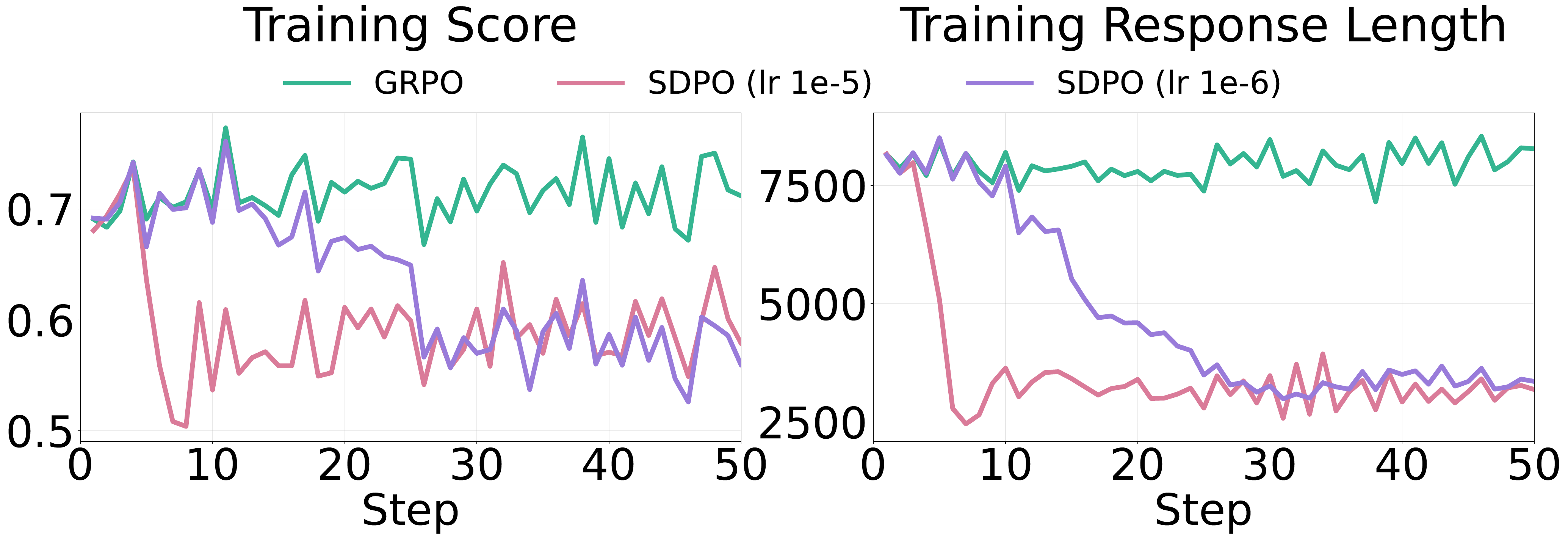}
        \caption{Effect of learning rate ($1\mathrm{e}{-5}$ vs.\ $1\mathrm{e}{-6}$).}
        \label{fig:ablation_lr}
    \end{subfigure}
    \caption{Ablation study on top-$k$ distillation and learning rate. Both modifications fail to prevent the convergence toward degraded reasoning behavior; a lower learning rate only delays the process.}
\end{figure}

To examine the effect of various training hyperparameters on self-distillation behavior, we conduct additional experiments by varying the top-$k$ distillation parameter and the learning rate. As shown in Figure~\ref{fig:ablation_top_k}, increasing top-$k$ from 100 to 256 yields no significant difference in training dynamics or final performance. Meanwhile, as illustrated in Figure~\ref{fig:ablation_lr}, reducing the learning rate from $1\mathrm{e}{-5}$ to $1\mathrm{e}{-6}$ slows the rate of degradation, but the model ultimately converges to the same reasoning behavior.

\newpage
\section{Additional Analysis of Epistemic Verbalization} 


\subsection{Per-Token Analysis of Epistemic Verbalization} \label{appendix:more_epistemic_behavior}

\begin{figure}[h!]
    \centering
    \includegraphics[width=\linewidth]{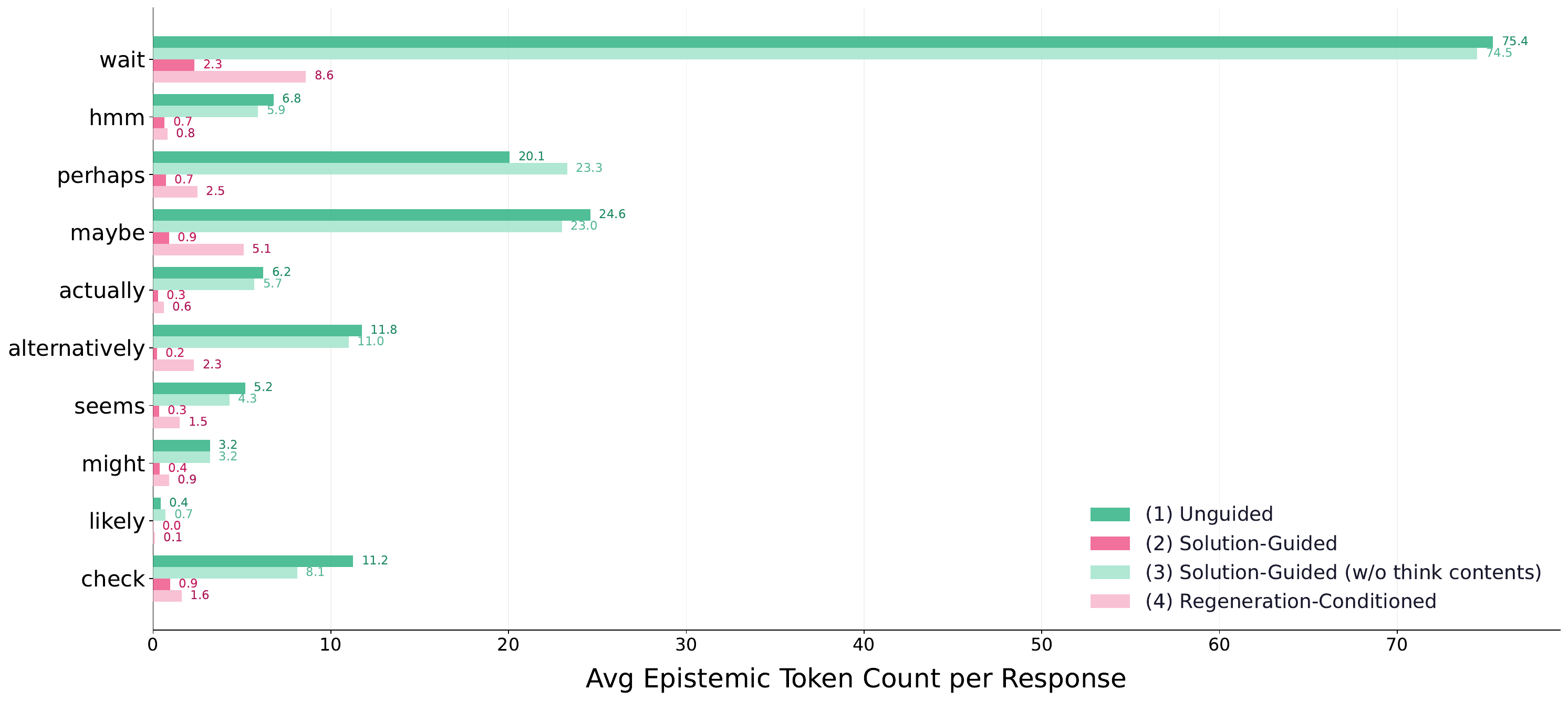}
    \caption{Per-token breakdown of epistemic token usage across the four generation settings. Each bar represents the average number of occurrences per response for an individual epistemic token. All tokens follow the same ordering as the aggregate trend, with \textit{wait}, \textit{maybe}, and \textit{perhaps} showing the largest variation across settings.}
    \label{fig:reasoning_behavior_epistemic_breakdown}
\end{figure}

In Table \ref{tab:reasoning_behavior} of Section \ref{sec:llm-reasoning-behavior}, we compared the average number of ten epistemic tokens per response. Figure \ref{fig:reasoning_behavior_epistemic_breakdown} further extends this analysis by examining how the average per-response count of each individual token changes under varying levels of conditioning information. When examining the per-token counts, all tokens exhibit a consistent trend:
\begin{equation}
\mathbb{E}\bigl[E(y)\bigr]\Big|_{(1)}
>
\mathbb{E}\bigl[E(y)\bigr]\Big|_{(3)}
>
\mathbb{E}\bigl[E(y)\bigr]\Big|_{(4)}
>
\mathbb{E}\bigl[E(y)\bigr]\Big|_{(2)},
\end{equation}
where tokens such as \textit{wait}, \textit{maybe}, and \textit{perhaps} are particularly prominent.

\subsection{Comparison of Epistemic Token Usage Across Models} \label{appendix:comparison_across_models}

Following the analysis of DeepSeek-R1-Distill-Qwen-7B (DeepSeek-Distill-7B) in Section~\ref{sec:llm-reasoning-behavior}, we further compare epistemic token usage across three settings: DeepSeek-Distill-7B, Qwen3-8B (Thinking Mode: ON), and Qwen3-8B (Thinking Mode: OFF).

\begin{figure}[h!]
    \centering
    \includegraphics[width=\linewidth]{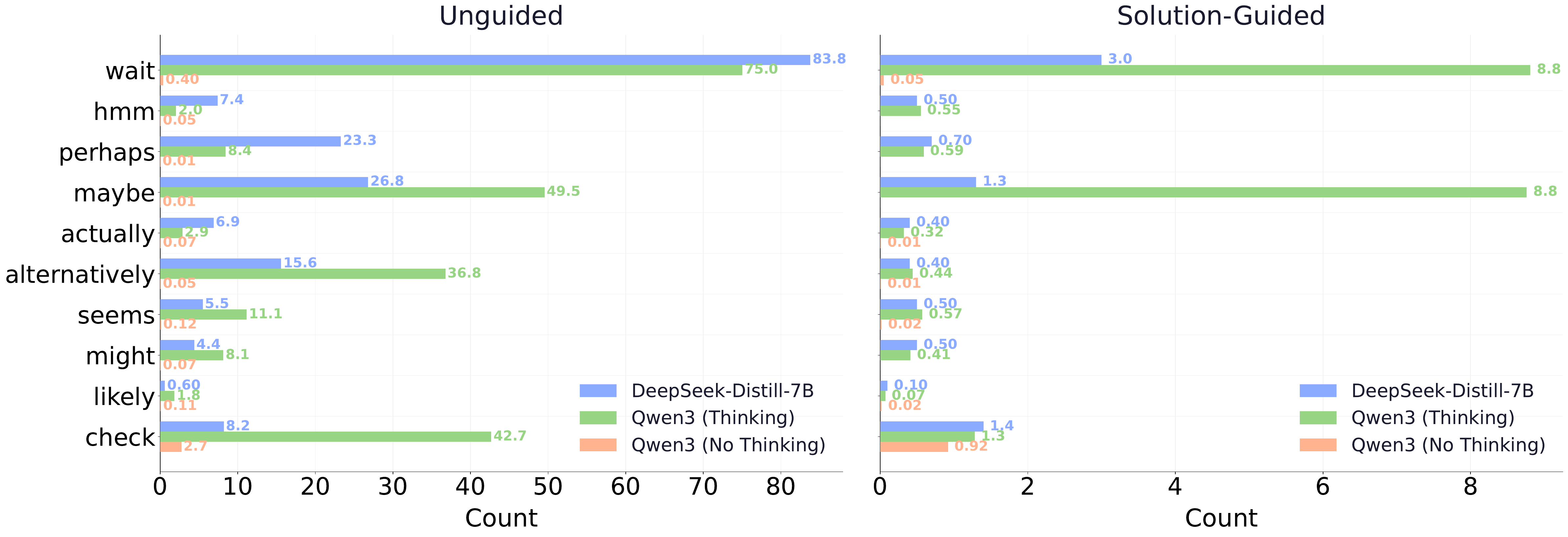}
    \caption{Comparison of epistemic token usage across DeepSeek-R1-Distill-Qwen-7B, Qwen3-8B (thinking enabled), and Qwen3-8B (thinking disabled).}
    \label{fig:epistemic_token_across_models}
\end{figure}

As shown in Figure~\ref{fig:epistemic_token_across_models}, both DeepSeek-Distill-7B and Qwen3-8B (Thinking Mode: ON) produce substantially more epistemic tokens than Qwen3-8B (Thinking Mode: OFF). While the two thinking-enabled models share a similar tendency to express uncertainty, they differ in their preferred epistemic tokens. For instance, DeepSeek-Distill-7B frequently uses \textit{wait} and employs \textit{perhaps} and \textit{maybe} at comparable rates, whereas Qwen3-8B uses \textit{perhaps} relatively less and favors \textit{maybe}. Qwen3-8B also uses \textit{alternatively} and \textit{check} far more than DeepSeek-Distill-7B, and overall embeds a greater amount of uncertainty within its reasoning.

Extending the discussion in Section~\ref{sec:llm-reasoning-behavior}, we also observe that Qwen3-8B generates far fewer epistemic tokens under solution-guided generation than under unguided generation. Across all three settings, Qwen3-8B (Thinking Mode: ON) produces the most epistemic tokens, followed by DeepSeek-Distill-7B, and then Qwen3-8B (Thinking Mode: OFF).

\subsection{Analysis of GRPO and SDPO Reasoning Patterns by Task Coverage} \label{appendix:more_epistemic_generalization}

\begin{figure}[h!]
    \centering
    \includegraphics[width=\linewidth]{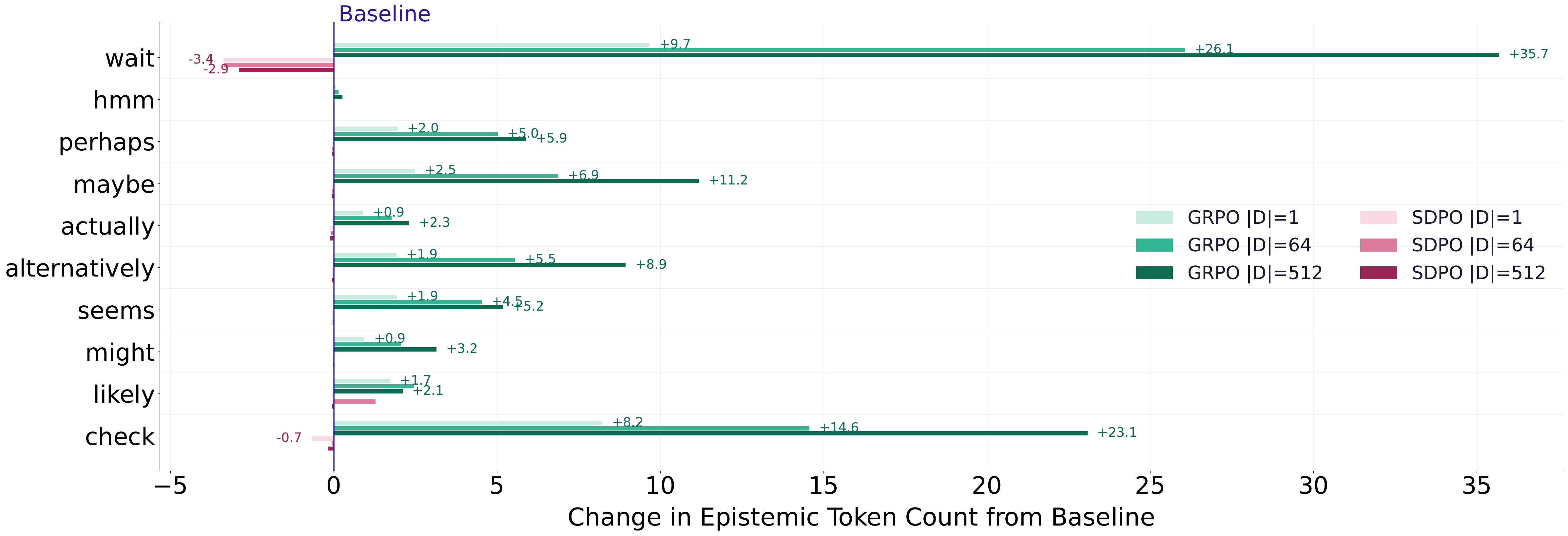}
    \caption{Change in epistemic token counts relative to the base model on AIME24 for GRPO and SDPO with $|D| \in \{1, 64, 512\}$.}
    \label{fig:epistemic_diff_aime24}
\end{figure}

To provide a more in-depth analysis of the results in Figure~\ref{fig:generalization_ood_eval} (Section~\ref{sec:task_coverage_performance}), we compare the change in epistemic token counts relative to the base model on AIME24 across six training configurations: GRPO and SDPO, each with $|D| \in \{1, 64, 512\}$. Figure~\ref{fig:epistemic_diff_aime24} shows that GRPO tends to increase epistemic token usage relative to the base model as $|D|$ grows, whereas SDPO reduces epistemic token usage, with smaller $|D|$ leading to a greater reduction.

\subsection{Comparative Analysis of Epistemic and Overall Word Usage} \label{appendix:total_vs_epistemic}
In addition to analyzing epistemic tokens, we examine how the overall word distribution in model responses shifts across the base model, GRPO-trained model, and SDPO-trained model. We first extract English words from each response using regular expressions and construct a reference vocabulary $\mathcal{V}$ from all unique words appearing in the base model responses. For each word $w \in \mathcal{V}$ and model $m \in \{\text{base}, \text{GRPO}, \text{SDPO}\}$, we compute the average per-response frequency:
\begin{equation}
    \mu_m(w) = \frac{1}{N}\sum_{i=1}^{N} f(w, r_i^{(m)}),
\end{equation}
where $f(w, r)$ denotes the number of occurrences of word $w$ in response $r$. We then define the frequency shift $\Delta_m(w) = \mu_m(w) - \mu_{\text{base}}(w)$ to capture how much each word's usage changes relative to the base model.

Using these frequency shifts, we compare the behavior of the predefined set of epistemic tokens $T$ (as defined in Section~\ref{sec:llm-reasoning-behavior}) against the rest of the vocabulary. Specifically, we compare $\{\Delta_m(w) \mid w \in T\}$ against $\{\Delta_m(w) \mid w \in \mathcal{V} \setminus T\}$ to determine whether the changes in epistemic token usage are disproportionately large relative to the overall shift in word distribution. This allows us to distinguish whether the training procedure specifically amplifies epistemic expressions or merely induces a uniform change across the vocabulary.
\begin{figure}
    \centering
    \includegraphics[width=\linewidth]{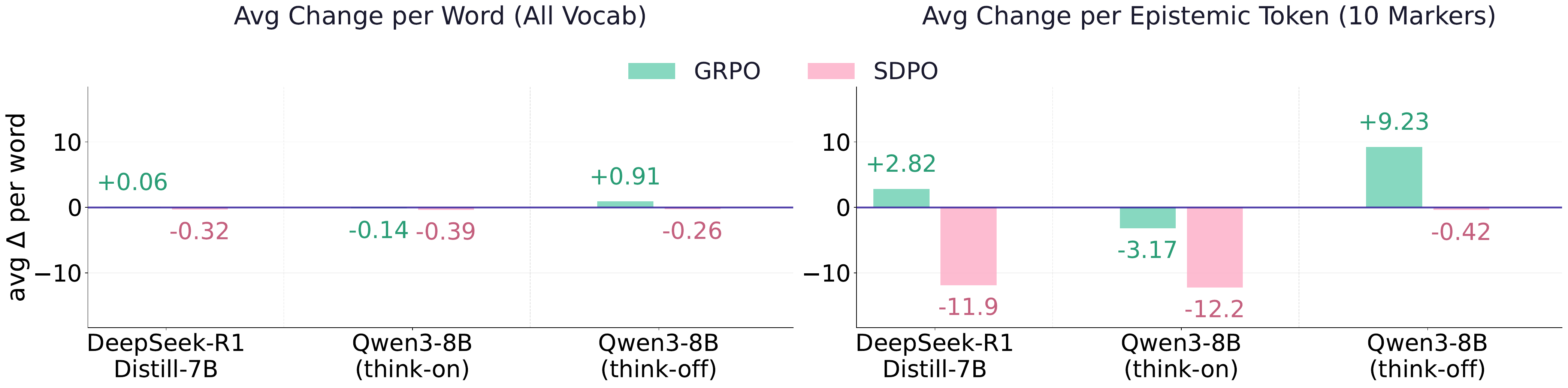}
    \caption{Comparison of per-word frequency shifts ($\Delta_m(w)$) for epistemic tokens ($T$) versus non-epistemic tokens ($\mathcal{V} \setminus T$) in GRPO- and SDPO-trained models relative to the base model. Epistemic tokens exhibit disproportionately large frequency changes compared to the general vocabulary, indicating that the training procedures specifically amplify or suppress epistemic expressions rather than inducing uniform changes across the word distribution.}
    \label{fig:token_change_comparison}
\end{figure}

As shown in Figure~\ref{fig:token_change_comparison}, the disparity between the two panels confirms that the observed epistemic shifts are not byproducts of general vocabulary drift. Across all model-method pairs, the average per-word shift over the full vocabulary remains near zero ($|\Delta| < 1$), whereas epistemic tokens exhibit shifts up to $30\times$--$40\times$ larger in magnitude (e.g., $-11.9$ and $-12.2$ for SDPO). This indicates that both training objectives selectively target epistemic expressions rather than uniformly perturbing the output distribution.

\subsection{Uncertainty Annotation via LLM-as-Judge} \label{appendix:epistemic_llm_as_judge}

We employed epistemic markers as indicators of uncertainty
expressions, as they serve as practical and interpretable cues
for regions where uncertainty externalization is likely to occur.
To more comprehensively capture diverse forms of uncertainty
expression, we additionally leveraged GPT-5.4~\citep{gpt5} as
an automated annotator. Specifically, GPT-5.4 was prompted to
identify and count uncertainty expressions in each
model-generated response. The prompt used for this annotation
is shown below.

\begin{tcolorbox}[
  colback=white,
  title=\small{Prompt for Uncertainty Annotation},
  colframe=gray,
  coltitle=black,
  colbacktitle=gray!10,
  breakable,
]
\small
You are an expert linguistic annotator. Your task is to count uncertainty expressions in a model-generated math reasoning response.

Uncertainty expressions include:
\begin{itemize}[nosep, leftmargin=*]
  \item \textbf{Hedging expressions:} expressions that indicate uncertainty (e.g., \textit{wait, hmm, perhaps, maybe})
  \item \textbf{Doubt markers:} expressions that explicitly convey uncertainty (e.g., \textit{I'm not sure, not certain, double-check, verify this} when used to express doubt)
\end{itemize}

Do NOT count:
\begin{itemize}[nosep, leftmargin=*]
  \item Confident statements such as ``we know'', ``clearly''
  \item Mathematical conditionals such as ``if $x = 0$, then\ldots''
\end{itemize}

Return ONLY a JSON object with exactly these fields:

\begin{verbatim}
{
  "count": <integer, total number of uncertainty expressions>,
  "examples": [<up to 5 short quoted phrases from the text>]
}
\end{verbatim}
\end{tcolorbox}

\begin{table}[h!]
\centering
\small
\caption{Representative examples of uncertainty expressions
identified by GPT-5.4. Expressions are drawn from model-generated
reasoning traces across all evaluated methods.}
\label{tab:uncertainty-examples}
\begin{tabular}{@{}l@{\hskip 3em}l@{}}
\toprule
\textbf{Uncertainty Expression} &  \\
\midrule
\textit{``maybe there's a simpler way''} & \textit{``Wait, hold on. Is that correct?''} \\
\textit{``I think the key here is to find $s$ and $t$ first, right?''} & \textit{``I'm not entirely sure''} \\
\textit{``perhaps there's a relationship here''} & \textit{``Hmm, that seems contradictory''} \\
\textit{``perhaps I made a miscalculation''} & \textit{``it seems a bit tricky''} \\
\textit{``Wait, but this might get complicated''} & \textit{``it seems less likely''} \\
\textit{``maybe we can assume some symmetry''} & \textit{``Alternatively, perhaps it's better} \\
\textit{``maybe we don't consider $n = 0$''} & \textit{\phantom{``}to think of it\ldots''} \\
\textit{``might not satisfy the surface area constraint''} &
\textit{``maybe I need another approach''} \\ 
\textit{``actually, no. Let me think again.''} \\
\bottomrule
\end{tabular}
\end{table}

Table~\ref{tab:uncertainty-examples} presents representative
examples of uncertainty expressions identified by the annotator.
The majority of identified expressions coincide with the
epistemic tokens examined in the previous analysis
(e.g., \textit{maybe}, \textit{perhaps}, \textit{wait}),
while others manifest as multi-token phrases such as
\textit{``I'm not entirely sure''} that cannot be captured
by single-token frequency alone. This confirms that epistemic
markers serve as a reliable proxy for uncertainty
externalization, while the LLM-as-Judge approach provides
complementary coverage over more diverse expression forms.

\begin{table}[h!]
\centering
\small
\begin{minipage}[t]{0.45\columnwidth}
\centering
\caption{Average number of uncertainty expressions per response,
as annotated by GPT-5.4.}
\label{tab:uncertainty-counts}
\begin{tabular}{@{}lc@{}}
\toprule
\textbf{Method} & \textbf{Avg.\ Unc.\ Expr.} \\
\midrule
Baseline & 21.36 \\
GRPO     & 24.36 \\
SDPO     & \textbf{9.18} \\
\bottomrule
\end{tabular}
\end{minipage}%
\hfill
\begin{minipage}[t]{0.5\columnwidth}
\centering
\vspace{0pt}
\includegraphics[width=0.9\linewidth]{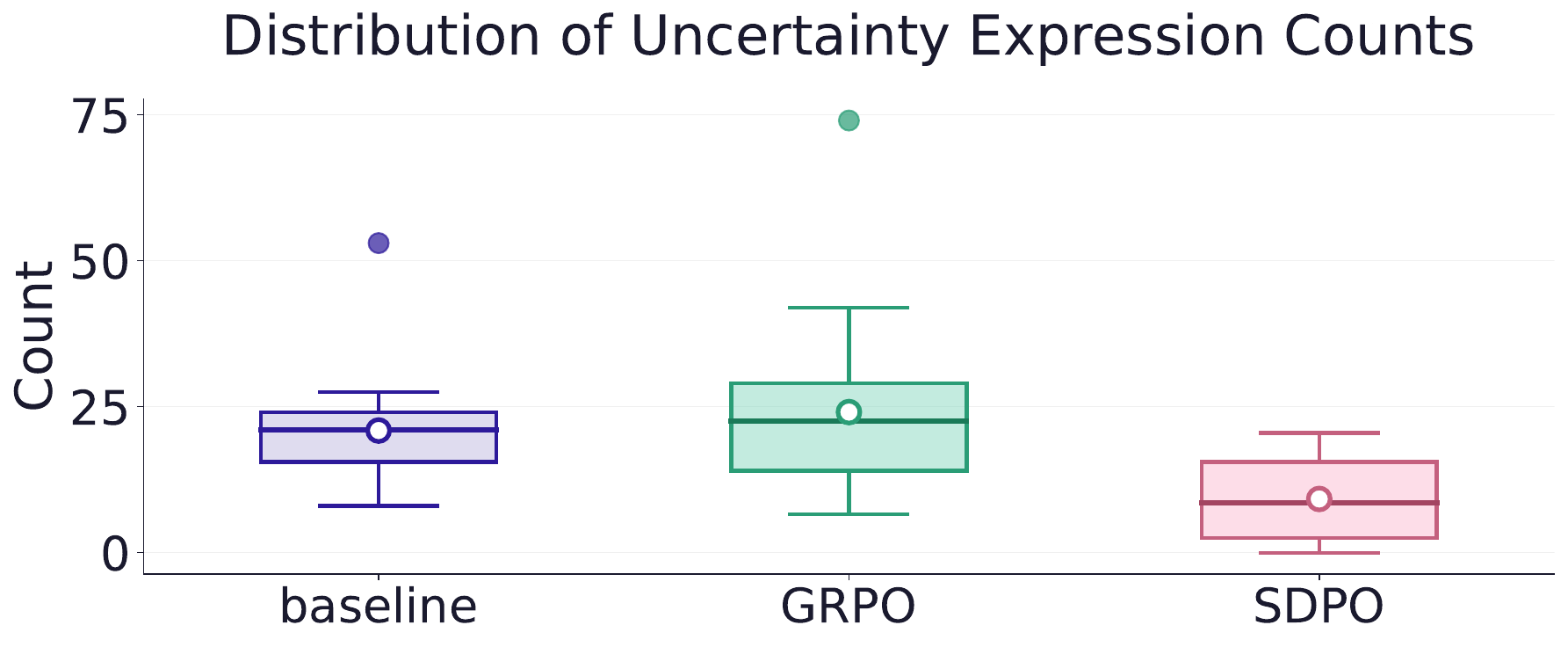}
\captionof{figure}{Box plot of uncertainty expression counts across methods.}
\label{fig:uncertainty-boxplot}
\end{minipage}
\end{table}

Table~\ref{tab:uncertainty-counts} reports the average number of
such expressions per response for each method. When evaluated
through LLM-as-Judge annotation, SDPO again substantially reduces
uncertainty expressions compared to both the baseline and GRPO,
consistent with the token-level analysis.
Figure~\ref{fig:uncertainty-boxplot} complements this with a box plot
comparison across methods: notably, GRPO exhibits considerably greater variance
than the baseline, suggesting that RLVR can amplify inconsistency in hedging behavior.
In contrast, SDPO maintains consistently low counts with a narrow
spread, confirming that on-policy self-distillation
suppresses unnecessary hedging and self-doubt during reasoning.

\newpage
\section{Student–Teacher KL Divergence on Epistemic Tokens} \label{appendix:kl}
We additionally measure the per-token KL divergence between the student $\pi_\theta(\cdot \mid x, y_{<t})$ and the self-teacher $\pi_\theta(\cdot \mid x, c, y_{<t})$ for each surface-form variant of the epistemic tokens in $\mathcal{T}$. As shown in Table~\ref{tab:uncertainty_kl}, every variant exceeds the corpus-wide mean KL of $0.076$, ranging from $1.9\times$ (\texttt{`Wait'}) to $6.8\times$ (\texttt{`\,Hmm'}). Notably, sentence-initial variants (e.g., \texttt{`\,Hmm'} at $6.8\times$, \texttt{`\,wait'} at $3.8\times$) tend to exhibit larger divergence than mid-sentence forms, suggesting that the self-teacher most strongly disagrees at positions where the student initiates epistemic deliberation. This provides a quantitative explanation for the suppression of epistemic verbalization under SDPO.

\begin{table}[h]
\centering
\small
\caption{KL divergence between student and self-teacher for epistemic tokens $\mathcal{T}$. The corpus-wide mean KL across all 1{,}642 frequent tokens is $0.076$.}
\label{tab:uncertainty_kl}
\begin{tabular}{llccc}
\toprule
Token & Variant & Mean KL & Count & Ratio \\
\midrule
\multirow{2}{*}{hmm}
  & \texttt{`\,Hmm'}          & 0.516 &    542 & $6.8\times$ \\
  & \texttt{`Hmm'}            & 0.337 &    887 & $4.4\times$ \\
\midrule
\multirow{3}{*}{maybe}
  & \texttt{`Maybe'}          & 0.432 &     52 & $5.7\times$ \\
  & \texttt{`\,Maybe'}        & 0.192 &    937 & $2.5\times$ \\
  & \texttt{`\,maybe'}        & 0.173 &  4,676 & $2.3\times$ \\
\midrule
actually
  & \texttt{`\,actually'}     & 0.188 &  1,134 & $2.5\times$ \\
\midrule
\multirow{2}{*}{alternatively}
  & \texttt{`\,Alternatively'} & 0.194 &   179 & $2.6\times$ \\
  & \texttt{`Alternatively'}   & 0.175 & 3,471 & $2.3\times$ \\
\midrule
might
  & \texttt{`\,might'}        & 0.171 &    684 & $2.3\times$ \\
\midrule
\multirow{2}{*}{perhaps}
  & \texttt{`Perhaps'}        & 0.255 &     54 & $3.4\times$ \\
  & \texttt{`\,perhaps'}      & 0.165 &  4,173 & $2.2\times$ \\
\midrule
\multirow{3}{*}{wait}
  & \texttt{`\,wait'}         & 0.285 &  1,066 & $3.8\times$ \\
  & \texttt{`\,Wait'}         & 0.196 &  1,314 & $2.6\times$ \\
  & \texttt{`Wait'}           & 0.146 & 15,000 & $1.9\times$ \\
\midrule
seems
  & \texttt{`\,seems'}        & 0.152 &    904 & $2.0\times$ \\
\bottomrule
\end{tabular}
\end{table}

\newpage

\section{Analysis of Changes in Model Internal Uncertainty under Self-Distillation}

To examine whether self-distillation also reduces the model's internal uncertainty, we further analyze token-level entropy as a measure of internal uncertainty. Building on the analysis in Section~\ref{sec:llm-reasoning-behavior}, we revisit the comparison between unguided and solution-guided generation to assess whether access to richer information is associated with lower predictive uncertainty.

\begin{table}[h!]
    \centering
    \small
    \caption{Token-level entropy under unguided and solution-guided generation.}
    \label{tab:entropy_guided}
    \begin{tabular}{lcc}
        \toprule
        Model &
        Unguided &
        Solution-guided \\
        \midrule
        DeepSeek-R1-Distill-Qwen-7B &
        0.385 &
        0.218 \\
        \bottomrule
    \end{tabular}
\end{table}

As shown in Table~\ref{tab:entropy_guided}, solution-guided generation reduces average token-level entropy by approximately 43\%, indicating substantially lower predictive uncertainty when richer information is available. Moreover, under unguided generation, epistemic tokens have an average entropy of 1.34, approximately $3.5\times$ higher than the overall average, indicating that epistemic verbalization tends to occur at positions of high predictive uncertainty.

We next examine whether this lower-entropy generation behavior is transferred to the student through self-distillation.

\begin{table}[h!]
    \centering
    \small
    \caption{Token-level entropy before and after GRPO and SDPO training.}
    \label{tab:entropy_training}
    \begin{tabular}{lccc}
        \toprule
        Model &
        Initial &
        GRPO &
        SDPO \\
        \midrule
        DeepSeek-R1-Distill-Qwen-7B &
        0.378 &
        0.401 &
        \textbf{0.248} \\
        Qwen3-8B (Thinking Mode: OFF) &
        0.100 &
        0.291 &
        \textbf{0.099} \\
        \bottomrule
    \end{tabular}
\end{table}

\begin{figure}[h!]
    \centering
    \includegraphics[width=0.33\linewidth]{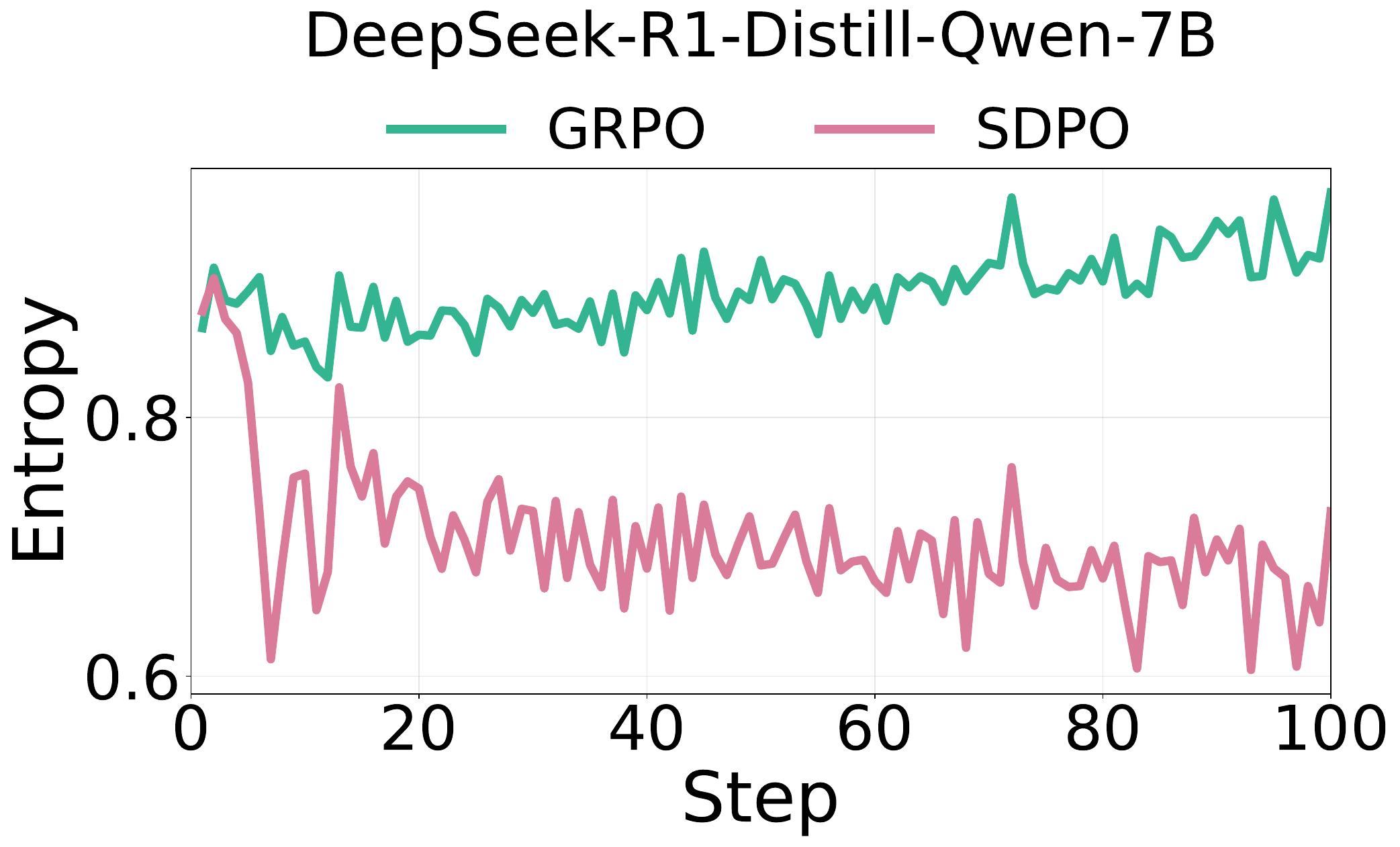}
    \includegraphics[width=0.33\linewidth]{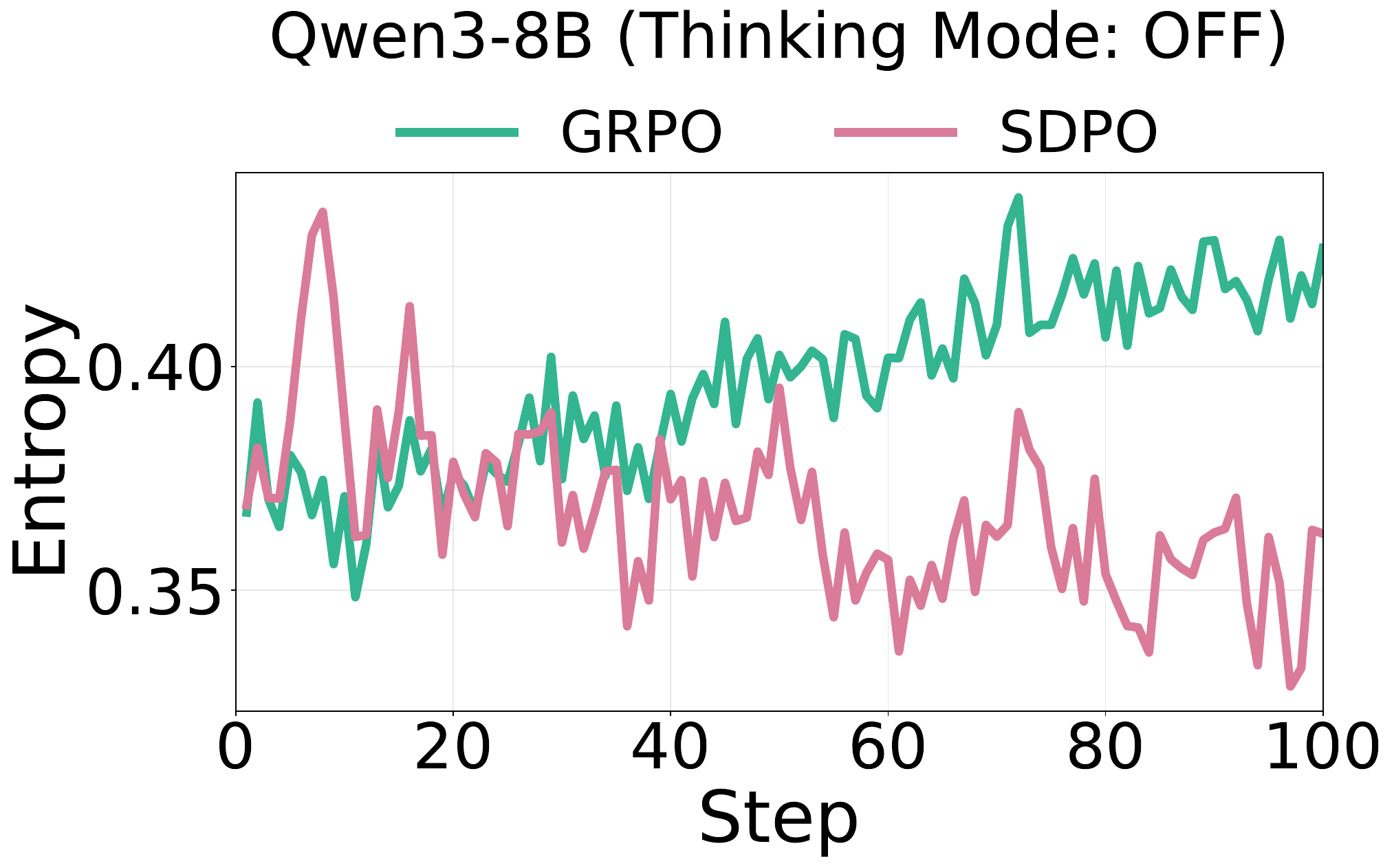}
    \caption{Average token-level entropy during GRPO and SDPO training.}
    \label{fig:entropy_training}
\end{figure}

Table~\ref{tab:entropy_training} and Figure~\ref{fig:entropy_training} show a consistent pattern in which GRPO increases token-level entropy for both models, whereas SDPO reduces it, and this distinction persists throughout training. These results indicate that the suppression of uncertainty expressions through self-distillation is accompanied by a genuine reduction in the model's internal uncertainty.

\newpage
\section{Can Uncertainty-Encouraging Prompts Mitigate the Suppression?} \label{appendix:different_prompt}

We further investigate whether prompting the teacher to produce more verbose reasoning with explicit expressions of uncertainty can mitigate the performance degradation caused by the suppression of uncertainty expressions. To this end, we design a modified regeneration prompt that encourages detailed, step-by-step reasoning with uncertainty, and repeat the reasoning behavior analysis (Section \ref{sec:llm-reasoning-behavior}) and on-policy self-distillation experiments (Section \ref{sec:opsd}) using this prompt.

\begin{table}[h!]
\centering
\small
\begin{tabular}{p{3.2cm} p{9.8cm}}
\toprule
Prompt for unguided generation & \texttt{\{question\}} \newline Please reason step by step, and put your final answer within \textbackslash boxed\{\}. \\
\midrule
Regeneration prompt (followed the prompt in \cite{SDPO}) & \texttt{\{question\}} \newline Please reason step by step, and put your final answer within \textbackslash boxed\{\}. \newline \textcolor{c-pink-dark}{Correct solution: \texttt{\{previously correct solution\}}} \newline Correctly solve the original question. \\
\midrule
Regeneration prompt with encouraging uncertainty & \texttt{\{question\}} \newline Please reason step by step, and put your final answer within \textbackslash boxed\{\}. \newline \textcolor{c-pink-dark}{Correct solution: \texttt{\{previously correct solution\}}} \newline Solve the original question \textbf{\textcolor{c-green-dark}{from scratch with a full, detailed reasoning trace, without skipping any steps, and include as much uncertainty or ambiguity in the explanation.}} \\
\bottomrule
\end{tabular}
\end{table}

\begin{table}[h!]
    \centering
    \small
    \renewcommand{\arraystretch}{1.3}
    \caption{Response characteristics of solution-guided generation with and without the uncertainty-encouraging prompt.}
    \label{tab:reprompt_behavior}
    \begin{tabular}{p{7cm}|P{1.85cm}P{3.5cm}}
        \hline
        & Avg. Length & Epistemic Token Count \\
        \hline
        Unguided & 13,054 & 182.5 \\
        Solution-Guided (vanilla) & 1,873 & 8.8 \\
        Solution-Guided (w/ encouraged uncertainty) & 3,098 & 31.4 \\
        \hline
    \end{tabular}
\end{table}

\begin{figure}[h!]
    \centering
    \begin{subfigure}[b]{0.49\linewidth}
        \centering
        \includegraphics[width=\linewidth]{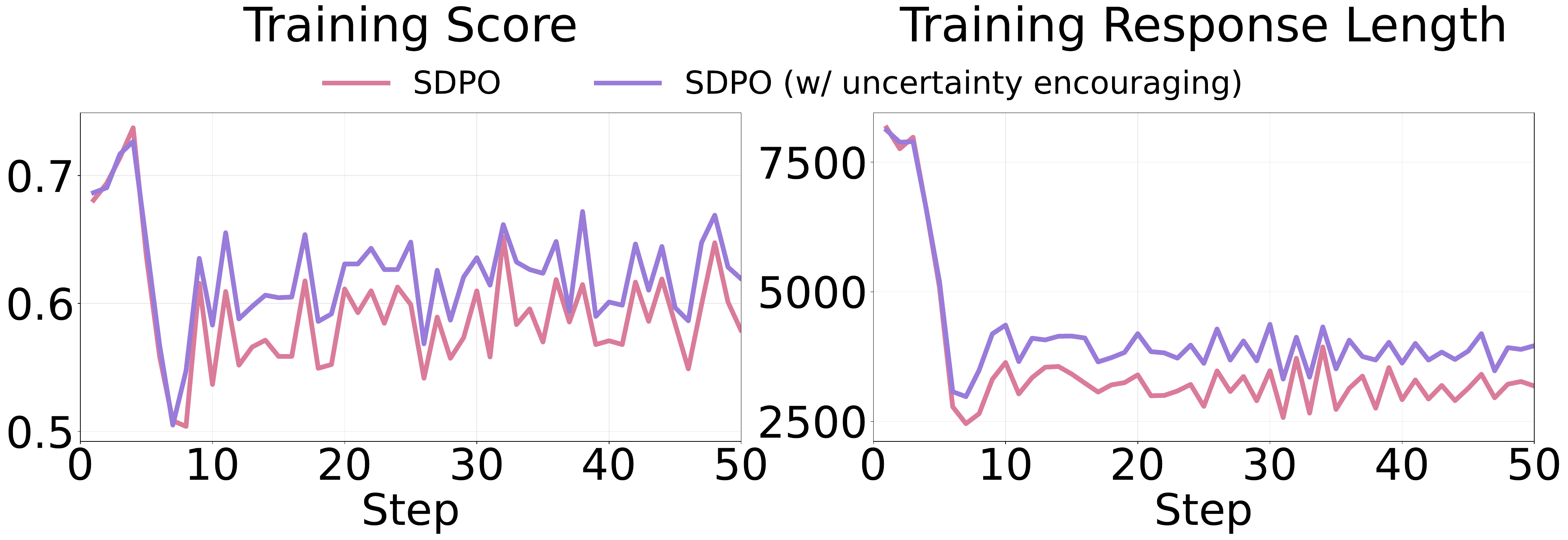}
        \caption{DeepSeek-Distill-7B Training Comparison}
        \label{fig:deepseek_reprompt_score_length}
    \end{subfigure}
    \begin{subfigure}[b]{0.49\linewidth}
        \centering
        \includegraphics[width=\linewidth]{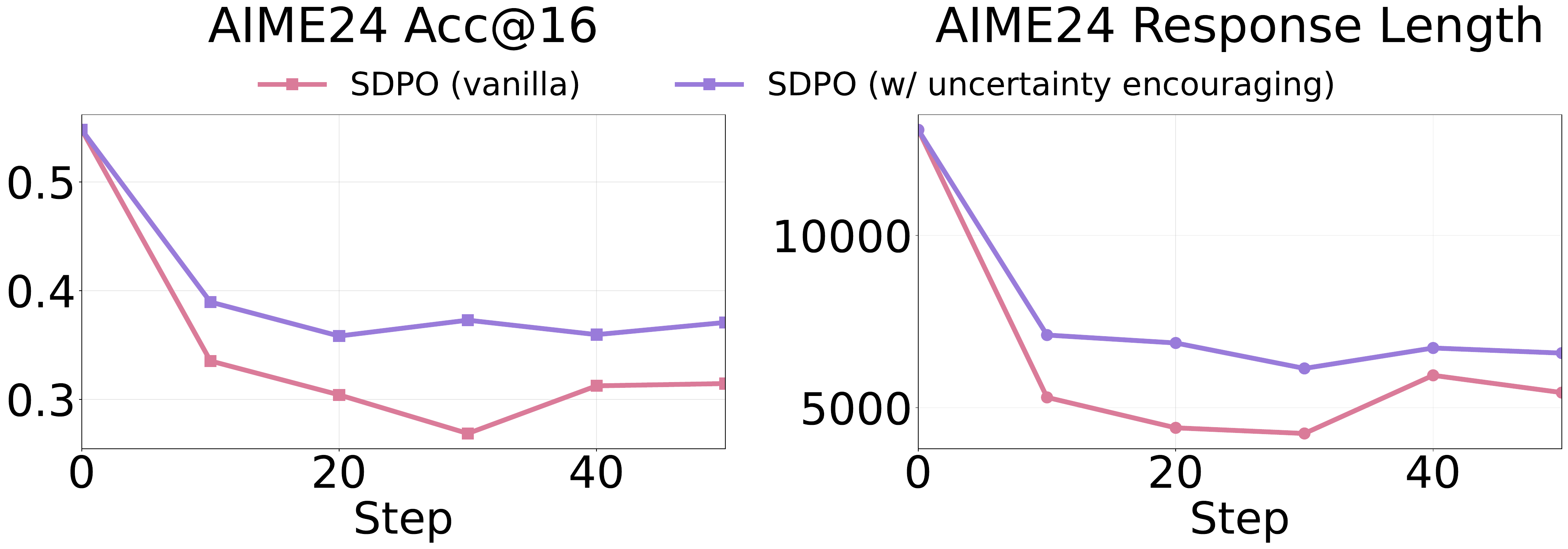}
        \caption{DeepSeek-Distill-7B AIME24 Comparison}
        \label{fig:deepseek_reprompt_aime24}
    \end{subfigure}
    \caption{Training dynamics and AIME24 evaluation of on-policy self-distillation using the uncertainty-encouraging regeneration prompt. (a) Training score and response length over training steps. (b) AIME24 Acc@16 and response length over training steps. While the uncertainty-encouraging prompt partially mitigates response shortening, the overall performance degradation persists.}
    \label{fig:deepseek_reprompt}
\end{figure}

Table \ref{tab:reprompt_behavior} shows that the uncertainty-encouraging prompt partially recovers the suppressed reasoning behavior: compared to the vanilla solution-guided prompt, it produces longer responses (3,098 vs.\ 1,873 tokens) and more epistemic expressions (31.4 vs.\ 8.8). However, both metrics remain far below those of unguided generation, indicating that the presence of a reference solution fundamentally constrains the model's reasoning verbosity regardless of the prompt instruction.

Figure \ref{fig:deepseek_reprompt} shows that the uncertainty-encouraging prompt leads to slightly higher training scores and longer responses compared to the vanilla setting, and this improvement carries over to AIME24 evaluation. Nevertheless, both variants still suffer from substantial performance degradation, suggesting that prompting alone is insufficient to resolve the fundamental issue of uncertainty suppression in solution-guided generation.

\newpage
\section{Comparison with OPSD from \citet{zhao2026self}} \label{appendix:comparison_opsd}

Recently, \citet{zhao2026self} proposed OPSD and demonstrated performance gains in mathematical reasoning through on-policy self-distillation, particularly on the Qwen3 series. Unlike our setup, where both the student and the teacher either enable or disable thinking mode, OPSD adopts a hybrid configuration in which the student operates with thinking mode disabled while the teacher has it enabled. 

As our experiments also confirm, enabling thinking mode produces substantially longer responses with a greater number of epistemic tokens, making this hybrid setup function more akin to conventional teacher–student distillation, despite using the same underlying model. We note that this configuration is inherently limited to model families such as Qwen3 that support toggling thinking mode on and off.

For training efficiency, OPSD trains only on a response prefix (1,024 tokens by default), rather than the full student response. Unlike OPSD, SDPO and our setup use full fine-tuning with verl \citep{hybridflow}, whereas OPSD adopts LoRA fine-tuning with trl \citep{trl} and a smaller batch size of 32 (vs. 256 in our setup), resulting in higher training efficiency but smaller parameter updates per step.

\begin{figure}[h!]
    \centering
    \begin{subfigure}[b]{0.4\linewidth}
        \centering
        \includegraphics[width=0.87\linewidth]{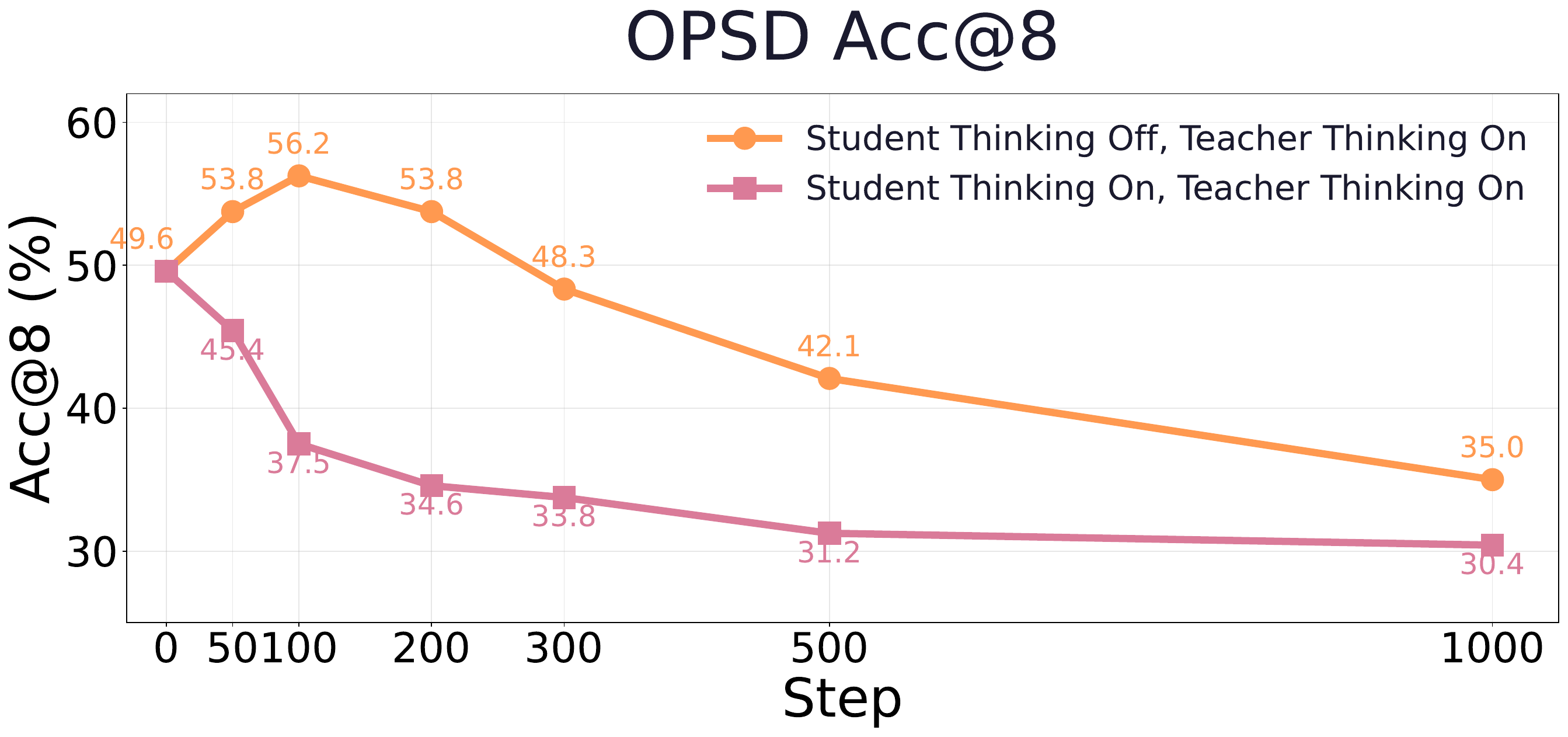}
        \caption{Hybrid vs. homogeneous setup performance over training.}
        \label{fig:opsd_avg8}
    \end{subfigure}
    \hfill
    \begin{subfigure}[b]{0.57\linewidth}
        \centering
        \includegraphics[width=\linewidth]{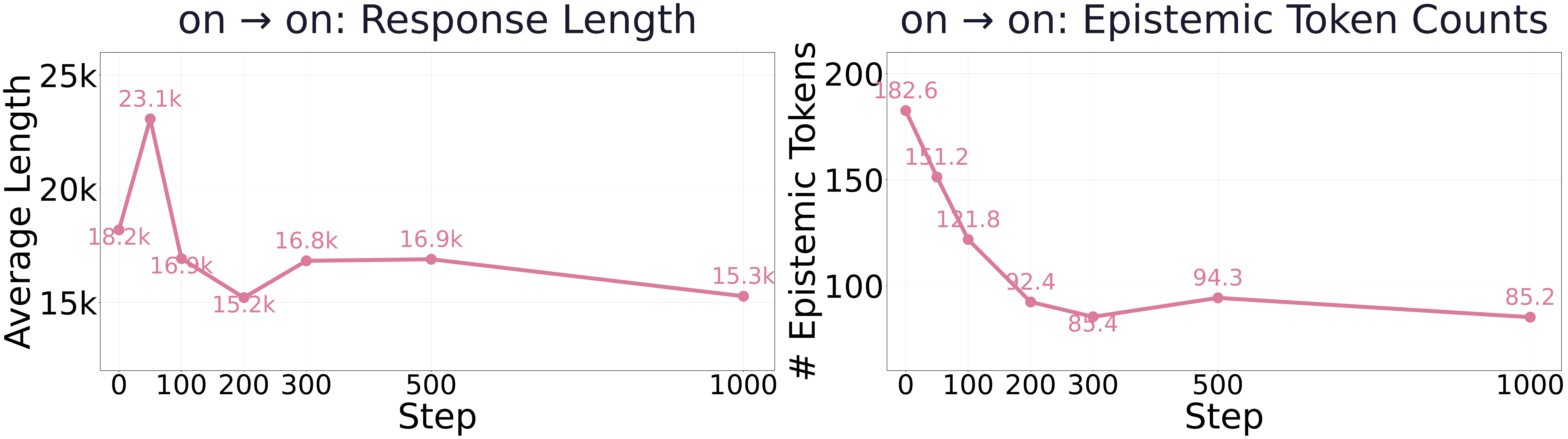}
        \caption{Response length and epistemic token usage in the homogeneous setup.}
        \label{fig:opsd_len}
    \end{subfigure}
    \caption{Training dynamics of OPSD hybrid distillation on Qwen3-1.7B compared with our homogeneous chat-template setup. (a) In the hybrid setup, the thinking-enabled teacher initially improves student performance, but the gains diminish over time, whereas the homogeneous setup shows a consistent decline. (b) Response length and epistemic token usage under the homogeneous setup.}
    \label{fig:opsd}
\end{figure}

Under this hybrid setup with prefix learning in Qwen3-1.7B, as shown in Figure \ref{fig:opsd_avg8} (\textcolor{orange}{orange line}), we observe an intriguing training dynamic. In the early stages, the thinking-enabled teacher encourages the student to generate longer responses while simultaneously improving performance, highlighting the effectiveness of hybrid distillation during the initial phase of training. However, as training progresses, performance steadily deteriorates despite this early improvement. In contrast, under our homogeneous setup, where both the student and teacher operate with thinking mode enabled (\textcolor{custompink}{pink line}), performance declines consistently throughout training. Response length and epistemic token count likewise decrease steadily, consistent with our analysis.

While this hybrid distillation setup for the Qwen3 series presents an interesting research direction with its own unique training dynamics, such as why performance initially improves before declining and whether this stems from changes in reasoning behavior or chat template mismatch, a thorough investigation is beyond the scope of this work and is left for future exploration.

\newpage
\section{More On-Policy Self-Distillation Results}

\subsection{Qwen3-8B (Thinking Mode: OFF)} \label{appendix:more_on_policy_qwen3_8b_nothink}
\begin{figure}[h!]
    \centering
    \begin{subfigure}[b]{0.49\linewidth}
        \centering
        \includegraphics[width=\linewidth]{figure/on-policy-distillation/qwen3-8b-nothink/score_length_comparison.pdf}
        \caption{Training score-length comparison}
        \label{fig:extended_nothink_score_length}
    \end{subfigure}
    \begin{subfigure}[b]{0.49\linewidth}
        \centering
        \includegraphics[width=\linewidth]{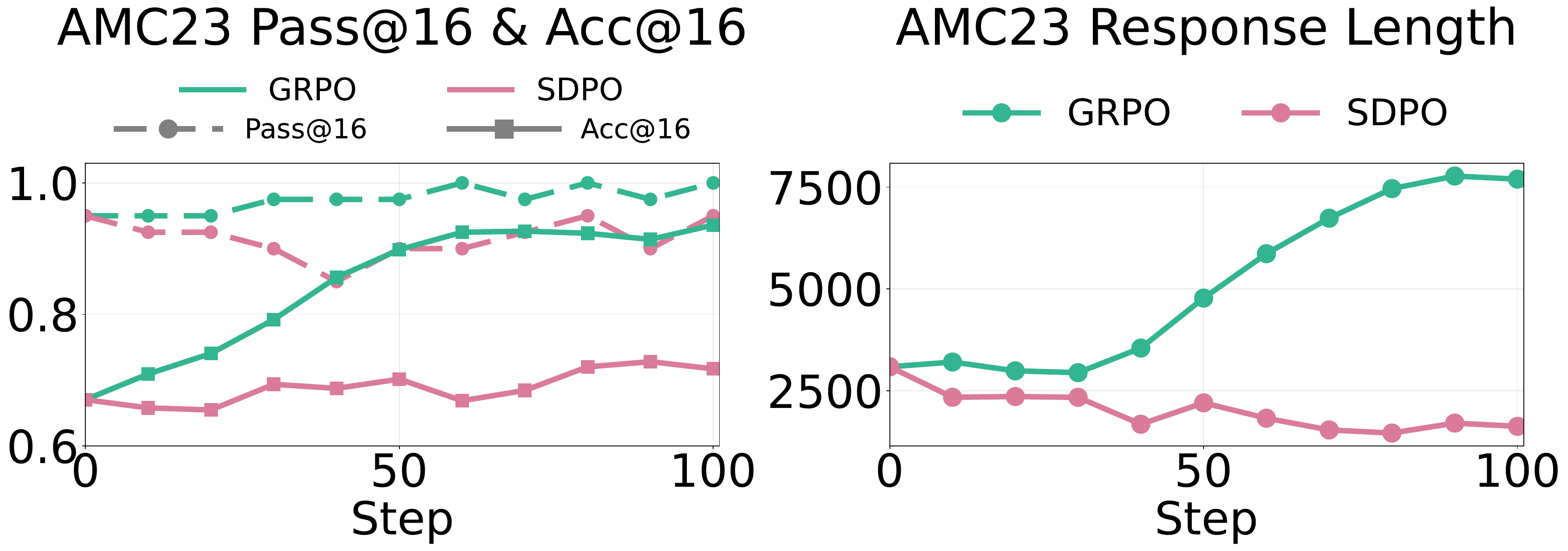}
        \caption{AMC23 score and response length}
        \label{fig:nothink_amc23}
    \end{subfigure} \\
    \vspace{0.3cm}
    \begin{subfigure}[b]{0.49\linewidth}
        \centering
        \includegraphics[width=\linewidth]{figure/on-policy-distillation/qwen3-8b-nothink/Qwen3_8B_think_off_aime24_combined.pdf}
        \caption{AIME24 score and response length}
        \label{fig:extended_nothink_aime24}
    \end{subfigure}
    \begin{subfigure}[b]{0.49\linewidth}
        \centering
        \includegraphics[width=\linewidth]{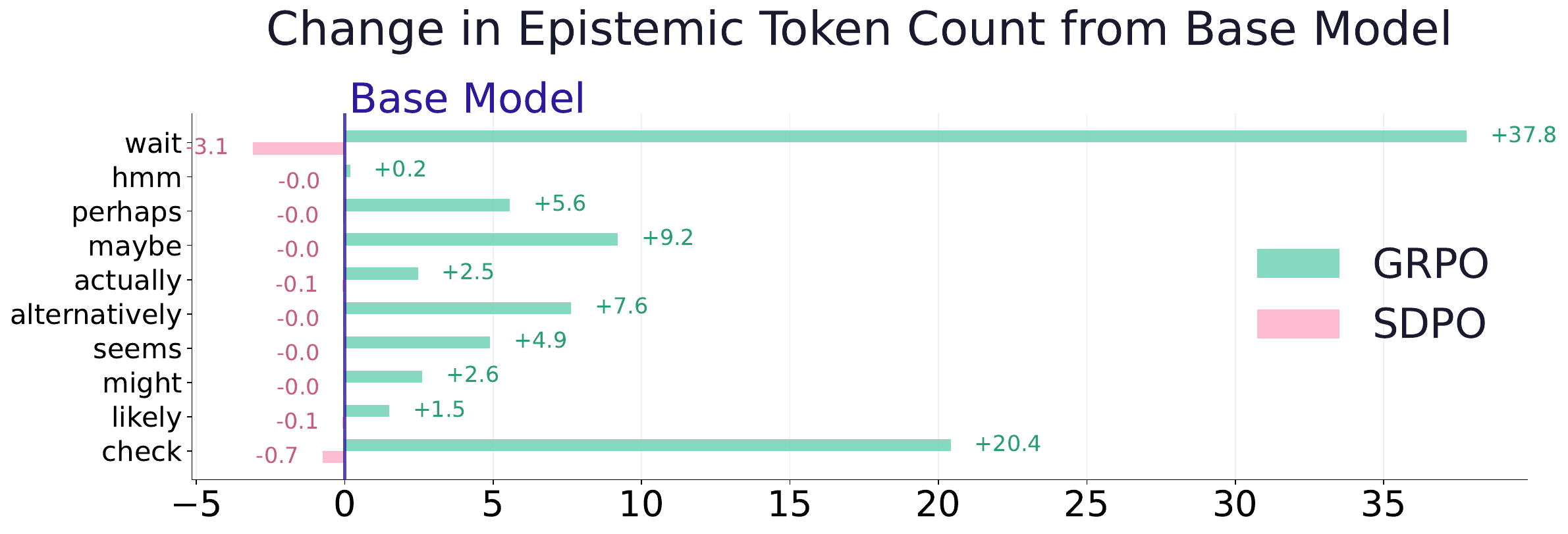}
        \caption{Change in epistemic token usage on AIME24}
        \label{fig:nothink_epistemic_change}
    \end{subfigure}
    \caption{Extended results for Qwen3-8B (thinking mode off): training score-length trade-off, evaluation performance on AMC23 and AIME24, and change in epistemic token usage.}
    \label{fig:nothink_extended}
\end{figure}

As an extension of Figure~\ref{fig:qwen3-8b-nothink} in Section~\ref{sec:qwen3-8b-nothink}, we additionally compare AMC23 evaluation scores and the change in epistemic token usage on AIME24. As shown in Figure~\ref{fig:nothink_extended}, GRPO significantly increases response length and yields substantial performance gains on both benchmarks. In contrast, SDPO exhibits divergent trends: on AMC23, acc@16 increases from 0.67 to 0.73 while reducing response length by roughly half, whereas on AIME24, acc@16 slightly decreases from 0.25 to 0.23, with pass@16 dropping more substantially. Notably on AMC23, SDPO achieves approximately a 6-point improvement with shorter responses, whereas GRPO obtains a much larger improvement of around 36 points at the cost of considerably longer responses. Achieving large performance gains while maintaining a reasonable response length remains an open challenge.

\subsection{Qwen3-1.7B (Thinking Mode: ON)}

\begin{figure}[h!]
    \centering
    \begin{subfigure}[b]{0.49\linewidth}
        \centering
        \includegraphics[width=\linewidth]{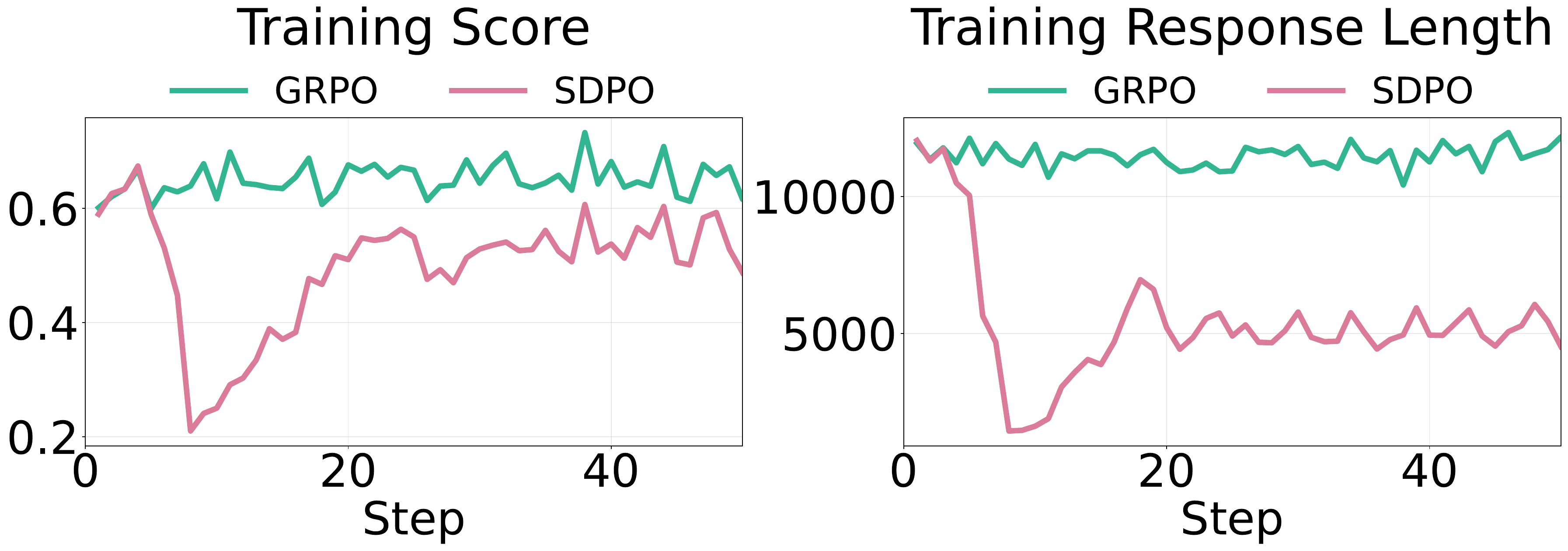}
        \caption{Training score-length comparison}
        \label{fig:qwen_1.7b_score_length}
    \end{subfigure}
    \begin{subfigure}[b]{0.49\linewidth}
        \centering
        \includegraphics[width=\linewidth]{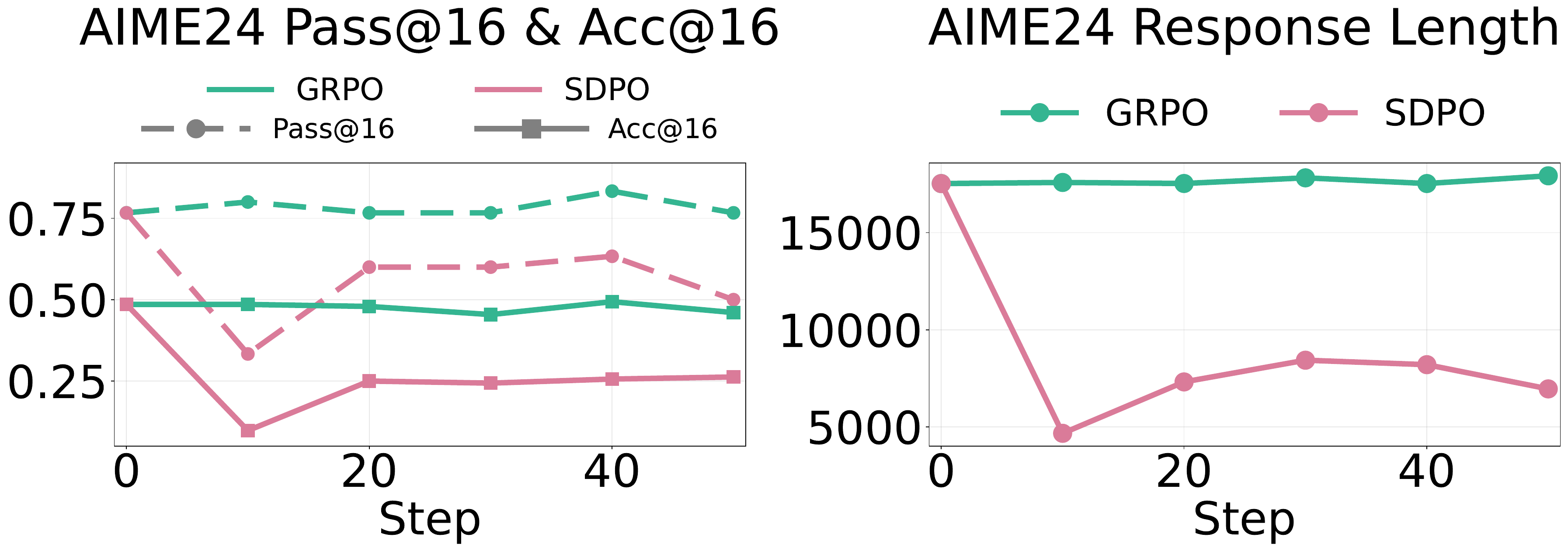}
        \caption{AIME24 score and response length}
        \label{fig:qwen_1.7b_aime24}
    \end{subfigure} \\
    \caption{GRPO and SDPO results on Qwen3-1.7B (Thinking Mode: ON): training score and response length, and OOD evaluation on AIME24. Self-distillation consistently reduces response length but leads to a substantial drop in performance, with a more pronounced degradation compared to larger models.}
    \label{fig:qwen_1.7b}
\end{figure}

To investigate whether our findings generalize to smaller models, we conducted the same on-policy self-distillation experiment on Qwen3-1.7B. As shown in Figure~\ref{fig:qwen_1.7b}, our finding that self-distillation reduces uncertainty and makes outputs more concise, while simultaneously degrading mathematical reasoning performance, remains consistent across model sizes. Notably, the performance degradation is more severe for the 1.7B model (45.9\%) compared to Qwen3-8B with Thinking Mode ON (12.1\%) and DeepSeek-R1-Distill-Qwen-7B (40.7\%). This suggests that the externalization of uncertainty may be particularly important for robust reasoning in smaller models, which possess more limited inherent capabilities.

\subsection{Olmo-3-7B-Instruct} \label{appendix:olmo}

\begin{figure}[h!]
    \centering
    \begin{subfigure}[b]{0.49\linewidth}
        \centering
        \includegraphics[width=\linewidth]{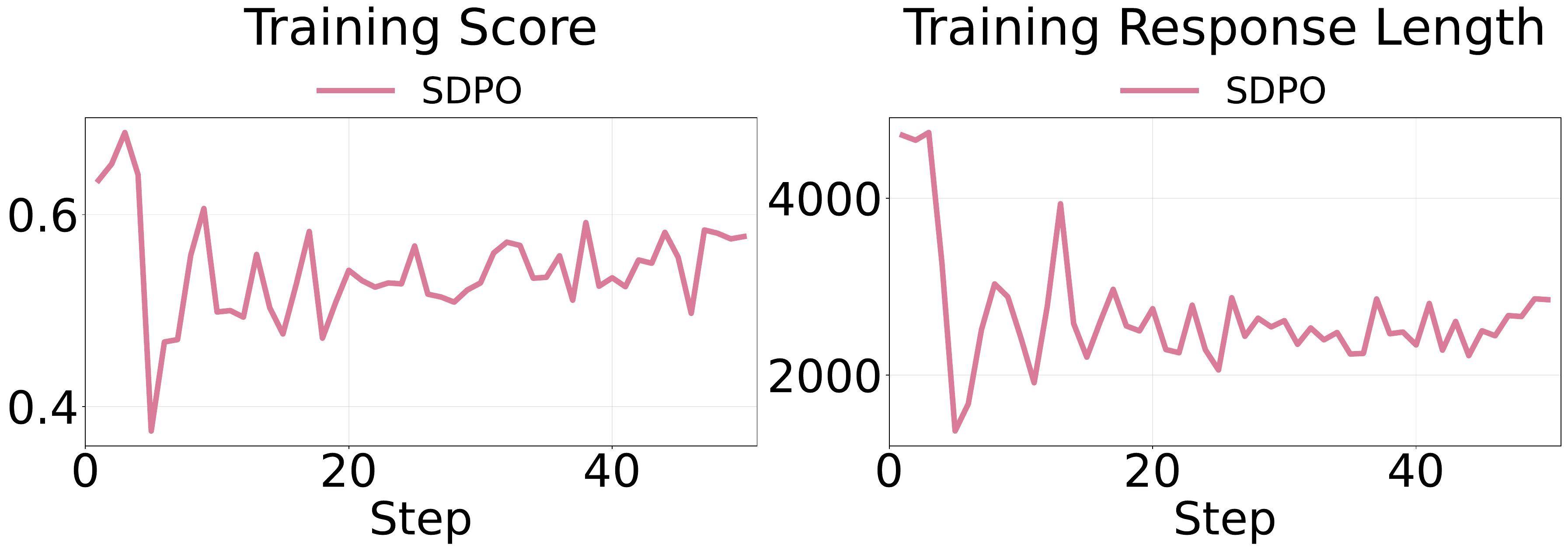}
        \caption{Training score-length comparison}
        \label{fig:olmo_score}
    \end{subfigure}
    \begin{subfigure}[b]{0.49\linewidth}
        \centering
        \includegraphics[width=\linewidth]{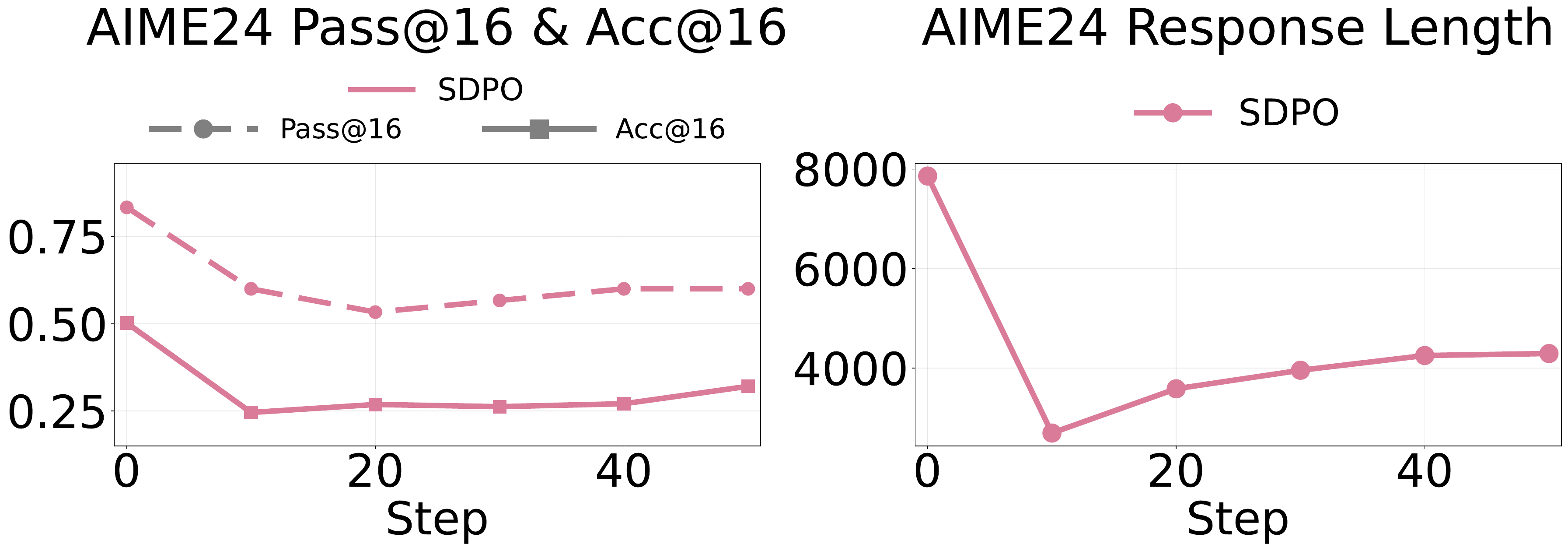}
        \caption{AIME24 score and response length}
        \label{fig:olmo_aime24}
    \end{subfigure}
    \caption{SDPO results on OLMo-3-7B-Instruct: training score and response length, and OOD evaluation on AIME24.}
    \label{fig:olmo_sdpo}
\end{figure}

In addition to DeepSeek-R1-Distill-Qwen-7B and Qwen3-8B, we further evaluate on-policy self-distillation on OLMo-3-7B-Instruct from a different model family. As shown in Figure~\ref{fig:olmo_sdpo}, consistent with our previous analyses, SDPO also degrades reasoning performance on this model, with OOD evaluation scores falling below those of the base model. This confirms that our findings are not model-dependent but reflect robust characteristics of reasoning behavior across diverse model families.

\subsection{Pass@16 Score}

\begin{figure}[h!]
    \centering
    \begin{subfigure}[b]{0.49\linewidth}
        \centering
        \includegraphics[width=\linewidth]{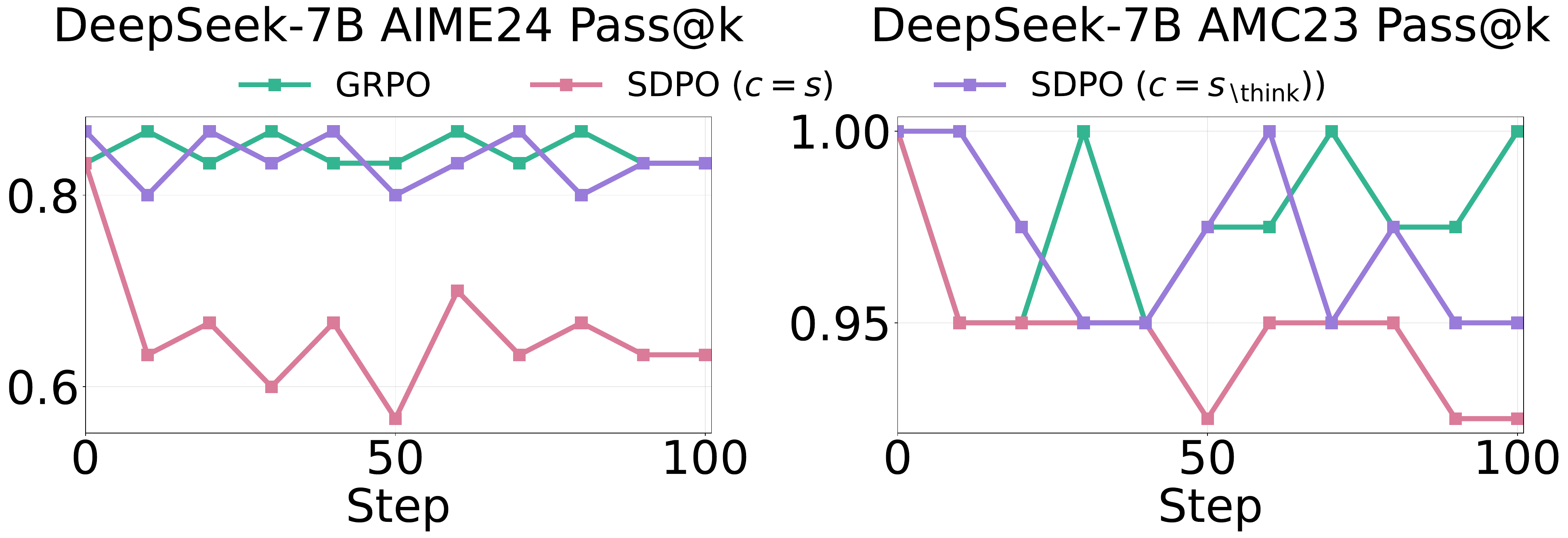}
        \caption{DeepSeek-Distill-7B}
        \label{fig:pass16_deepseek}
    \end{subfigure}
    \begin{subfigure}[b]{0.49\linewidth}
        \centering
        \includegraphics[width=\linewidth]{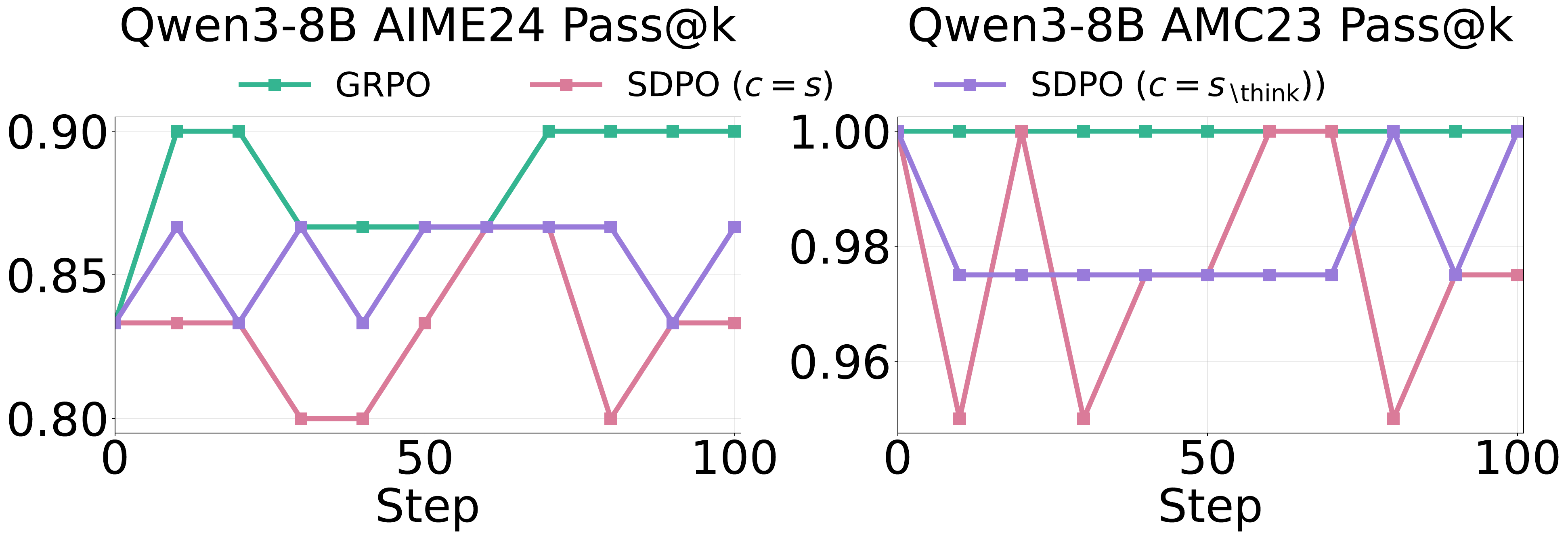}
        \caption{Qwen3-8B (thinking mode enabled)}
        \label{fig:pass16_qwen3}
    \end{subfigure}
    \caption{Pass@16 on AMC23 and AIME24 for GRPO and SDPO across training steps.}
    \label{fig:pass16_comparison}
\end{figure}

In addition to the acc@16 scores for DeepSeek-Distill-7B and Qwen3-8B (Thinking Mode: ON) presented in Figures~\ref{fig:ds_amc23},~\ref{fig:ds_aime24},~\ref{fig:qwen3_8b_think_amc23}, and~\ref{fig:qwen3_8b_think_aime24}, we report pass@16 scores in Figure~\ref{fig:pass16_comparison}. The gap between GRPO and SDPO in pass@16 is larger for DeepSeek-Distill-7B than for Qwen3-8B, and more pronounced on the harder benchmark (AIME24) compared to AMC23.

\newpage
\section{Qualitative Analysis}

\subsection{Unguided vs.\ Solution-Guided Generation} \label{appendix:qualitative_analysis}
We analyze examples of $\mathcal{D}_{\text{ug}}$ (Unguided responses) and $\mathcal{D}_{\text{sg}}$ (Solution-guided responses) introduced in Section~\ref{sec:sft}. Below, we present the two responses generated for the same problem. As illustrated, the unguided response exhibits substantially more uncertainty and exploratory reasoning, frequently revisiting assumptions, considering alternative approaches, and self-correcting along the way. In contrast, the solution-guided response proceeds in a more concise and direct manner, arriving at each intermediate result without detours.

\vspace{4pt}
\noindent\textbf{Problem.} How many ways can $8$ mutually non-attacking rooks be placed on a $9 \times 9$ chessboard so that all $8$ rooks are on squares of the same color? (Two rooks are said to be attacking each other if they are placed in the same row or column of the board.) Please reason step by step, and put your final answer within $\boxed{\phantom{0}}$.
\vspace{4pt}

\begin{tcolorbox}[
  colback=white,
  title=\small{Unguided Response -- Exploratory Reasoning},
  colframe=c-green-dark,
  coltitle=black,
  colbacktitle=c-green-light,
  breakable,
]
\small
 
Alright, so I have this problem where I need to figure out how many ways I can place 8 mutually non-attacking rooks on a 9$\times$9 chessboard, and all of them have to be on squares of the same color. Hmm, let me break this down.
 
First, I know that a chessboard is 8$\times$8, but this is a 9$\times$9 chessboard. So, it's one square larger on each side. Also, non-attacking rooks mean that no two rooks can be in the same row or column. Since it's 9$\times$9, if I were placing 9 rooks, they would each occupy their own row and column. But here, we're placing 8 rooks, so one row and one column will be empty.
 
But wait, the problem also says all rooks must be on squares of the same color. So, not only do they have to not attack each other, but they all have to be on, say, white squares or all on black squares.
 
[\ldots]
 
So, perhaps, the number of ways to place 8 non-attacking rooks on white squares would be equal to the number of permutation matrices on the white squares.
 
Similarly for black squares.
 
But how do we compute that?
 
I recall that in combinatorics, the number of permutation matrices avoiding certain positions is given by the inclusion-exclusion principle, but that might get complicated.
 
Alternatively, maybe we can model this as a bipartite graph matching problem.
 
[\ldots]
 
But I don't know the exact formula for that.
 
Alternatively, perhaps we can use the principle of inclusion-exclusion or recursion.
 
Alternatively, perhaps notice that the white squares form a bipartite graph which is a union of two complete bipartite graphs, but that might not hold.
 
Wait, let me consider the structure of the white squares. [\ldots] So, the bipartite graph is actually composed of two complete bipartite graphs: one connecting even rows to even columns, and the other connecting odd rows to odd columns.
 
Therefore, the bipartite graph is actually two disjoint complete bipartite graphs: one between the odd rows (5 rows) and odd columns (5 columns), and another between the even rows (4 rows) and even columns (4 columns).
 
So, the white squares form two separate complete bipartite graphs: $K_{5,5}$ and $K_{4,4}$.
 
[\ldots]
 
So, to get a matching of size 8, we need to have $k$ edges from $K_{5,5}$ and $8 - k$ edges from $K_{4,4}$, where $k$ ranges from 4 to 5.
 
If $k=4$, then $8 - 4 = 4$ edges from $K_{4,4}$.
 
If $k=5$, then $8 - 5 = 3$ edges from $K_{4,4}$.
 
[\ldots]
 
So, for $K_{5,5}$, the number of matchings of size 4 is $\binom{5}{4}^2 \cdot 4! = 25 \times 24 = 600$.
 
Similarly, the number of matchings of size 5 in $K_{5,5}$ is $\binom{5}{5}^2 \cdot 5! = 1 \times 120 = 120$.
 
For $K_{4,4}$, the number of matchings of size 4 is $\binom{4}{4}^2 \cdot 4! = 1 \times 24 = 24$.
 
And the number of matchings of size 3 is $\binom{4}{3}^2 \cdot 3! = 16 \times 6 = 96$.
 
Therefore: $600 \times 24 + 120 \times 96 = 14{,}400 + 11{,}520 = 25{,}920$.
 
[\ldots]
 
In $K_{5,4}$, the number of perfect matchings (size 4) is $\binom{5}{4} \cdot \binom{4}{4} \cdot 4! = 5 \times 1 \times 24 = 120$.
 
Similarly, in $K_{4,5}$, the number of perfect matchings is $\binom{4}{4} \cdot \binom{5}{4} \cdot 4! = 1 \times 5 \times 24 = 120$.
 
Therefore, the number of matchings of size 8 in the black square bipartite graph is $120 \times 120 = 14{,}400$.
 
Therefore, total is $25{,}920 + 14{,}400 = 40{,}320$.
 
But wait, $40{,}320$ is $8! = 40{,}320$. Hmm, that's interesting.
 
But why?
 
Is there a combinatorial interpretation where the number of ways is $8!$ multiplied by something?
 
Wait, $8!$ is 40320, which is exactly the number of ways we got.
 
But why is it exactly $8!$? It seems coincidental.
 
[\ldots]
 
Wait, perhaps it's best to just accept that the total number is $40{,}320$.
 
So, the final answer is $\boxed{40{,}320}$.
\end{tcolorbox}
 
\begin{tcolorbox}[
  colback=white,
  title=\small{Solution-Guided Response -- Concise, Direct Reasoning},
  colframe=c-pink-dark,
  coltitle=black,
  colbacktitle=c-pink-light,
  breakable,
]
\small
 
Alright, so I have this problem where I need to figure out how many ways I can place 8 mutually non-attacking rooks on a 9$\times$9 chessboard, and all of them have to be on squares of the same color. Hmm, okay, let's break this down step by step.
 
First, I know that a rook can move any number of squares vertically or horizontally, so non-attacking rooks must be placed on different rows and different columns. That means for 8 rooks, we need 8 distinct rows and 8 distinct columns.
 
[\ldots] Specifically, there are 41 squares of one color and 40 of the other. [\ldots] So white squares are 41 and black squares are 40.
 
So, I need to calculate the number of ways to place 8 non-attacking rooks on white squares and then on black squares separately, and then add them together.
 
Starting with white squares. [\ldots] The white square bipartite graph has two components: $K_{5,5}$ and $K_{4,4}$. Similarly, [\ldots] the black square bipartite graph has two components: $K_{5,4}$ and $K_{4,5}$.
 
[\ldots] The number of ways to choose 4 edges from $K_{5,5}$ is $\binom{5}{4}^2 \cdot 4! = 25 \times 24 = 600$. The number of ways to choose 5 edges from $K_{5,5}$ is $\binom{5}{5}^2 \cdot 5! = 1 \times 120 = 120$. Similarly, for $K_{4,4}$, the number of ways to choose 4 edges is $\binom{4}{4}^2 \cdot 4! = 1 \times 24 = 24$, and choosing 3 edges is $\binom{4}{3}^2 \cdot 3! = 16 \times 6 = 96$.
 
So, the total number of ways for white squares is $(600 \times 24) + (120 \times 96) = 14{,}400 + 11{,}520 = 25{,}920$.
 
For black squares, [\ldots] the total number of ways is $14{,}400$.
 
Adding both cases together, we get $25{,}920 + 14{,}400 = \boxed{40{,}320}$.
\end{tcolorbox}

\subsection{GRPO vs.\ SDPO Generation} \label{appendix:qualitative_grpo_sdpo}

\paragraph{Problem (AIME24).}
Find the number of ways to place a digit in each cell of a 2$\times$3 grid so that the sum of the two numbers formed by reading left to right is 999, and the sum of the three numbers formed by reading top to bottom is 99. The grid below is an example of such an arrangement because $8+991=999$ and $9+9+81=99$:
$$\begin{array}{|c|c|c|} \hline 0 & 0 & 8 \\ \hline 9 & 9 & 1 \\ \hline \end{array}$$
The correct answer is $\boxed{45}$.
 
GRPO explores with repeated reflections, deriving the extra constraint $a+b+c=8$ beyond the pairwise sums, and correctly counts $\binom{10}{2}=45$ solutions. SDPO proceeds directly and stops after deriving $a+d=b+e=c+f=9$, multiplying $10\times10\times10=1000$ without checking the row-sum equation, thereby missing the additional constraint $a+b+c=8$.
 
\begin{tcolorbox}[
  colback=white,
  title=\small{DeepSeek-R1-Distill-Qwen-7B GRPO Generation (Correct: 45)},
  colframe=c-green-dark,
  coltitle=black,
  colbacktitle=c-green-light,
  breakable,
]
\small
 
Okay, so I need to figure out how many ways there are to fill a 2$\times$3 grid with digits such that two conditions are satisfied: (1) the sum of the two numbers formed by reading left to right is 999, and (2) the sum of the three numbers formed by reading top to bottom is 99.
 
Let me visualize the grid as
$\begin{array}{|c|c|c|} \hline a & b & c \\ \hline d & e & f \\ \hline \end{array}$.
The first condition gives $(100a+10b+c) + (100d+10e+f) = 999$, and the second gives $(10a+d) + (10b+e) + (10c+f) = 99$.
 
[\ldots]
 
So we have:
 
Equation (A): $100\,S_1 + 10\,S_2 + S_3 = 999$, where $S_1 = a+d$, $S_2 = b+e$, $S_3 = c+f$.
 
Equation (B): $10T + U = 99$, where $T = a+b+c$, $U = d+e+f$.
 
[\ldots] Since $S_1$ is $a+d$, $S_1$ can be from 0 to 18. But if $S_1$ is 10, then $100 S_1 = 1000$, which is already larger than 999. Therefore, $S_1$ can be from 0 to 9.
 
For $S_1 = 9$: $100 \cdot 9 = 900$. Then $10 S_2 + S_3 = 99$. 
 
For $S_1 = 8$: $100 \cdot 8 = 800$. Then $10 S_2 + S_3 = 199$. But $10 S_2 + S_3$ can be at most $10\cdot 18 + 18 = 198$. So 199 is too big. Therefore $S_1$ can't be 8.
 
For $S_1 = 9$, $10 S_2 + S_3 = 99$. Since $S_2$ and $S_3$ are between 0 and 18, $10 S_2 + S_3$ can be between 0 and 198. So 99 is possible.
 
[\ldots] Therefore $S_2$ must be 9. Hence, $S_2 = 9$, then $S_3 = 99 - 90 = 9$.
 
Let me check other $S_1$ values. For $S_1 = 7$: $10 S_2 + S_3 = 299$, even larger. Hence impossible. Similarly for lower $S_1$. Hence, only $S_1 = 9$ is possible.
 
\textbf{Wait, let me double-check that $S_1 = 9$ is the only possible value.} For $S_1 = 9$, we have $10 S_2 + S_3 = 99$, which requires $S_2 = 9$ and $S_3 = 9$. [\ldots] So only $S_2 = 9$ and $S_3 = 9$.
 
Therefore, Equation (A) only holds when $S_1 = S_2 = S_3 = 9$.
 
[\ldots] From $T + U = 27$ and $10T + U = 99$: subtracting gives $9T = 72$, so $T = 8$, $U = 19$.
 
[\ldots] Substituting $d = 9-a$, $e = 9-b$, $f = 9-c$, Equation 5 ($d+e+f=19$) is automatically satisfied. The main constraint reduces to $a + b + c = 8$ with $a,b,c \in \{0,\ldots,9\}$.
 
Since $8 < 9$, the upper bound is automatic, so the count is the number of non-negative integer solutions to $a+b+c=8$, which is $\binom{10}{2} = 45$.
 
Since $d = 9 - a$ and $a \le 8$, $d$ ranges from 1 to 9 --- all valid digits. Similarly for $e, f$. Therefore, all 45 solutions are valid.
 
\textbf{Wait, let me check if there are constraints I missed.} When forming the numbers $N_1$ and $N_2$, they can have leading zeros, which is fine since the example itself has $008$ in the top row. [\ldots] So there is no problem with leading zeros.
 
Let me verify with the given example: $a=0, b=0, c=8$, so $d=9, e=9, f=1$. Columns: $09 + 09 + 81 = 99$ \checkmark. Rows: $008 + 991 = 999$ \checkmark.
 
Another example: $a=8, b=0, c=0$. Then $d=1, e=9, f=9$. Top: 800, bottom: 199. $800+199=999$ \checkmark. Columns: $81+9+9=99$ \checkmark.
 
Another: $a=0, b=1, c=7$. Then $d=9, e=8, f=2$. $17+982=999$ \checkmark. $9+18+72=99$ \checkmark.
 
\textbf{Wait, one more check --- what if leading zeros aren't allowed?} Then the top row would need $a \neq 0$, but the example given has $a=0$, so that interpretation is invalid. Therefore, leading zeros are allowed, and the answer is indeed 45.
 
Therefore, I think the correct answer is $\boxed{45}$.
 
\end{tcolorbox}

\begin{tcolorbox}[
  colback=white,
  title=\small{DeepSeek-R1-Distill-Qwen-7B SDPO Generation (Incorrect: 1000)},
  colframe=c-pink-dark,
  coltitle=black,
  colbacktitle=c-pink-light,
  breakable,
]
\small
 
Okay, so I need to figure out how many ways there are to fill a 2$\times$3 grid with digits so that the sum of the two numbers formed by reading left to right is 999, and the sum of the three numbers formed by reading top to bottom is 99.
 
Labeling the grid as Row 1: $a, b, c$ and Row 2: $d, e, f$, the two conditions give:
\begin{align*}
\text{(1)}\quad & 10(a+b+c) + (d+e+f) = 99 \\
\text{(2)}\quad & 100(a+d) + 10(b+e) + (c+f) = 999
\end{align*}
 
Let $S = a+b+c$, $T = d+e+f$, and $x = a+d$, $y = b+e$, $z = c+f$. Then (1) becomes $10S + T = 99$ and (2) becomes $100x + 10y + z = 999$.
 
Since $x+y+z = S+T$, substituting $T = 99 - 10S$ gives $x+y+z = 99 - 9S$. Subtracting this from (2):
$$99x + 9y = 900 + 9S \;\Rightarrow\; 11x + y = 100 + S.$$
 
[\ldots] From $0 \le 99 - 10S \le 27$ we get $S \in \{8, 9\}$.
 
\textbf{Case 1: $S = 8$.} Then $T = 19$, $11x + y = 108$, and $x + y + z = 27$. Substituting $y = 108 - 11x$ gives $z = 10x - 81$. The constraint $0 \le z \le 18$ forces $x = 9$, so $y = 9$, $z = 9$. Therefore $a+d = 9$, $b+e = 9$, $c+f = 9$.
 
For each of the three pairs, there are 10 possibilities (since for each digit from 0 to 9, there's exactly one digit that adds up to 9). So for each of the three pairs, there are 10 possibilities. Therefore, the total number of solutions in this case is $10 \times 10 \times 10 = 1000$.
 
\textbf{Case 2: $S = 9$.} Then $T = 9$, $11x + y = 109$, and $x + y + z = 18$. Substituting gives $z = 10x - 91$. The constraint $0 \le z \le 18$ forces $x = 10$, but then $y = 109 - 110 = -1$, which is invalid. Therefore, this case is invalid.
 
Thus, only Case 1 is valid, leading to 1000 possible configurations.
 
The digits $a,b,c,d,e,f$ must be digits between 0 and 9. In Case 1, we have $a+d=9$, so each pair $(a,d)$ can be any pair of digits adding to 9, which is valid. Similarly for $(b,e)$ and $(c,f)$. So there are no invalid digit values here. Therefore, all 1000 configurations are valid.
 
Let me check with the example given in the problem:
$\begin{array}{|c|c|c|} \hline 0 & 0 & 8 \\ \hline 9 & 9 & 1 \\ \hline \end{array}$.
Horizontal sum: $008 + 991 = 999$ \checkmark. Vertical sum: $09 + 09 + 81 = 99$ \checkmark. Each of the pairs $(a,d) = (0,9)$, $(b,e) = (0,9)$, $(c,f) = (8,1)$ adds to 9, which fits Case 1. So this example is counted in the 1000 configurations.
 
Hence, the answer should be $\boxed{1000}$.
 
\end{tcolorbox}

\section{Disclosure of LLM Usage}

We used LLMs for conducting the LLM-as-a-Judge experiments described in Appendix \ref{appendix:epistemic_llm_as_judge}, refining plot designs, and improving the clarity of our writing. LLMs were not involved in research ideation, and their role in writing was limited to polishing author-written text.

\end{document}